\documentclass[preprint,3p,12pt]{elsarticle}

\usepackage{amssymb}
\usepackage{amsmath}
\usepackage{amsmath,amsfonts}
\usepackage{algorithmic}
\usepackage{algorithm}
\usepackage{array}
\usepackage[caption=false,font=footnotesize,labelfont=sf,textfont=sf]{subfig}
\usepackage{textcomp}
\usepackage{stfloats}
\usepackage{url}
\usepackage{verbatim}
\usepackage{graphicx}
\usepackage{threeparttable}
\usepackage{booktabs}
\usepackage{pifont}
\usepackage{rotating}
\usepackage{multirow}
\usepackage{dashrule}
\newtheorem{Theorem}{Theorem}
\usepackage{bbding}
\usepackage{amsmath}
\usepackage{amssymb}
\usepackage{diagbox}
\usepackage{xcolor}
\usepackage{makecell}

\newtheorem{Proposition}{Proposition}
\newtheorem{Remark}{Remark}
\newtheorem{Proof}{Proof}

\journal{Neurocomputing}

\begin{document}
\begin{sloppypar}
\begin{frontmatter}



\title{Enhancing Transformer-based models for Long Sequence Time Series Forecasting via Structured Matrix}
\author[1]{Zhicheng Zhang}
\author[1]{Yong Wang\corref{cor}}
\ead{cla@uestc.edu.cn} 
\cortext[cor]{Corresponding author}
\author[1]{Shaoqi Tan}
\author[1]{Bowei Xia}
\author[1]{Yujie Luo}

\affiliation[1]{
	organization={Laboratory of Intelligent Collaborative Computing Technology, University of Electronic Science and Technology of China},
	city={Chengdu},
	postcode={611731}, 
	state={Sichuan},
	country={China}}

\begin{abstract}
Recently, Transformer-based models for long sequence time series forecasting have demonstrated promising results. The self-attention mechanism as the core component of these Transformer-based models exhibits great potential in capturing various dependencies among data points. Despite these advancements, it has been a subject of concern to improve the efficiency of the self-attention mechanism. Unfortunately, current specific optimization methods are facing the challenges in applicability and scalability for the future design of long sequence time series forecasting models. Hence, in this article, we propose a novel architectural framework that enhances Transformer-based models through the integration of Surrogate Attention Blocks (SAB) and Surrogate Feed-Forward Neural Network Blocks (SFB). The framework reduces both time and space complexity by the replacement of the self-attention and feed-forward layers with SAB and SFB while maintaining their expressive power and architectural advantages. The equivalence of this substitution is fully demonstrated. The extensive experiments on 10 Transformer-based models across five distinct time series tasks demonstrate an average performance improvement of 12.4\%, alongside 61.3\% reduction in parameter counts.\footnote{Our code is publicly available at \url{https://github.com/newbeezzc/MonarchAttn}}.
\end{abstract}


\end{frontmatter}



\section{Introduction}
Time series data represent a sequence of values for a specific statistical indicator that are arranged systematically in chronological order. It can offer us immensely valuable insights for strategic planning and decision-making, if we are able to accurately predict the future data in vital sectors such as traffic management\cite{Liu2023cikm}, healthcare monitoring\cite{Piccialli2021inffus} and financial analysis\cite{MATHEMATICS2023He}. 

Time series forecasting is to predict the future values $\hat{Y} = \{y_{N+1}, \cdots, y_{N+l}\}$ in proximity to the ground truth, based on the available observed data $Z = \{z_{1}, \cdots, z_{N}\}$. Here, $l$ represents the number of future steps to be predicted, $N$ denotes the length of the observed data sequence. With the advent of the big data era, long sequence time series forecasting (LSTF) is proposed to handle the prediction tasks characterized by both large prediction steps (large $l$) and extensive historical data (large $N$). 

In recent years, Transformer-based models have demonstrated remarkable success in LSTF tasks. These models leverage self-attention mechanisms to capture intricate trend, seasonal and residual patterns, which excel at uncovering latent dependencies between arbitrary data points in a time series sequence. However, the lack of inductive biases critically undermines its ability to capture temporal causality and trend features, which in turn results in slower model convergence during the training phase. Moreover, its quadratic time and memory complexity with respect to the sequence length $N$ renders these models computationally intensive and difficult to deploy in resource-constrained environments. As the length of time series data increases, the computational burden grows exponentially, creating a significant barrier to model scalability and real-world implementation. In addition, training a Transformer-based model to achieve satisfactory forecasting results requires extensive data samples. To optimize the Transformer and enhance its efficiency, current research predominantly focuses on two fundamental properties of self-attention mechanisms in LSTF tasks: locality \cite{Vaswani2021cvpr} and sparsity \cite{Liu2021corr}. 

The locality property suggests that tokens primarily interact with their neighbors. With such observation, the works in \cite{Cordonnier2020iclr, Han2022iclr} theoretically proved that multi-head self-attention and convolution are similar in expressiveness, which suggests that it can obtain better efficiency by using convolutional layers to replace self-attention layers. Similarly, other works like PANet\cite{Zhao2022IEEE} and Autoformer\cite{Wu2021NIPS} further integrated convolutional operations into self-attention layers guided by this property, which led to reduced model complexity and better prediction accuracy. However, convolution is confined to a fixed receptive field and cannot capture the long-term dependence of the time series. The sparsity arises from the fact that most elements in the attention scoring matrix are close to zero, indicating that only few tokens are crucial for LSTF tasks. Exploiting this property, researchers in Linformer\cite{Sinong2020corr}, Informer\cite{Zhou2021AAAI} and LogTrans\cite{Li2019NeurIPS} have employed techniques such as down-sampling and memory compression to alleviate the time and memory complexity of the Transformer-based models. These two properties offer valuable insights into optimizing Transformer-based models for LSTF, ultimately leading to improved performance and efficiency. Current works focus on the model-level design, apparently ignoring the success of previous Transformer-based models in the LSTF tasks.

Another branch of work attempts to create a lightweight model through parameter quantification\cite{koster2017flexpoint, jacob2018quantization, zhou2016dorefa}, LSTF-specific knowledge distillation\cite{hinton2015distilling, ba2014deep} and network pruning\cite{lecun1989optimal, han2015learning}. These works are model and task specific because they assume the existence of a well-trained model. Furthermore, they have to strike a balance between efficiency, accuracy and expressiveness.

In order to inherit the successful experience of existing Transformer-based models and enhance their computational efficiency without sacrificing performance, this research aims to address the following critical challenges:
(1) Can we develop an innovative computational approach that achieves sub-quadratic computational complexity with respect to both sequence length and model dimension, while simultaneously ensuring hardware efficiency?
(2) Based on such an approach, is it feasible to design modular components that can seamlessly replace the computationally intensive layers in current Transformer-based architectures, thereby preserving the fundamental strengths of the original design?
(3) To what extent can the proposed replacement method maintain the representational capacity and expressive power equivalent to the original Transformer-based models?

In addressing these questions, we have condensed our research into three main aspects:

\begin{itemize}
	\item We introduce a generalizable optimization framework which is capable of systematic reduction in computational and memory consumption across diverse Transformer-based architectures, while preserving their prediction accuracy. The framework comprises two innovative modules: the Surrogate Attention Block (SAB) and the Surrogate Feed-Forward Network Block (SFB). By presenting a model-agnostic optimization strategy, our approach offers a versatile and principled solution to mitigate the inherent computational inefficiencies of Transformer-based models.
	
	\item Through a comprehensive and systematic analysis, the broad applicability of our proposed optimization framework is demonstrated across various Transformer-based architectures. We rigorously establish the mathematical equivalence between original and optimized variants via the SAB. Moreover, it is mathematically proved that SAB preserves the critical capability of capturing both long-range and short-range dependencies in sequential data unlike traditional convolutional operations with fixed receptive fields. Finally, SAB is proved to be a linear time-invariant (LTI) system with favorable training dynamics and theoretical convergence properties.
	
	\item Extensive comparative experiments are conducted with the proposed framework by applying it to 5 distinct downstream tasks. The experimental evaluation results, comprising 2,769 performance tests, demonstrate that the models optimized by our framework consistently outperform their original counterparts in 72.4\% of the evaluated tasks. Notably, our approach achieves an average performance improvement of 12.4\%, accompanied by 66.1\% reduction in FLOPS and a substantial 61.3\% reduction in model parameters.
	
\end{itemize}

The rest of the article is organized as follows: The related works are reviewed in Section \ref{sec:related_works}. Section \ref{sec:preliminaries} reviews the prior knowledge needed for the article. Section \ref{sec:model_architecture} describes the two surrogate blocks and proves the equivalence of the attention layer and the convolution. Section \ref{sec:theoretical_analysis} contains a proof of the expressive power of the substitution module, an analysis of the complexity and a proof of stability. The experimental evaluations are presented in Section \ref{sec:experiments}. Finally, some discussion and conclusions are drawn in Section \ref{sec:conclusions}.

\section{Related works}\label{sec:related_works}
Currently, there is a substantial body of Transformer-based research in the time series domain. These innovations can be broadly categorized into two main types: optimizations of the attention mechanism and advancements in embedding techniques and architectural design.

\subsection{Attention Mechanism Optimizations in Time Series Forecasting}
The computational challenges of Transformer-based models in long sequence time series forecasting have prompted diverse optimization strategies. Pioneering works like LogTrans\cite{Li2019NeurIPS} and Informer\cite{Zhou2021AAAI} introduced groundbreaking techniques to reduce computational complexity by strategically leveraging the sparsity and locality properties of attention scoring matrices, successfully reducing algorithmic complexity from quadratic to $O(N\log N)$. These seminal contributions not only addressed critical computational bottlenecks but also significantly expanded the potential applications of Transformer models in time series forecasting.

Building upon these foundational insights, subsequent research explored more radical architectural modifications. TCCT\cite{Shen2022Neurocomputing}, GCformer\cite{Zhao2023CoRR}, Autoformer\cite{Wu2021NIPS} and MODERNTCN\cite{anonymous2024iclr} proposed to replace self-attention layers with convolutional layers, recognizing their efficiency in capturing local dependencies and potential for reduced parameter counts. This line of research highlighted the viability for alternative architectural approaches to address the computational bottlenecks of vanilla Transformers.

\subsection{Embedding and Architectural Modification}
The challenge of capturing complex temporal dependencies led researchers to develop sophisticated embedding and architectural strategies. Enriched positional embedding mechanisms, explored by the works in \cite{Zerveas2021kdd, Lim2021IJoF, Zhou2021AAAI, Zhou2022PMLR, Tonekaboni2020nips, Cheng2023iclr} aimed to improve the model's ability to capture both local patterns and long-range dependencies in time series sequences.

Hybridization emerged as another key approach, combining neural network components such as \cite{Dai2021conf, Li2019NeurIPS} and LSTM\cite{Yang2022tkde} with attention mechanisms. More than this, Crossformer\cite{zhang2023ICLR} proposed a novel cross-dimension interaction mechanism that captures intricate relationships across different feature dimensions. Pyraformer\cite{Liu2022ICLR} develops a pyramid-structured approach that enables multi-scale temporal feature extraction. These approaches sought to address the limitations of vanilla Transformer architectures by introducing complementary neural network paradigms, ultimately aiming to enhance model capacity and generalization.

Complementing architectural innovations, researchers developed sophisticated matrix factorization techniques to enhance model efficiency. Frequency domain transformation emerged as a promising approach, with the works like FEDformer\cite{Zhou2022PMLR} and ETSformer\cite{Gerald2022corr} leveraging Fast Fourier Transform (FFT) to reconstruct time series representations. These methods demonstrated the potential of domain transformation in reducing computational complexity while preserving critical temporal information. Parallel to frequency-based approaches, matrix decomposition techniques such as Singular Value Decomposition\cite{abdelhakim2018optimized, braconnier2011towards, zhang2018stabilizing}, Non-negative Matrix Factorization\cite{devarajan2008nonnegative, lin2020optimization} and Sparse Coding\cite{aharon2006k, lee2006efficient} provided alternatives for reducing high-dimensional self-attention scoring matrices. These techniques shared a common goal of reducing computational overhead by projecting complex representations into more manageable lower-dimensional spaces.

Despite the significant progress made in Transformer-based time series forecasting, most existing models have primarily focused on optimizing computational complexity and memory usage, with relatively little attention given to hardware-friendly characteristics. This oversight limits the practical deployment of these models, particularly in resource-constrained environments where hardware compatibility is critical. Addressing this gap, Table \ref{tab:xformers} compares our method with other Transformer-based models on various time series tasks from recent years. Our approach not only achieves a computational complexity of $O(N^{3/2})$ but also incorporates hardware-friendly design principles without requiring modifications to the original architecture. This combination allows our method to enhance efficiency while retaining the architectural strengths of the original Transformer models, making it more suitable for deployment across diverse hardware platforms.
This allows our method to efficiently handle long sequence time series forecasting tasks, where other Transformer-based models may encounter limitations in terms of computational efficiency or memory requirements.

In addition to the previously discussed optimization techniques, there is anohter branch of techniques that can be employed to improve computational efficiency. Branching\cite{he2014learning, sprangle1997agree} provides advanced architectural optimization, allowing for more efficient network structures. Quantification\cite{koster2017flexpoint, jacob2018quantization, zhou2016dorefa} enables model compression through parameter reduction, significantly reducing computational requirements. The process of knowledge distillation\cite{hinton2015distilling, ba2014deep} is useful in transferring complex model knowledge to smaller models. Pruning\cite{lecun1989optimal, han2015learning} involves removing non-critical neurons or connections to streamline the network to improve computational efficiency.

\begin{table}
	\caption{Comparison of Transformer Variants. P is the patch size. N generally refers to the sequence length. D is the dimension of the data. $ L_{seg}$ represents segment length. $\omega$ denotes fixed window size. $c$ is a hyperparameter relying on the characteristics of the time series dataset. For task types, `F' denotes forecasting, `I' denotes imputation, `C' denotes classification and `A' denotes anomaly detection.}
	\begin{center}
		\scriptsize
		\resizebox{\textwidth}{!}{
			\begin{tabular}{ccccccccc}
				\toprule
				\textbf{Model} & \textbf{Year} & \textbf{Complexity} & \textbf{Task Type} & \textbf{Locality} & \textbf{Sparsity} & \textbf{Attention Optimization} & \textbf{Architectural Modification} & \textbf{Hardware-friendly} \\
				\midrule
				LogTrans\cite{Li2019NeurIPS}  & 2019  &  $ O(N\log{N}) $  & F     & \ding{51}  & \ding{51}  &   \ding{51}  &   \ding{51}  &   \ding{55}  \\
				AST\cite{Wu2020NEURIPS}  & 2020  &  $ O(N\log{N}) $  & F     & \ding{55}  & \ding{51}  &   \ding{51}  &   \ding{51}  &   \ding{55}  \\
				Informer\cite{Zhou2021AAAI}  & 2021  &  $ O(N\log{N}) $  & F     & \ding{55}  & \ding{51}  &   \ding{51}  &   \ding{51}  &   \ding{55}  \\
				Autoformer\cite{Wu2021NIPS}  & 2021  &  $ O(N\log{N}) $   & F     & \ding{51}  & \ding{55}  &   \ding{51}  &   \ding{51}  &   \ding{55} \\
				TFT\cite{Lim2021IJoF}  & 2021  &  $ O(N^2) $  & F     & \ding{55}  & \ding{55}  &   \ding{55}  &   \ding{51}  &   \ding{55}  \\
				Yformer\cite{Kiran2021CoRR}  & 2021  &  $ O(N\log{N}) $  & F     & \ding{55}  & \ding{51}  &   \ding{51}  &   \ding{51}  &   \ding{55}  \\
				Spacetimeformer\cite{Grigsby2021CoRR}  & 2021  &  $ O(N^2) $  & F     & \ding{55}  & \ding{55}  &   \ding{55}  &   \ding{51}  &   \ding{55}  \\
				FEDformer\cite{Zhou2022PMLR}  & 2022  &  $ O(N) $  & F     & \ding{51}  & \ding{55}  &   \ding{51}  &   \ding{51}  &   \ding{55}  \\
				NST\cite{Liu2022NEURIPS}  & 2022  &  $ O(N^2) $  & F     & \ding{55}  & \ding{55}  &   \ding{55}  &   \ding{51}  &   \ding{55}  \\
				Pyraformer\cite{Liu2022ICLR}  & 2022  &  $ O(N) $  & F     & \ding{51}  & \ding{51}  &   \ding{51}  &   \ding{51}  &   \ding{55}  \\
				ETSformer\cite{Gerald2022corr}  & 2022  &  $ O(N\log{N}) $  & F     & \ding{55}  & \ding{55}  &   \ding{51}  &   \ding{51}  &   \ding{55}  \\
				TCCT\cite{Shen2022Neurocomputing}  & 2022  &  $ O(\frac{1}{2} N^2) $  & F     & \ding{51}  & \ding{55}  &   \ding{51}  &   \ding{51}  &   \ding{55}  \\
				Quatformer\cite{Chen2022KDD}  & 2022  &  $ O(2cN) $  & F     & \ding{55}  & \ding{55}  &   \ding{51}  &   \ding{51}  &   \ding{55}  \\
				TDformer\cite{Zhang2022NeurIPS}  & 2022  &  $ O(N\log{N}) $  & F     & \ding{55}  & \ding{55}  &   \ding{51}  &   \ding{51}  &   \ding{55}  \\
				Crossformer\cite{zhang2023ICLR}  & 2023  &  $ O(\frac{D}{L_{seg}^2} N^2 ) $  & F     & \ding{55}  & \ding{55}  &   \ding{51}  &   \ding{51}  &   \ding{55}  \\
				FPPformer\cite{Shen2023IEEE}  & 2023  &  $ O(NP) $   & F     & \ding{55}  & \ding{55}  &   \ding{51}  &   \ding{51}  &   \ding{55}  \\
				GCformer\cite{Zhao2023CoRR}  & 2023  &  $ O(N^2) $  & F     & \ding{51}  & \ding{55}  &   \ding{51}  &   \ding{51}  &   \ding{55}  \\
				Conformer\cite{Li2023ICDE}  & 2023  &  $ O(\omega N) $  & F     & \ding{51}  & \ding{55}  &   \ding{51}  &   \ding{51}  &   \ding{55}  \\
				PDFormer\cite{Jiang2023AAAI}  & 2023  &  $ O(N^2) $  & F     & \ding{55}  & \ding{55}  &   \ding{55}  &   \ding{51}  &   \ding{55}  \\
				Preformer\cite{Du2023ICASSP}  & 2023  & $ O(\frac{(N^2}{L_{seg}}) $  & F     & \ding{55}  & \ding{55}  &   \ding{51}  &   \ding{51}  &   \ding{55}  \\
				Taylorformer\cite{Nivron2023CoRR}  & 2023  &  $ O(N^2) $  & F     & \ding{51}  & \ding{55}  &   \ding{55}  &   \ding{51}  &   \ding{55}  \\
				PatchTST\cite{Nie2023iclr}  & 2023  &  $ O(N^2) $  & F     & \ding{51}  & \ding{55}  &   \ding{55}  &   \ding{51}  &   \ding{55}  \\
				iTransformer\cite{liu2024ICLR}  & 2024  &  $ O(N^2) $  & F     & \ding{55}  & \ding{55}  &   \ding{55}  &   \ding{51}  &   \ding{55}  \\
				PAttn\cite{Tan2024NeurIPS}  & 2024  &  $ O(N^2) $  & F     & \ding{51}  & \ding{55}  &   \ding{55}  &   \ding{51}  &   \ding{55}  \\
				TimeXer\cite{wang2024NeurIPS}  & 2024  &  $ O(N^2) $  & F     & \ding{51}  & \ding{55}  &   \ding{55}  &   \ding{51}  &   \ding{55}  \\
				Ours  & 2024  &  $ O(N^{3/2}) $  & F+I+C+A & \ding{55}  & \ding{55}  &   \ding{51}  &   \ding{55}  &   \ding{51}  \\
				\bottomrule
			\end{tabular}%
		}
	\end{center}
	\label{tab:xformers}
\end{table}

\section{Preliminaries}\label{sec:preliminaries}

\subsection{Revisiting Transformer mechanisms}\label{subsec:MHSA}

\begin{figure}
	\centering
	\includegraphics[width=1.0\linewidth]{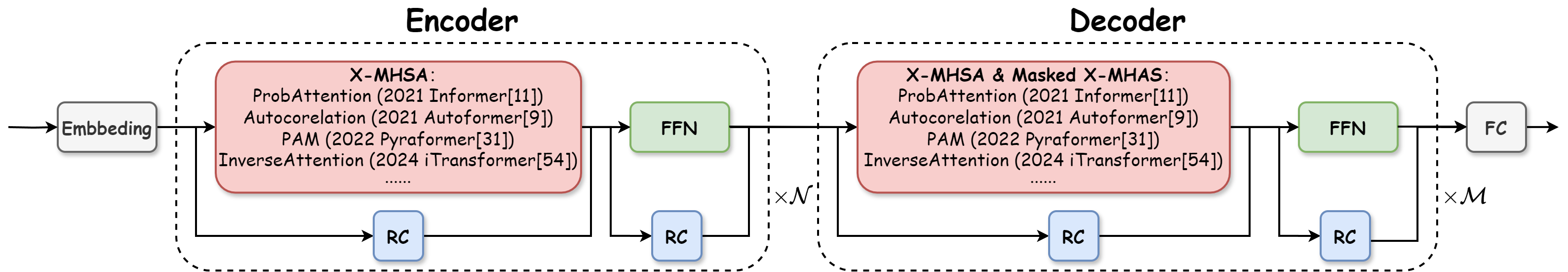}
	\caption{Abstract architecture for Transformer-based model. $\mathcal{N}$ denotes the number of Encoder Layers, $\mathcal{N}$ denotes the number of Decoder Layers.}
	\label{fig:x-former}
\end{figure}

Transformer-based models follow the encoding-decoding structure, which essentially comprises the components of Multi-Head Self-Attention (MHSA) layer, Feed-Forward Network (FFN) layer and Residual Connections (RC), Figure \ref{fig:x-former} illustrates the unified architecture of these models. 

An \textbf{MHSA layer} utilizes multiple self-attentions as heads to extract different types of features, which is widely used in practice. Let $z_t \in \mathbb{R}^m$ denote the observation of $m$ variables at time step $t$. Given a multivariate time series sequence $\mathbf{Z} = \{z_1, z_2, ..., z_N\} \in \mathbb{R}^{N \times m}$ for $N$ time steps, $\mathbf{Z}$ can be projected as the representation $\mathbf{X} \in \mathbb{R}^{N \times D_{in}}$. For the input representation $\mathbf{X}$, self-attention matches the sequence of queries $Q = \mathbf{X}W_{qry} \in \mathbb{R}^{N \times D_k}$ against the sequence of keys $K = \mathbf{X}W_{key} \in \mathbb{R}^{N \times D_k}$ by scaled dot-product. The output of an MHSA layer can be formulated as:\footnote{For easy-to-understand presentation, biases are excluded in the following equations.}
\begin{equation}
	\begin{aligned}
		SA^{(h)}(\mathbf{X}) := Softmax\left(\frac{Q^{(h)}\cdot(K^{(h)})^T}{\sqrt{D_k}} \right) \cdot \mathbf{X} \cdot W^{(h)}_{val}
	\end{aligned}
	\nonumber
\end{equation}
\begin{equation}
	\label{Eq:MHSA-X}
	\begin{aligned}
		MHSA(\mathbf{X}) := \mathop{concat}\limits_{h \in [H]}[SA^{(h)}(\mathbf{X})]\cdot W_{out} 
	\end{aligned}
	\nonumber
\end{equation}
where $W^{(h)}_{val} \in \mathbb{R}^{D_{in} \times D_v}$ and $W_{out} \in \mathbb{R}^{(HD_v) \times D_{out}}$ are learnable projection matrices. $H$ is the number of heads. 

An \textbf{FFN layer} typically consists of two-layer neural networks, which can be expressed as:
\begin{equation}					
	\label{Eq:FFN}
	\begin{aligned}
		FFN(\mathbf{X})=\sigma(\mathbf{X}\cdot W_1)\cdot W_2 ^T
	\end{aligned}
	\nonumber
\end{equation}
where $W_1$, $W_2$ $\in \mathbb{R}^{D_{in} \times D_m}$ are learnable parameter matrices, $\sigma$ is a non-linearity function such as $ReLU$.

An \textbf{RC} connects the inputs and outputs of a sub-layer in transformers such as FFN and MHSA. Considering the position of the normalization layer in transformers, there are currently two major definitions of residual connections, which are: 
\begin{equation}
	\begin{aligned}
		\mathbf{X} = LN(\mathbf{X}+F(\mathbf{X})) 
	\end{aligned}
	\nonumber
\end{equation}
and
\begin{equation}
	\begin{aligned}
		\mathbf{X} = \mathbf{X}+F(LN(\mathbf{X})) 
	\end{aligned}
	\nonumber
\end{equation}
where $LN(\cdot)$ is the layer normalization function, $F(\cdot)$ is a sub-layer (e.g., FFN or MHSA).

\subsection{Structured matrices}\label{subsec:Monarch}

Structured matrices are those with a sub-quadratic number of parameters and runtime. Large classes of structured matrices (e.g., Toeplitz-like\cite{Sindhwani2015nips}, ACDC\cite{Moczulski2015corr}, Fastfood\cite{Le2013pmlr} and Butterfly\cite{Dao2020iclr}) have demonstrated their capacities of replacing dense weight matrices in large neural networks, which can reduce their computation and memory requirements. Very recently, Monarch matrices were proposed to capture a wide class of linear transforms, including Hadamard transforms, Toeplitz matrices, ACDC matrices and convolutions. They are a sub-quadratic class of structured matrices that are hardware-efficient and expressive \cite{Fu2023corr}. A Monarch matrix $\mathbf{M}\in \mathbb{R}^{N\times N}$ of order-$p$ is defined as:
\begin{equation}
	\begin{aligned}
		\mathbf{M} = \left(\prod_{i=1}^p \mathbf{P}_i \mathbf{B}_i\right)\mathbf{P}_0
	\end{aligned}
	\nonumber
\end{equation}
where each $\mathbf{P}_i$ is associated with the `base $\sqrt[p]{N}$' variant of the bit-reversal permutation and $\mathbf{B}_i$ is a block-diagonal matrix with a block size of $b$. When setting $p = 2$ and $b = \sqrt{N}$, Monarch matrices require $O(N^{3/2})$ computed in a time series sequence of length $N$.

\subsection{Acceleration objectives}\label{subsec:obj}

Compute-bound and memory-bound \cite{Konstantinidis2015pdp} are two classes of operations on GPUs that mainly affect the runtime performance of deep learning models. The FLOP/s is used as a metric to determine the speed of these operations. Therefore, the objectives of accelerating long sequence time series forecasting models are sub-quadratic scaling with the input length $N$ and high FLOP utilization. Let $O_{\text{computation}}(\cdot)$ and $O_{\text{memory}}(\cdot)$ be the computation and memory complexity of a model. For a given well-designed Transformer-based model $f$, it is accelerated or enhanced as $f^{\star}$ that satisfies:

\begin{equation}
	\begin{aligned}
		O_{\text{computation}}\left( f^{\star}(\mathbf{X})\right) < O_{\text{computation}}\left( f(\mathbf{X})\right)
	\end{aligned}
	\nonumber
\end{equation}
\begin{equation}
	\begin{aligned}
		O_{\text{memory}}\left(f^{\star}(\mathbf{X})\right) < O_{\text{memory}}\left( f(\mathbf{X})\right)
	\end{aligned}
	\nonumber
\end{equation}

Subject to:

\begin{equation}
	\begin{aligned}
		E_{\hat{Y}}(\left\| f^{\star}(\mathbf{X}) - f(\mathbf{X}) \right\| ) < \varepsilon
	\end{aligned}
	\nonumber
\end{equation}
where $E(\left\|\cdot\right\|)$ is an evaluation function that measures the accuracy of $h$ steps time series forecasting results on labeled sequences $\hat{Y}=\{ \hat{y}_{N+1}, \hat{y}_{N+2},\cdots, \hat{y}_{N+l} \}$, such as mean square error. $\varepsilon$ indicates the difference between the output of $f^{\star}$ and $f$. Ideally, we suppose $\varepsilon$ can be ``ignored". The major notations used in this article are presented in Table \ref{Tab:symbols}.

\begin{table}[ht]
	\centering
	\caption{Notations and Descriptions}
	\begin{tabular}{cc}
		\toprule
		Notations & Descriptions \\
		\midrule
		$\mathbf{Z}$ & Input time series  \\
		$\mathbf{Y}$ & Ground truth future time series for forecasting  \\
		$\hat{\mathbf{Y}}$ & Predicted future time series  \\
		$\mathbf{X}$ & Embedding in model  \\
		$W$ & Weight matrix \\
		$\mathbf{M}$ & Structured matrix \\
		$Q, K, V$ & Query, key and value matrix in attention layers \\
		$N$ & Length of the input time series  \\
		$m$ & Dimension of the input time series  \\
		$l$ & Length of prediction  \\
		$D_{in}$ & Dimension of a layer input  \\
		$D_{out}$ & Dimension of a layer output  \\
		$D_{k}$ & Dimension of queries and keys \\
		$D_{v}$ & Dimension of values \\
		$H$ & Number of heads  \\
		$h$ & Index of head  \\
		$p$ & Order of Monarch matrix \\
		$q$ & Index of time step in queries \\
		$k$ & Index of time step in keys \\
		\bottomrule
	\end{tabular}
	\label{Tab:symbols}
\end{table}

\section{Methodology}\label{sec:model_architecture}
\begin{figure}[ht]
	\centering
	\includegraphics[width=1.0\linewidth]{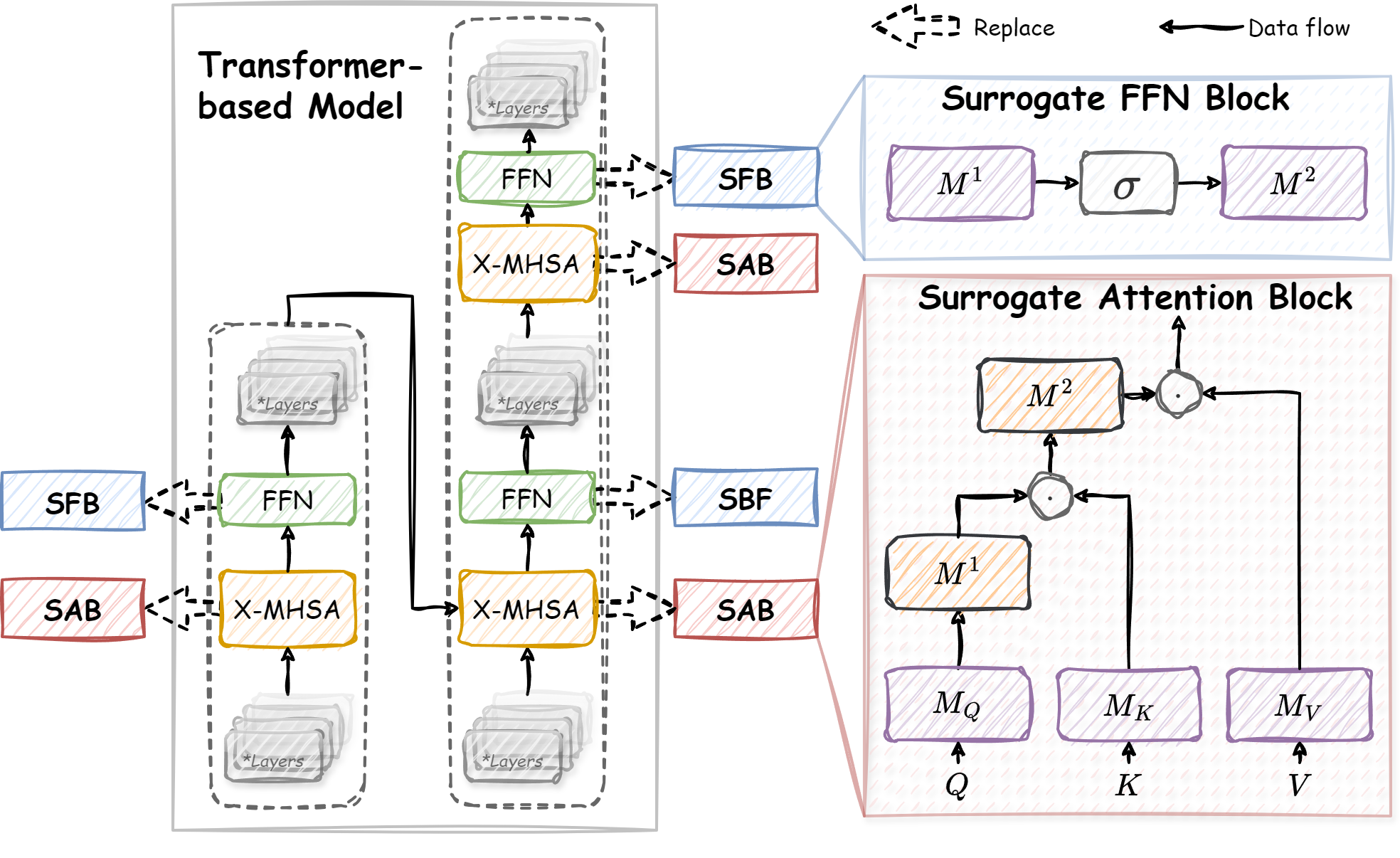}
	\caption{Overview of the proposed enhancement process for a Transformer-based model. For ease of presentation, here only one X-MHSA layer and one FNN layer are illustrated in the Transformer-based model, neglecting the encoder-decoder architecture, multi-layer stacking and other layers that were not modified.}
	\label{fig:overview}
\end{figure}

Our approach focuses on systematically reducing the computational complexity of key neural network layers without compromising model performance. 
Figure \ref{fig:overview} provides a comprehensive visualization of our substitution framework, it is universally applicable, capable of being integrated with various Transformer-based models.
The core innovation lies in a strategic substitution methodology that replaces standard matrix computations with more efficient structured matrix operations. Specifically, our framework targets two critical components of the Transformer architecture:

1. Attention Mechanism Layer: We replace the computationally expensive multi-head self-attention calculations with a novel Surrogate Attention Block. This block maintains the essential information processing capabilities while significantly reducing computational overhead.

2. Feed-Forward Network (FFN) Layer: A Surrogate FFN Block is designed to optimize the linear transformation processes. By introducing carefully crafted structured matrices, we can dramatically reduce the computational complexity of this layer.

\subsection{Surrogate Attention Block}\label{subsec:SAB}
The Surrogate Attention Block consists of two main steps: linear projection substitution and attention  substitution.

\subsubsection{Linear Projection Substitution}
Monarch matrices have been demonstrated to capture Hadamard transforms\cite{Dao2022icml}:
\begin{Remark}
	\label{remark: had=M}
	Let $H_n$ be the Hadamard Transform of size $n$. Then, $H_n \in \mathcal{M}$.
\end{Remark}

\begin{Proposition}
	\label{prop: Linear=M}
	Let $W$ be a weight matrix; a linear projection $LinearProj(\mathbf{X}) = \mathbf{X}W$ is equivalent to a structured linear projection $StructuredLinearProj(\mathbf{X}) = \mathbf{X}\mathbf{M}$, where $\mathbf{M}$ is a structured matrix.
\end{Proposition}

\begin{Proof}
	Proof follows from Remark \ref{remark: had=M}.
\end{Proof}

In accordance with Proposition \ref{prop: Linear=M}, the linear projection in MHSA can be reformulated as:
\begin{equation}
	\label{Eq:projection-m2}
	\begin{aligned}
		Q^{(h)} = \mathbf{X} {\mathbf{M}_Q}^{(h)}, \	K^{(h)} = \mathbf{X} {\mathbf{M}_K}^{(h)}, \ V^{(h)} = \mathbf{X} {\mathbf{M}_V}^{(h)}
	\end{aligned}
\end{equation}
where ${\mathbf{M}^*}^{(h)}$ denotes the Structured matrix on $h$-th head.

\subsubsection{Attention Substitution}
In order to reduce the computational consumption in the attention mechanism, take inspiration from articles utilizing gated long convolution to replace the attention mechanism \cite{Poli2023icml, Fu2023iclr, Fu2023icml, Fu2023corr}. Based on the FFT convolution theorem\footnote{$\mathbf{K}*\mathbf{X} = FFT^{-1}\left( FFT\left( \mathbf{X} \right) * FFT\left( \mathbf{K}\right) \right)$}, we propose substituting X-MHSA with:
\begin{equation}
	\label{Eq:monarch-attn}
	\begin{aligned}
		&SA^{(h)}(\mathbf{X}) := \mathbf{M}^2\left( \mathbf{M}^1Q^{(h)} \odot K^{(h)}\right) \odot V^{(h)}\\
		&MHSA(\mathbf{X}) := \sum_{h \in [H]} SA^{(h)}(\mathbf{X}) W_{out}^{(h)} 
	\end{aligned}
\end{equation}
where $\mathbf{M}^1, \mathbf{M}^2 \in \mathbb{R}^{N\times N}$ are Structured matrices represent $FFT$ and $FFT^{-1}$, $\odot$ denotes element-wise multiplication and $W_{out}^{(h)}$ denotes the output projection matrix. Here, the key matrix $K$ is treated as the convolutional kernel. 

Since the attention mechanism and convolution are not exactly equivalent, to demonstrate the validity of this substitution, the question must be answered: \textit{under what conditions can an MHSA layer be replaced by a (several) convolutional layer(s) in time series tasks?} 

Under observations of patterns in the self-attention scoring matrix of LSTF, they can be categorized into four distinct types\cite{Wang2024apin}: diagonal, vertical, block and heterogeneous.

Within the context of the diagonal pattern, self-attention aggregates local information for each query to capture seasonal patterns within the region centered on itself, reflecting element-level local dependencies. When all heads in an MHSA layer exhibit a diagonal pattern, it is termed a diagonal MHSA layer. On the $h$-th head of a diagonal MHSA layer, the scoring matrices can be expressed as:
\begin{equation}
	\label{Eq:diagonal A}
	\begin{aligned}
		A^{(h)}_{q,k} = \left\lbrace 
		\begin{array}{ll}
			f^{(q,h)}(q-k) & \left( q-k\right)\in \Delta\\
			0 & otherwise
		\end{array}\right. 
	\end{aligned}
\end{equation}
where $A^{(h)}_{q,k}$ denotes the element at position $(q, k)$ in the scoring matrix, $\Delta = \{-\lfloor \lambda/2 \rfloor, \cdots,\lfloor \lambda/2 \rfloor\}$ contains all the corresponding shifts in the diagonal local region with size $\lambda$ and $f^{(q,h)}$ is a set of functions: $f^{(q,h)}: \Delta \rightarrow (0,1]$. For fixed $q$ and $h$, $\sum_{\delta\in \Delta} f^{(q,h)}(\delta)=1$.

\begin{Theorem}
	\label{Theo: diagonal attn=conv}
	Suppose $\exists h\in[H],\ \forall q_1, q_2\in[N], \ f^{(q_1,h)} = f^{(q_2,h)}$, a diagonal MHSA layer with the diagonal local region size$\lambda$, the heads number $H$, the head dimension $D$ and output dimension $D_{out}$ is equivalent to a \textbf{sum of convolutional layer} of kernel size $\lambda$ and $D_{out}$ output channels, i.e., $MHSA^D(\mathbf{X}) = \sum_{h\in [H]}Conv^{(h)}(\mathbf{X})$.
\end{Theorem}

Under the vertical pattern, self-attention maps the specific local regions to each query to learn the relationships between different local regions, which reflect the region-level local dependencies. When all heads in an MHSA layer exhibit a distinct vertical pattern, it is termed a vertical MHSA layer. On the $h$-th head of a vertical MHSA layer, the scoring matrices can be expressed as:
\begin{equation}
	\label{Eq:vertical A}
	\begin{aligned}
		A_{q,k}^{(h)} =
		\left\{\begin{array}{ll}
			g^{(q,h)}(k)   & k\in \widetilde{K}^{(h)}
			\\ 0 &  otherwise
		\end{array}\right.
	\end{aligned}
\end{equation}
where $A^{(h)}_{q,k}$ denotes the element at position $(q, k)$ in the scoring matrix, $\widetilde{K}^{(h)} = \left\lbrace \widetilde{k}_1^{(h)}, \cdots, \widetilde{k}_\lambda^{(h)}\right\rbrace $ contains all $\lambda$ significant columns and $g^{(q,h)}$ is a set of functions: $g^{(q,h)}: \widetilde{K}^{(h)} \rightarrow (0,1] $. For fixed $q$ and $h$, $\sum_{k\in \widetilde{K}^{(h)}} g^{(q,h)}(k)=1$.

\begin{Theorem}
	\label{Theo: vertical attn=conv}
	Suppose $\exists h\in[H],\ \forall q_1, q_2\in[N], \ g^{(q_1,h)} = g^{(q_2,h)}$, a vertical MHSA layer with the significant colums number $\lambda$, the heads number $H$, the head dimension $D$ and output dimension $D_{out}$ is equivalent to a \textbf{sum of convolutional layer} of kernel size $\lambda$ and $D_{out}$ output channels, i.e., $MHSA^V(\mathbf{X}) = \sum_{h\in [H]}Conv^{(h)}(\mathbf{X})$.
\end{Theorem}

Theorems \ref{Theo: diagonal attn=conv} and \ref{Theo: vertical attn=conv} are proved by incorporating Eq. (\ref{Eq:diagonal A}) and Eq. (\ref{Eq:vertical A}) into the expression for MHSA (Eq. (\ref{Eq:MHSA-X})). By crafting a bijective mapping and leveraging the law of the union of matrix multiplications, each head of a diagonal or vertical MHSA layer can be represented as a convolution expression. Please refer to \ref{appendix:diagonal proof} and \ref{appendix:vertical proof} for a comprehensive elucidation of the proof procedure.

Under the block and heterogeneous pattern which appear less frequently in LSTF\cite{Wang2024apin}, it can be considered that each block in the attention map is a convolutional kernel (where a heterogeneous pattern is regarded as one block). Since the block size is large, it is challenging to regard this as meeting the requirement of capturing local features in the convolution. However, it aligns with the concept of a long convolution with a massive kernel. Therefore, the block and the heterogeneous MHSA layer can be equivalent to a long convolution. Nevertheless, this equivalence is not strict, as it does not meet the requirement of weight sharing in a long convolution.

\subsection{Surrogate FFN Block}\label{subsec:ffn}
The Transformer-based architecture stipulated the inclusion of a feed-forward neural network (FFN) layer following each self-attention layer, a configuration retained by many X-formers. Following Proposition \ref{prop: Linear=M}, the dense weight matrices can simply swapped out in the FFN block with structured matrices\cite{Fu2023corr}:
\begin{equation}
	\label{Eq:ffn-m2}
	\begin{aligned}
		\mathbf{Y}= \sigma(\mathbf{X} \mathbf{M}^1)\mathbf{M}^2
	\end{aligned}
\end{equation}
where $\sigma$ is an optional point-wise non-linearity (e.g. ReLU), $\mathbf{M}^1,\mathbf{M}^2 \in \mathbb{R}^{D\times D}$ are Structured matrices. The leftmost side of Figure \ref{fig:overview} illustrates the replacement of FFNs.

\section{Theoretical analysis}\label{sec:theoretical_analysis}
\subsection{Expressiveness}\label{subsec:expresivity}
This section demonstrates the expressiveness of the Surrogate Attention Block on time series tasks by describing its parameterization for solving time series forecasting tasks.

Consider a simple time series with 4 time steps : $\mathbf{X} = \{x_0, x_1, x_2, x_3\}$. If this time series exhibits \textbf{short-term dependence}, it indicates a strong relationship between the value at one time step and the value at the previous time step. Taking $x_1$ as an example, its output $y_1$ must be related to $x_0$. Recall Eq.(\ref{Eq:monarch-attn}) and $\mathbf{M}^1 = \mathbf{P}\mathbf{L}^1\mathbf{P}\mathbf{R}^1\mathbf{P}, \mathbf{M}^2 = \mathbf{P}\mathbf{L}^2\mathbf{P}\mathbf{R}^2\mathbf{P}$, the following exist:
\begin{equation}
	\begin{aligned}
		y_1 = v_1( \mathbf{L}^2_{2,2}\left( \mathbf{R}^2_{1,0}a_0+\mathbf{R}^2_{1,1}a_2\right) +
		\mathbf{L}^2_{2,3}\left( \mathbf{R}^2_{3,2}a_1 + \mathbf{R}^2_{3,3}a_3\right) ) 
	\end{aligned}
	\nonumber
\end{equation}
where the subscripts of $\mathbf{L}$ and $\mathbf{L}$ denote the elements in the corresponding positions in the matrix and $a_*$ denotes the vector of scoring matrix per time step, computed by the following equation:
\begin{equation}
	\begin{aligned}
		a_0 = k_0( \mathbf{L}^1_{0,0}\left( \mathbf{R}^1_{0,0}q_0+\mathbf{R}^1_{0,1}q_2\right) +
		\mathbf{L}^1_{0,1}\left( \mathbf{R}^1_{2,2}q_1 + \mathbf{R}^1_{2,3}q_3\right) ) 
	\end{aligned}
	\nonumber
\end{equation}
\begin{equation}
	\begin{aligned}
		a_1 = k_1( \mathbf{L}^1_{2,2}\left( \mathbf{R}^1_{1,0}q_0+\mathbf{R}^1_{1,1}q_2\right) +
		\mathbf{L}^1_{2,3}\left( \mathbf{R}^1_{3,2}q_1 + \mathbf{R}^1_{3,3}q_3\right) ) 
	\end{aligned}
	\nonumber
\end{equation}
\begin{equation}
	\begin{aligned}
		a_2 = k_2( \mathbf{L}^1_{1,0}\left( \mathbf{R}^1_{0,0}q_0+\mathbf{R}^1_{0,1}q_2\right) +
		\mathbf{L}^1_{1,1}\left( \mathbf{R}^1_{2,2}q_1 + \mathbf{R}^1_{2,3}q_3\right) ) 
	\end{aligned}
	\nonumber
\end{equation}
\begin{equation}
	\begin{aligned}
		a_3 = k_3( \mathbf{L}^1_{3,2}\left( \mathbf{R}^1_{1,0}q_0+\mathbf{R}^1_{1,1}q_2\right) +
		\mathbf{L}^1_{3,3}\left( \mathbf{R}^1_{3,2}q_1 + \mathbf{R}^1_{3,3}q_3\right) ) 
	\end{aligned}
	\nonumber
\end{equation}
where $q_*$ denotes the vector of query matrix per time step

For simplistic considerations, we ignore projections of queries, keys and values. To express the intuition, we set $\mathbf{L}^*$ and $\mathbf{R}^*$ to a very simple case, i.e.,
\begin{equation}
	\begin{aligned}
		\mathbf{L}^1 = \begin{bmatrix}
			1 & 0 &  & \\
			0 & 0 &  & \\
			&  & 0 & 0\\
			&  & 0 & 0\\
		\end{bmatrix}, 
		\mathbf{R}^1 = \begin{bmatrix}
			1 & 0 &  & \\
			0 & 0 &  & \\
			&  & 0 & 0\\
			&  & 0 & 0\\
		\end{bmatrix}, 
		\mathbf{L}^2 = \begin{bmatrix}
			0 & 0 &  & \\
			0 & 0 &  & \\
			&  & 1 & 0\\
			&  & 0 & 0\\
		\end{bmatrix}, 
		\mathbf{R}^2 = \begin{bmatrix}
			0 & 0 &  & \\
			1 & 0 &  & \\
			&  & 0 & 0\\
			&  & 0 & 0\\
		\end{bmatrix}
	\end{aligned}
	\nonumber
\end{equation}

The design of these matrices is deliberate. In $\mathbf{L}^1$ and $\mathbf{R}^1$, the $(0,0)$ and $(1,1)$ entries are set to create a connection between consecutive time steps. This allows information from $x_0$ to influence $x_1$ directly. In $\mathbf{L}^2$ and $\mathbf{R}^2$, the matrix configuration ensures that $y_1$ depends on previous time steps. Specifically, it creates a path for $x_0$ to influence $y_1$.

With these matrices, we derive $y_1=x_0^2 x_1$, which establishes a strong correlation between $y_1$ and $x_0$. This demonstrates that the Surrogate Attention Block can effectively learn and capture short-term dependencies in time series data.  

\textbf{Long-term dependence} means that $\mathbf{X}$ will exhibit periodicity. For example, if the period of $\mathbf{X}$ is assumed to be $2$, then $x_2$ will be strongly correlated with $x_0$. Recall Eq.(\ref{Eq:monarch-attn}) and $\mathbf{M}^1 = \mathbf{P}\mathbf{L}^1\mathbf{P}\mathbf{R}^1\mathbf{P}, \mathbf{M}^2 = \mathbf{P}\mathbf{L}^2\mathbf{P}\mathbf{R}^2\mathbf{P}$, the following relations exist:
\begin{equation}
	\begin{aligned}
		y_2 = v_2( \mathbf{L}^2_{1,0}\left( \mathbf{R}^2_{0,0}a_0+\mathbf{R}^2_{0,1}a_2\right) +
		\mathbf{L}^2_{1,1}\left( \mathbf{R}^2_{2,2}a_1 + \mathbf{R}^2_{2,3}a_3\right) ) 
	\end{aligned}
\end{equation}
where $a_0, a_1,  a_2, a_3$ are the same as those shown above.

Paralleling our approach in demonstrating short-term dependencies, we ignore projections and carefully design the following matrices:
\begin{equation}
	\begin{aligned}
		\mathbf{L}^1 = \begin{bmatrix}
			1 & 0 &  & \\
			0 & 0 &  & \\
			&  & 0 & 0\\
			&  & 0 & 0\\
		\end{bmatrix}, 
		\mathbf{R}^1 = \begin{bmatrix}
			1 & 0 &  & \\
			0 & 0 &  & \\
			&  & 0 & 0\\
			&  & 0 & 0\\
		\end{bmatrix}, 
		\mathbf{L}^2 = \begin{bmatrix}
			0 & 0 &  & \\
			1 & 0 &  & \\
			&  & 0 & 0\\
			&  & 0 & 0\\
		\end{bmatrix}, 
		\mathbf{R}^2 = \begin{bmatrix}
			1 & 0 &  & \\
			0 & 0 &  & \\
			&  & 0 & 0\\
			&  & 0 & 0\\
		\end{bmatrix}
	\end{aligned}
	\nonumber
\end{equation}

Through this carefully constructed configuration, we derive $y_2 = x_2 x_0^2$, which establishes a strong correlation between time steps 2 and 0. This result substantiates a critical capability of the Surrogate Attention Block: the ability to learn and capture long-term dependencies across distant time steps.

The strategic placement of non-zero entries in these projection matrices enables information to traverse across non-consecutive time steps, demonstrating the model's potential to extract meaningful long-range temporal relationships. Our theoretical analysis reveals how the Surrogate Attention Block can transcend the limitations of traditional models that struggle to maintain context over extended temporal distances.

We further extend this demonstration by examining the general case with an arbitrary time step $N$ in Appendix \ref{appendix:expresivity}, providing a comprehensive validation of the model's long-term dependency learning capabilities.

In addition, we also attempt to empirically demonstrate that our proposed method can adapt to long-term and short-term time series. A long-term datasets are generated using a sine function with a period of 96 and short-term datasets using a sine function with a period of 7. Subsequently, we conducted experiments on these two artificial datasets using the improved Transformer model (see Table \ref{tab:expresivity}). The prediction errors for each task were close to zero, empirically demonstrating that our method effectively captures long-term and short-term dependencies in time series data.

\begin{table}[]
	\centering
	\caption{The experimental results of validating the replacement method on the artificial datasets.}
	\begin{tabular}{c|c|cc}
		\toprule  
		\multicolumn{2}{c}{Metric} & MSE& MAE\\
		\midrule
		\multirow{4}{*}{Long}
		& 96  & 0.0046 & 0.0619\\
		& 192  & 0.0063 & 0.0723 \\
		& 336  & 0.0079 & 0.0779\\
		& 720  & 0.0107 & 0.0907\\
		\midrule
		\multirow{4}{*}{Short}
		& 96  & 0.0094 & 0.0882\\
		& 192  & 0.0097 & 0.0901 \\
		& 336  & 0.0132 & 0.1049 \\
		& 720  & 0.0106 & 0.0912\\
		\bottomrule
	\end{tabular}
	\label{tab:expresivity}
\end{table}

\subsection{Complexity}
Since this paper mainly utilizes the Monarch matrix, here we first show that an order-$2$ Monarch matrix $\mathbf{M} = \mathbf{P}\mathbf{L}\mathbf{P}^{\mathsf{T}}\mathbf{R}\mathbf{P}\in \mathbb{R}^{n\times n}$ is described by $2n^{3/2}$ parameters: both $\mathbf{L},\mathbf{R}$ have $n^{1/2}$ dense blocks of size $n^{1/2} \times n^{1/2}$, each with a total number of parameters of $n^{3/2}$. The permutation $\mathbf{P}$ is fixed, so no parameters are added. In order to multiply by $\mathbf{M}$, permute, multiply by the block diagonal matrix $\mathbf{R}$, permute, multiply by the block diagonal matrix $\mathbf{L}$ and finally permute are needed (when right-multiplying by $\mathbf{M}$, this order is reversed). When these five steps are effectively executed, the total time complexity is $O(n^{3/2})$.

Table \ref{tab:complexity} shows the summary of complexity variations. Our proposed replacement method will decrease the time complexity from $O(N^2D+N D^2)$ to $O(N^{3/2}D+N D^{3/2})$ and the space complexity from $O(D^2)$ to $O(D^{3/2}+N^{3/2})$. The following subsections provide detailed explanations of the complexity changes in three neural networks.

\begin{table}[!htbp]
	\centering
	\caption{Summary of complexity variations.}
	\resizebox{0.7\linewidth}{!}{
		\begin{tabular}{ccc}
			\toprule 
			Layer                & Time Complex & Space Complex\\
			\midrule
			LP    & $O(N D^2)\rightarrow O(N D^{\frac{3}{2}})$  & $O(D^2)\rightarrow O(D^{\frac{3}{2}})$  \\
			Attention & $O(N^2D)\rightarrow O(N^{\frac{3}{2}}D)$    &   $0\rightarrow O(N^{\frac{3}{2}})$      \\
			FFN &  $O(N D^2)\rightarrow O(N D^{\frac{3}{2}})$ &  $O(D^2)\rightarrow O(D^{\frac{3}{2}})$\\
			\midrule
			Total &  $O(N^2D+N D^2)\rightarrow O(N^{\frac{3}{2}}D+N D^{\frac{3}{2}})$ & $O(D^2)\rightarrow O(D^{\frac{3}{2}}+N^{\frac{3}{2}})$\\
			\bottomrule
		\end{tabular}
	}
	\label{tab:complexity}
\end{table}

\subsubsection{Linear Projection (LP)}
In a linear projection layer, the input is multiplied by a weight matrix, its time complexity is $O(N D^2)$, space complexity is only related to the size of the weight matrix, i.e. $O(D^2)$. After substituting the weight matrix by Monarch matrix, the time complexity of linear projection is reduced to $O(N D^{3/2})$ and the space complexity is reduced to $O(D^{3/2})$.

\subsubsection{Attention}
The attention layer consists of three steps: (1) \textbf{scoring matrix calculation}: Multiplication of matrices of size $N\times D$ and $D\times N$, with a complexity of $O(N^2D)$. (2) \textbf{Softmax}: Softmax calculation for each row of the scoring matrix, with a complexity of $O(N)$ for one softmax calculation, thus the complexity for $N$ rows is $O(N^2)$. (3) \textbf{Weighted sum}: Multiplication of matrices of size $N\times D$ and $D\times N$, with a complexity of $O(N^2D)$. In total, the time complexity is $O(N^2D)+O(N^2)+O(N^2D)=O(N^2D)$. The surrogate attention block includes two matrix dot products and two Monarch matrix multiplications, with a total time complexity of $2\times O(ND)+2\times O(N^{3/2}D) = O(N^{3/2}D)$. The original attention layer does not occupy any other parameter space except for input data, while the surrogate attention block occupies the space of two Monarch matrices, i.e., $O(N^{3/2})$.

\subsubsection{FFN}
The FFN layer consists of two linear layers and an activation function, with a total time complexity of $O(N D^2)$ and space complexity of $O(D^2)$. Similar to linear layers, the surrogate FFN block reduces them to $O(N D^{3/2})$ and $O(D^{3/2})$.

\subsection{Trainability}
In order to prove that Surrogate Attention Block is capable of being trained, it needs to be shown that it is a linear time-invariant (LTI) system, which can be described by the following equations:
\begin{equation}
	\begin{aligned}
		x_{t+1} &= \mathbf{A}x_t + \mathbf{B}u_{t+1} \\
		y_{t+1} &= \mathbf{C}x_{t+1} + \mathbf{D}u_{t+1}
	\end{aligned}
\end{equation}
where $\mathbf{A}$, $\mathbf{B}$, $\mathbf{C}$ and $\mathbf{D}$ are time-invariant matrices representing the system dynamics, $x_t$ is the state vector at time $t$,  $u_{t+1}$ is the input vector at time $t+1$, $y_{t+1}$ is the output vector at time $t+1$.

The relationship between Surrogate Attention Block and LTI systems is demonstrated in \ref{appendix:LTI}.

\section{Experimental evaluations}\label{sec:experiments}
\subsection{Experimental Setup}

\begin{table*}[]
	\centering
	
	\resizebox{\linewidth}{!}{
		\begin{tabular}{c|c|c|c}
			\toprule
			Tasks                  & Datasets                              & Metrics                     & Series Length \\ \midrule
			Long-term Forecasting &
			ETT(4 subsets), Electricity,Traffic, Weather, Exchange, ILI &
			MSE, MAE, R2, DTW &
			\begin{tabular}[c]{@{}c@{}}96, 192, 336, 720\\ (ILI: 24, 36, 48, 60)\end{tabular} \\ 
			Short-term Forecasting & M4 (6 subsets)                        & SMAPE, MASE, OWA            & 6$\sim$48     \\ 
			Imputation             & ETT (4 subsets), Electricity, Weather & MSE, MAE, R2             & 96            \\
			Classification         & UEA (10 subsets)                      & Accuracy                    & 29$\sim$1751       \\
			Anomaly Detection      & SMD, MSL, SMAP, SWaT, PSM             & Precision, Recall, F1-Socre & 100\\   
			\bottomrule       
		\end{tabular}
	}
	\caption{Summary of experiment benchmarks.}
	\label{tab:benchmark}
\end{table*}

\textbf{Benchmark}: In general, we refer to the experimental setup in TimesNet\cite{Wu2023iclr}. Two new metrics are added to the long-term forecasting task and the imputation task: R-Square (\textbf{R2}) and Dynamic Time Wrapping (\textbf{DTW}). R2 evaluates the goodness of fit of a model to the data. It quantifies the proportion of the variance in the dependent variable that is explained by the model. R-Square values range from 0 to 1, with higher values indicating that the model captures a larger portion of the variation in the data. It helps assess how well the model represents the underlying data patterns. DTW is a method used for comparing two time series with potentially different lengths and time axes. It determines the optimal alignment of elements in the two series, minimizing their paired distances. Therefore, DTW can be used for measuring the waveform similarity between two time series. Table \ref{tab:benchmark} summarizes the datasets, metrics and series length settings for the five tasks. More details can be found in the \ref{appendix:exp}. 

\textbf{Baseline}: We extensively modify and compare all the widely recognized advanced Transformer-based models, including 
Transformer\cite{Vaswani2017nips}, Informer\cite{Zhou2021AAAI}, Autoformer\cite{Wu2021NIPS}, Crossformer\cite{zhang2023ICLR}, Pyraformer\cite{Liu2022ICLR}, Non-stationary Transformer\cite{Liu2022NEURIPS} (abbreviated as NST), PatchTST\cite{Nie2023iclr}, iTransformer\cite{liu2024ICLR}, PAttn\cite{Tan2024NeurIPS} and TimeXer\cite{wang2024NeurIPS}. 
Details of these advanced Transformer-based models can be found in \ref{appendix:detail}.

\textbf{Setup}: All datasets are divided chronologically into training, validation and test sets in a ratio of 6:2:2. The hyperparameters in these baselines are their reported default settings.


\textbf{Platform}: All the models were trained and tested on two Nvidia RTX A6000 GPUs with 16 GB of RAM, which ensured that all models ran successfully without the limitation of hardware resources.

\subsection{Main result} \label{subsec:main result}

\subsubsection{Long-term Forecasting}
\textbf{Result:} Long-term forecasting plays a crucial role in weather forecasting, traffic and electricity consumption planning. We follow the benchmarks used in Autoformer\cite{Wu2021NIPS}, including ETT\cite{Zhou2021AAAI}, Electricity\cite{ecldata}, Traffic\cite{trafficdata}, Weather\cite{weatherdata}, Exchange Rate\cite{Lai2018sigir} and ILI\cite{ilidata}, across a spectrum of five practical domains. The experimental results are shown in Table \ref{tab:full_forecasting_results}. 

\begin{table}[h]
	\centering
	\caption{Full results for the long-term forecasting task. The table data shows the improvement in performance percentage after our method is applied to the original model. For brevity, we omit the percent signs (\%) from the data. Positive numbers indicate performance improvement, while negative numbers indicate the opposite, regardless of whether the evaluation metric is higher-the-better or lower-the-better. The \emph{Count} represents the quantity of tasks in a model where improvements are positive. - indicates the result cannot be obtained due to memory overflow.}
	\resizebox{\textwidth}{!}{
		\begin{threeparttable}
			\begin{small}
				\renewcommand{\multirowsetup}{\centering}
				\setlength{\tabcolsep}{1pt}
				\begin{tabular}{cc|cccc|cccc|cccc|cccc|cccc|cccc|cccc|cccc|cccc|cccc|cccc|}
					\toprule
					\multicolumn{1}{c|}{} &       & \multicolumn{4}{c|}{Trans.}    & \multicolumn{4}{c|}{Log.}      & \multicolumn{4}{c|}{In.}       & \multicolumn{4}{c|}{Auto.}     & \multicolumn{4}{c|}{Cross.}    & \multicolumn{4}{c|}{Pyra.}     & \multicolumn{4}{c|}{Station.}  & \multicolumn{4}{c|}{Patch.}    & \multicolumn{4}{c|}{iTrans.}   & \multicolumn{4}{c|}{PAttn}     & \multicolumn{4}{c}{TimeXer} \\
					\midrule
					\multicolumn{1}{c|}{\multirow{4}[2]{*}{\begin{sideways}ETTh1\end{sideways}}} & 96    & 11.1 & 9.4 & 23.1 & 15.4 & 15.4 & 12.5 & 33.6 & 19.2 & 32.8 & 19.3 & 113.4 & 14.3 & 12.3 & 7.9 & 9.2 & 7.8 & 9.1 & 6.2 & 5.6 & 5.2 & 23.7 & 15.0 & 60.5 & 10.1 & -7.2 & -4.3 & -6.5 & -3.7 & -1.1 & 0.8 & -0.6 & 0.3 & 2.5 & 2.0 & 1.4 & 1.7 & 3.0 & 2.0 & 1.7 & 1.8 & 0.3 & 1.0 & 0.2 & \multicolumn{1}{c}{1.0} \\
					\multicolumn{1}{c|}{} & 192   & 11.1 & 9.9 & 31.9 & 12.3 & 24.3 & 18.1 & 178.7 & 19.1 & 27.1 & 15.2 & 239.5 & 8.1 & 6.9 & 3.8 & 5.8 & 4.4 & -2.3 & -2.2 & -1.9 & -1.7 & 14.3 & 7.4 & 50.1 & 0.2 & -3.9 & -3.7 & -4.9 & -7.8 & -2.8 & 0.6 & -1.7 & 0.0 & 0.2 & 0.8 & 0.1 & 0.5 & 3.0 & 2.7 & 2.0 & 2.0 & 1.0 & 1.2 & 0.6 & \multicolumn{1}{c}{0.9} \\
					\multicolumn{1}{c|}{} & 336   & 9.0 & 10.0 & 931.1 & 5.4 & 6.2 & 1.6 & 94.8 & 0.9 & 22.2 & 11.8 & 710.0 & 0.2 & -3.1 & -1.4 & -2.8 & -0.4 & 12.7 & 9.2 & 29.9 & 13.9 & 23.3 & 17.1 & 332.9 & 7.0 & 20.5 & 14.7 & 47.8 & 4.8 & -3.1 & 0.4 & -2.2 & -0.2 & -0.5 & 0.0 & -0.4 & 0.1 & 6.0 & 5.0 & 5.1 & 3.6 & 4.0 & 0.2 & 3.2 & \multicolumn{1}{c}{0.4} \\
					\multicolumn{1}{c|}{} & 720   & -7.7 & -8.9 & -60.0 & -1.0 & 2.1 & 2.9 & 1056.7 & 0.4 & 14.8 & 8.2 & 172.4 & 0.0 & -2.4 & 0.3 & -2.1 & 1.4 & -13.3 & -9.6 & -24.6 & -3.9 & 1.3 & 0.5 & 12.1 & -2.1 & 7.4 & 4.6 & 13.3 & -1.9 & 7.2 & 4.3 & 6.7 & 1.9 & 3.3 & 2.0 & 2.8 & 1.3 & 16.5 & 9.9 & 17.7 & 6.4 & 4.0 & 2.5 & 3.4 & \multicolumn{1}{c}{1.5} \\
					\midrule
					\multicolumn{1}{c|}{\multirow{4}[2]{*}{\begin{sideways}ETTh2\end{sideways}}} & 96    & 41.1 & 23.8 & 174.6 & 31.0 & 46.9 & 26.2 & 137.0 & 31.5 & 44.1 & 24.9 & 71.7 & 22.6 & -20.1 & -11.2 & -6.3 & -3.9 & 0.1 & 4.0 & 0.1 & -6.7 & -53.0 & -22.4 & -171.7 & -37.9 & -3.1 & 0.3 & -1.1 & 1.4 & -2.9 & -1.7 & -0.7 & -1.3 & 3.7 & 1.7 & 0.9 & 1.6 & 0.6 & 0.9 & 0.1 & 1.0 & -0.7 & -0.5 & -0.2 & \multicolumn{1}{c}{-0.4} \\
					\multicolumn{1}{c|}{} & 192   & 64.0 & 40.4 & 81.9 & 44.1 & 62.5 & 40.2 & 83.4 & 36.5 & 10.3 & 3.3 & 14.1 & -8.4 & 8.8 & 6.2 & 3.7 & 2.4 & 66.3 & 45.3 & 128.6 & 39.7 & 40.6 & 27.0 & 57.4 & -9.3 & 7.5 & 3.5 & -17.6 & 4.1 & 0.9 & 2.3 & 0.3 & 0.9 & 4.6 & 2.0 & 1.6 & 1.0 & 2.2 & 0.2 & 0.7 & -0.1 & 0.6 & -0.6 & 0.2 & \multicolumn{1}{c}{0.5} \\
					\multicolumn{1}{c|}{} & 336   & 37.9 & 22.9 & 59.4 & 23.9 & 57.3 & 37.0 & 82.5 & 33.1 & 24.7 & 13.8 & 35.3 & 14.1 & 0.6 & 2.0 & 0.3 & 3.8 & -69.9 & -28.8 & -274.6 & -22.8 & 36.0 & 18.2 & 53.1 & -8.1 & -4.7 & -1.3 & -2.8 & -0.1 & -2.5 & -0.8 & -0.9 & -0.4 & 4.8 & 3.3 & 1.9 & -1.1 & 0.8 & 1.8 & 0.3 & 1.3 & 1.6 & 2.5 & 0.6 & \multicolumn{1}{c}{1.5} \\
					\multicolumn{1}{c|}{} & 720   & 31.3 & 18.0 & 47.9 & 12.3 & 48.7 & 30.1 & 70.8 & 26.6 & 1.8 & 0.3 & 3.1 & -7.9 & 12.4 & 8.4 & 6.7 & 9.3 & -27.8 & -14.0 & -68.4 & -4.0 & 30.1 & 26.0 & 46.4 & -3.2 & -15.2 & -6.1 & -8.8 & -3.9 & 0.2 & 0.3 & 0.1 & 0.4 & 4.7 & 2.8 & 1.9 & -2.1 & 2.2 & 1.6 & 0.8 & 1.1 & 3.5 & 1.5 & 1.3 & \multicolumn{1}{c}{1.9} \\
					\midrule
					\multicolumn{1}{c|}{\multirow{4}[2]{*}{\begin{sideways}ETTm1\end{sideways}}} & 96    & 35.5 & 23.4 & 59.2 & 30.0 & 23.3 & 14.9 & 26.7 & 22.6 & 26.6 & 11.7 & 41.3 & 12.9 & 9.9 & 3.2 & 8.4 & 3.6 & 2.1 & 0.9 & 1.2 & -1.2 & 9.6 & 0.1 & 10.7 & -1.7 & 5.2 & 1.8 & 3.2 & 2.1 & -1.1 & -0.5
				 & -0.5 & -0.9 & 2.6 & 0.9 & 1.1 & -0.6 & 3.7 & 2.0 & 1.6 & 0.4 & -0.7 & 0.1 & -0.3 & \multicolumn{1}{c}{0.6} \\
					\multicolumn{1}{c|}{} & 192   & 40.0 & 27.6 & 115.0 & 30.7 & 33.2 & 22.4 & 70.5 & 23.3 & 32.3 & 20.5 & 81.2 & 20.8 & 17.8 & 8.2 & 20.3 & 7.6 & -9.7 & -11.2 & -6.1 & -11.6 & 14.6 & 3.4 & 23.1 & 3.8 & -15.0 & -8.4 & -11.2 & -6.4 & 0.0 & 1.9 & 0.0 & 1.5 & 1.8 & 0.6 & 0.9 & -0.6 & 0.4 & 0.9 & 0.2 & -0.5 & 2.0 & 1.3 & 1.1 & \multicolumn{1}{c}{1.1} \\
					\multicolumn{1}{c|}{} & 336   & 46.1 & 31.3 & 1811.4 & 34.8 & 44.1 & 29.8 & 1513.6 & 33.3 & 26.0 & 13.7 & 400.8 & 13.2 & 10.6 & 5.0 & 12.5 & 5.9 & -29.0 & -19.3 & -26.8 & -20.3 & 15.3 & 4.2 & 29.7 & 6.6 & -14.4 & -4.9 & -15.1 & -3.6 & -1.8 & 0.5 & -1.0 & 0.1 & 7.0 & 2.3 & 4.8 & 0.6 & 3.2 & 1.5 & 1.9 & 0.3 & 0.4 & 0.1 & 0.2 & \multicolumn{1}{c}{0.3} \\
					\multicolumn{1}{c|}{} & 720   & 39.5 & 24.5 & 622.9 & 26.6 & 42.4 & 28.7 & 708.9 & 32.2 & 30.9 & 18.0 & 386.3 & 15.4 & 13.0 & 5.9 & 13.9 & 3.5 & -3.4 & -1.1 & -5.8 & -2.1 & 34.1 & 13.8 & 941.1 & 13.3 & 19.6 & 7.6 & 35.3 & 7.8 & -1.7 & 1.6 & -1.2 & 1.3 & 3.0 & 1.2 & 2.5 & -0.3 & -0.1 & -0.4 & -0.1 & -0.2 & 0.9 & -0.1 & 0.7 & \multicolumn{1}{c}{-0.9} \\
					\midrule
					\multicolumn{1}{c|}{\multirow{4}[2]{*}{\begin{sideways}ETTm2\end{sideways}}} & 96    & -65.7 & -44.1 & -24.5 & -40.4 & 5.3 & -4.7 & 2.1 & -2.6 & 5.0 & 2.1 & 1.9 & -10.0 & -24.5 & -13.1 & -4.5 & -8.3 & 2.1 & -1.0 & 0.5 & -1.7 & -52.5 & -34.6 & -14.7 & -38.9 & 17.8 & 8.0 & 3.3 & 6.7 & 1.3 & 0.1 & 0.2 & -1.0 & 3.3 & 2.0 & 0.5 & -0.1 & 0.4 & 0.4 & 0.0 & 0.6 & 1.1 & -0.2 & 0.1 & \multicolumn{1}{c}{-0.2} \\
					\multicolumn{1}{c|}{} & 192   & 16.3 & 0.4 & 26.7 & -3.0 & 42.4 & 24.5 & 91.8 & 19.8 & 15.0 & 5.6 & 13.8 & -0.9 & 5.5 & 2.0 & 1.3 & -0.1 & -6.0 & -8.5 & -2.2 & -6.5 & -5.8 & -8.0 & -5.3 & -7.4 & 11.0 & 3.7 & 6.8 & 0.5 & 0.7 & -0.5 & 0.1 & -0.8 & 2.0 & 1.6 & 0.4 & -0.8 & 1.0 & 1.4 & 0.2 & 0.9 & 1.3 & 1.1 & 0.3 & \multicolumn{1}{c}{-0.1} \\
					\multicolumn{1}{c|}{} & 336   & -21.0 & -6.3 & -98.5 & -9.0 & -13.9 & -7.4 & -63.0 & -15.9 & 30.0 & 13.4 & 595.0 & 7.9 & -4.4 & -1.3 & -1.3 & 0.6 & 25.6 & 12.7 & 72.0 & 10.1 & -13.1 & -16.2 & -64.1 & -12.5 & -8.8 & -14.6 & -5.3 & -7.4 & 1.3 & 0.6 & 0.3 & 1.1 & -4.6 & -1.8 & -1.2 & -0.8 & 0.5 & 0.4 & 0.1 & 1.3 & 1.0 & 1.8 & 0.2 & \multicolumn{1}{c}{-0.8} \\
					\multicolumn{1}{c|}{} & 720   & -26.7 & -14.7 & -56.7 & -18.7 & 9.1 & 4.0 & 17.6 & 6.2 & 18.9 & 10.8 & 32.1 & 10.3 & -11.3 & -7.9 & -4.2 & -4.3 & 43.5 & 21.4 & 65.5 & 13.7 & 36.0 & 18.3 & 66.7 & -11.5 & 33.8 & 17.1 & 21.9 & 13.6 & 1.0 & 0.0 & 0.4 & -1.0 & -8.0 & -3.4 & -2.9 & -1.8 & 5.5 & 3.5 & 2.1 & 5.7 & 2.2 & 1.8 & 0.8 & \multicolumn{1}{c}{-0.3} \\
					\midrule
					\multicolumn{1}{c|}{\multirow{4}[2]{*}{\begin{sideways}EXC\end{sideways}}} & 96    & 42.4 & 19.0 & 22.3 & 16.4 & -5.8 & -7.0 & -3.3 & -0.7 & 24.6 & 14.6 & 30.6 & 13.4 & 13.1 & 5.6 & 1.5 & 10.4 & -24.5 & -17.5 & -4.2 & -19.5 & 7.2 & 4.7 & 2.8 & -4.1 & 18.7 & 5.0 & 1.8 & 8.2 & -1.3 & -0.5 & -0.1 & -3.1 & -8.3 & -4.6 & -0.4 & -12.1 & 3.8 & 1.4 & 0.2 & 2.0 & 0.9 & 1.6 & 0.0 & \multicolumn{1}{c}{-0.9} \\
					\multicolumn{1}{c|}{} & 192   & -14.8 & -11.4 & -20.9 & -0.5 & 32.4 & 11.3 & 51.0 & 11.7 & -4.1 & -3.8 & -7.8 & -6.0 & -0.2 & 1.2 & 0.0 & 2.6 & -138.3 & -53.2 & -52.3 & -30.4 & 4.5 & -1.1 & 8.5 & -7.5 & 12.5 & 4.6 & 2.3 & 9.1 & -4.9 & -2.1 & -0.6 & -9.8 & -1.1 & -0.8 & -0.1 & -2.6 & 5.1 & 2.3 & 0.7 & 5.0 & 1.2 & 0.3 & 0.1 & \multicolumn{1}{c}{3.0} \\
					\multicolumn{1}{c|}{} & 336   & 16.5 & 0.2 & 56.0 & 3.6 & 35.2 & 12.8 & 174.6 & 15.5 & 12.3 & 2.7 & 335.7 & 3.6 & 6.7 & 4.1 & 2.6 & 4.6 & 9.6 & 3.8 & 26.9 & -2.2 & 2.9 & -4.2 & 11.4 & -3.8 & 12.7 & 4.9 & 4.7 & 9.8 & -0.7 & 0.4 & -0.2 & 0.2 & -5.2 & -2.8 & -1.3 & -5.3 & 7.6 & 3.9 & 2.1 & 8.0 & 2.5 & 1.5 & 0.7 & \multicolumn{1}{c}{1.0} \\
					\multicolumn{1}{c|}{} & 720   & 48.6 & 28.8 & 132.6 & 19.8 & 30.3 & 18.9 & 53.5 & 7.4 & 45.7 & 28.7 & 114.9 & 18.3 & -5.2 & -1.8 & -9.9 & 1.1 & 8.4 & 2.2 & 107.5 & -5.5 & 27.2 & 13.2 & 102.0 & -0.9 & 28.6 & 9.3 & 310.3 & 18.6 & 1.0 & 0.2 & 1.1 & 10.7 & 2.7 & 0.5 & 3.2 & 0.5 & 2.2 & 1.2 & 2.6 & -1.7 & 0.2 & 0.0 & 0.2 & \multicolumn{1}{c}{6.3} \\
					\midrule
					\multicolumn{1}{c|}{\multirow{4}[2]{*}{\begin{sideways}ECL\end{sideways}}} & 96    & -7.5 & -3.8 & -2.6 & -3.5 & 7.8 & 4.1 & 3.0 & 3.7 & 1.6 & 0.6 & 0.8 & -1.6 & 2.9 & 2.2 & 0.7 & 0.5 & 7.3 & 6.4 & 1.2 & 4.5 & -6.1 & -3.5 & -2.4 & -2.5 & -13.9 & -10.1 & -2.8 & -6.5 & -6.3 & -3.2 & -1.4 & -2.8 & 25.1 & 16.2 & 6.1 & 13.3 & 2.0 & 0.2 & 7.7 & 6.4 & -1.7 & -2.3 & -0.3 & \multicolumn{1}{c}{-1.0} \\
					\multicolumn{1}{c|}{} & 192   & -5.1 & -5.0 & -1.8 & -2.1 & 4.7 & 3.9 & 1.8 & 2.3 & -3.4 & -2.4 & -1.8 & -2.5 & 6.5 & 4.7 & 1.9 & 3.0 & -4.4 & -2.7 & -0.9 & -1.6 & -6.4 & -3.9 & -2.7 & -3.3 & -10.2 & -7.8 & -2.3 & -4.9 & -5.2 & -3.2 & -1.2 & -2.5 & 20.8 & 13.3 & 5.4 & 11.1 & 5.1 & 3.2 & 8.8 & 6.2 & -1.5 & -1.1 & -0.3 & \multicolumn{1}{c}{-0.9} \\
					\multicolumn{1}{c|}{} & 336   & 0.7 & -1.8 & 0.3 & -0.5 & 2.1 & 2.8 & 0.8 & 1.8 & -1.3 & -0.7 & -0.7 & -1.9 & 20.0 & 10.6 & 7.5 & 6.3 & - & - & - & - & -4.3 & -2.9 & -1.9 & -2.4 & -12.1 & -7.2 & -3.0 & -5.4 & -4.2 & -9.9 & -1.0 & -1.4 & 21.3 & 13.7 & 6.2 & 10.8 & 4.3 & 4.3 & 9.6 & 6.0 & 2.4 & 1.4 & 0.6 & \multicolumn{1}{c}{1.3} \\
					\multicolumn{1}{c|}{} & 720   & -0.4 & -1.3 & -0.2 & -0.7 & 1.2 & 1.7 & 0.5 & 1.0 & 6.8 & 3.4 & 4.8 & -0.9 & 14.6 & 6.3 & 6.4 & 12.0 & - & - & - & - & 6.1 & 4.0 & 2.8 & 4.0 & 84.3 & 61.9 & 811.0 & -273.4 & -2.5 & -1.8 & -0.7 & -2.9 & 8.2 & -3.4 & 3.1 & 3.1 & 3.7 & 4.0 & 9.9 & 5.7 & 3.1 & 2.1 & 1.0 & \multicolumn{1}{c}{1.4} \\
					\midrule
					\multicolumn{1}{c|}{\multirow{4}[2]{*}{\begin{sideways}WTH\end{sideways}}} & 96    & 26.9 & 13.5 & 49.8 & 19.8 & 31.7 & 24.7 & 48.6 & 23.2 & 21.0 & 12.8 & 94.6 & 12.9 & -6.5 & -4.7 & -5.1 & -0.6 & 11.9 & 7.1 & 5.4 & 8.6 & 4.8 & 4.4 & 2.2 & 3.1 & -8.3 & -6.7 & -4.0 & -7.8 & -7.7 & -3.8 & -3.4 & -5.8 & -8.6 & -2.8 & -3.3 & -2.9 & 4.0 & 2.6 & 1.9 & 3.1 & 0.8 & 0.8 & 0.3 & \multicolumn{1}{c}{0.3} \\
					\multicolumn{1}{c|}{} & 192   & 38.1 & 21.3 & 679.8 & 24.6 & 32.4 & 22.2 & 332.0 & 24.0 & 3.2 & 0.7 & 43.7 & -0.8 & -5.5 & -2.1 & -6.8 & -6.7 & 11.2 & 6.5 & 7.8 & 2.2 & 4.5 & 5.4 & 3.1 & 0.2 & 6.1 & 3.8 & 4.7 & 4.0 & -6.2 & -2.4 & -3.9 & -3.8 & -5.6 & -1.8 & -3.2 & -0.7 & 1.0 & 0.4 & 0.7 & 0.6 & 3.2 & 1.3 & 1.9 & \multicolumn{1}{c}{0.7} \\
					\multicolumn{1}{c|}{} & 336   & 13.6 & 6.2 & 76.2 & 9.8 & 37.3 & 25.0 & 184.4 & 28.2 & -54.6 & -29.1 & -1028.1 & -24.6 & -10.6 & -7.3 & -21.4 & -12.8 & -0.8 & -2.1 & -0.7 & -2.0 & 6.4 & 4.1 & 7.0 & 1.7 & 8.1 & 4.3 & 11.7 & 5.2 & -2.2 & -0.7 & -2.1 & -1.9 & -4.4 & -1.7 & -3.8 & -1.0 & 1.6 & 0.5 & 1.6 & 1.6 & 0.3 & 0.1 & 0.2 & \multicolumn{1}{c}{0.9} \\
					\multicolumn{1}{c|}{} & 720   & 7.4 & -0.7 & 26.2 & 1.2 & 46.1 & 27.7 & 111.5 & 28.3 & -33.2 & -16.3 & -71.4 & -7.8 & -28.9 & -20.8 & -70.8 & -17.5 & 1.0 & 0.8 & 1.7 & 1.4 & 15.3 & 8.3 & 46.9 & 3.9 & -5.5 & -3.4 & -12.5 & -4.0 & -1.9 & -0.5 & -3.1 & -0.5 & -2.6 & -0.9 & -4.1 & 0.0 & 0.9 & 0.2 & 1.5 & -0.3 & 0.2 & 0.2 & 0.2 & \multicolumn{1}{c}{-0.5} \\
					\midrule
					\multicolumn{1}{c|}{\multirow{4}[2]{*}{\begin{sideways}Traffic\end{sideways}}} & 96    & -4.1 & -9.8 & -3.2 & -4.1 & 4.3 & 9.6 & 3.7 & 4.2 & -1.7 & -8.3 & -1.7 & -6.2 & 4.4 & -1.6 & 3.6 & -2.4 & 1.2 & 2.8 & 0.7 & 0.8 & -0.6 & -0.9 & -0.3 & 0.8 & 1.7 & 0.8 & 1.3 & 1.2 & 18.3 & 9.9 & 14.4 & 4.1 & 41.5 & 35.3 & 35.4 & 18.1 & 4.5 & 5.1 & 5.6 & 0.8 & 32.1 & 20.3 & 28.8 & \multicolumn{1}{c}{16.0} \\
					\multicolumn{1}{c|}{} & 192   & -3.4 & -9.9 & -2.8 & -5.7 & 3.0 & 8.4 & 2.6 & 5.0 & 0.6 & -4.1 & 0.6 & -4.3 & 6.6 & 7.2 & 5.8 & 0.4 & - & - & - & - & 1.8 & 1.1 & 1.8 & 0.1 & -3.2 & -2.0 & -2.5 & -2.0 & 13.1 & 5.9 & 9.4 & 2.9 & 36.2 & 32.2 & 28.4 & 16.9 & 4.3 & 4.9 & 6.0 & -0.4 & 25.8 & 13.9 & 20.1 & \multicolumn{1}{c}{13.1} \\
					\multicolumn{1}{c|}{} & 336   & 0.9 & -4.3 & 0.7 & -4.2 & -0.3 & 3.9 & -0.2 & 3.4 & 9.3 & 7.7 & 11.8 & -1.6 & 12.7 & 14.8 & 13.1 & 3.3 & - & - & - & - & -2.1 & -5.2 & -1.7 & -1.5 & 6.9 & 9.3 & 5.9 & -8.5 & 10.5 & 4.3 & 7.6 & 2.1 & 13.8 & 6.8 & 9.7 & 4.1 & 4.2 & 4.8 & 8.3 & 0.0 & 24.7 & 13.8 & 18.2 & \multicolumn{1}{c}{11.0} \\
					\multicolumn{1}{c|}{} & 720   & 7.3 & 14.7 & 7.4 & 1.9 & 8.9 & 18.9 & 8.9 & 4.6 & 0.1 & -0.3 & 2.0 & 1.0 & 9.5 & 11.8 & 9.2 & 2.0 & - & - & - & - & - & - & - & - & 18.3 & 16.0 & 23.5 & 1.9 & 8.7 & 2.6 & 7.1 & 1.8 & 13.1 & 5.8 & 10.3 & 4.2 & 4.0 & 7.8 & 8.8 & 94.9 & 73.1 & 29.7 & 59.0 & \multicolumn{1}{c}{59.2} \\
					\midrule
					\multicolumn{1}{c|}{\multirow{4}[2]{*}{\begin{sideways}ILI\end{sideways}}} & 24    & 31.2 & 24.7 & 153.0 & 10.4 & 20.3 & 16.0 & 81.0 & 2.3 & 4.9 & 2.4 & 31.8 & 3.7 & 3.1 & -2.8 & 10.4 & -1.4 & 7.9 & 5.3 & 42.4 & 1.5 & -34.4 & -23.4 & -335.2 & -34.7 & -3.1 & -7.8 & -5.1 & -12.7 & -9.2 & -19.5 & -9.7 & -21.7 & 28.2 & 26.2 & 83.2 & 23.4 & 5.0 & 7.1 & 4.2 & 4.5 & 11.4 & 0.2 & 9.2 & \multicolumn{1}{c}{-1.1} \\
					\multicolumn{1}{c|}{} & 36    & 26.8 & 19.0 & 150.6 & 1.9 & 9.3 & 5.8 & 64.5 & -3.0 & 10.9 & 7.8 & 58.6 & 4.4 & 0.8 & -4.0 & 2.5 & -4.4 & 5.2 & 3.8 & 24.7 & -4.5 & -24.5 & -17.0 & -125.4 & -35.9 & 10.1 & -0.5 & 24.3 & -4.8 & 11.8 & -1.5 & 13.8 & -6.3 & 28.9 & 23.1 & 81.2 & 18.1 & 1.7 & 2.1 & 1.0 & 1.1 & 2.3 & 4.8 & 1.8 & \multicolumn{1}{c}{3.9} \\
					\multicolumn{1}{c|}{} & 48    & 20.6 & 15.3 & 119.7 & 4.5 & 5.0 & 2.3 & 39.0 & -1.8 & 13.6 & 10.5 & 70.5 & 6.2 & -6.8 & -8.0 & -15.7 & -6.8 & 3.1 & 3.0 & 19.9 & -2.1 & -37.6 & -24.0 & -356.4 & -31.7 & 4.7 & -1.0 & 7.5 & -4.7 & 10.1 & -5.0 & 10.6 & -5.7 & 26.7 & 21.5 & 75.4 & 15.8 & 1.2 & 1.0 & 0.7 & 0.4 & 1.7 & -1.2 & 1.2 & \multicolumn{1}{c}{-1.6} \\
					\multicolumn{1}{c|}{} & 60    & 22.1 & 16.4 & 103.4 & 6.5 & 14.9 & 9.9 & 69.0 & 2.9 & 6.7 & 7.0 & 31.3 & 1.6 & -15.4 & -13.4 & -28.4 & -7.2 & 0.3 & 1.3 & 3.2 & -3.4 & -45.9 & -27.6 & -798.3 & -32.4 & -5.5 & -7.2 & -7.1 & -10.9 & 6.0 & -1.7 & 5.3 & -5.7 & 30.0 & 23.7 & 83.3 & 15.5 & 1.0 & 0.7 & 0.7 & 0.2 & 1.0 & -1.4 & 0.8 & \multicolumn{1}{c}{-1.6} \\
					\midrule
					\multicolumn{2}{c|}{Average} & \multicolumn{4}{c|}{\textbf{44.5\%}}   & \multicolumn{4}{c|}{\textbf{49.0\%}}   & \multicolumn{4}{c|}{\textbf{23.6\%}}   & \multicolumn{4}{c|}{\textbf{0.4\%}}    & \multicolumn{4}{c|}{-0.9\%}   & \multicolumn{4}{c|}{-1.2\%}   & \multicolumn{4}{c|}{\textbf{8.6\%}}    & \multicolumn{4}{c|}{-0.1\%}   & \multicolumn{4}{c|}{\textbf{7.4\%}}    & \multicolumn{4}{c|}{\textbf{3.4\%}}    & \multicolumn{4}{c}{\textbf{4.0\%}} \\
					\midrule
					\multicolumn{2}{c|}{$>$0\%Cnt} & \multicolumn{4}{|c|}{\textbf{98}} & \multicolumn{4}{c|}{\textbf{130}} & \multicolumn{4}{c|}{\textbf{108}} & \multicolumn{4}{c|}{\textbf{87}} & \multicolumn{4}{c|}{\textbf{67}} & \multicolumn{4}{c|}{\textbf{77}} & \multicolumn{4}{c|}{\textbf{74}} & \multicolumn{4}{c|}{\textbf{62}} & \multicolumn{4}{c|}{\textbf{98}} & \multicolumn{4}{c|}{\textbf{134}} & \multicolumn{4}{c}{\textbf{115}}\\
					\midrule
					\multicolumn{2}{c|}{$>$-5\%Cnt} & \multicolumn{4}{|c|}{\textbf{120}} & \multicolumn{4}{c|}{\textbf{138}} & \multicolumn{4}{c|}{\textbf{129}} & \multicolumn{4}{c|}{\textbf{113}} & \multicolumn{4}{c|}{\textbf{93}} & \multicolumn{4}{c|}{\textbf{101}} & \multicolumn{4}{c|}{\textbf{105}} & \multicolumn{4}{c|}{\textbf{129}} & \multicolumn{4}{c|}{\textbf{137}} & \multicolumn{4}{c|}{\textbf{144}} & \multicolumn{4}{c}{\textbf{144}}\\
					\bottomrule
				\end{tabular}%
			\end{small}
		\end{threeparttable}
	}
	\label{tab:full_forecasting_results}
\end{table}

Table \ref{tab:paras&flops} presents a detailed comparison between the original models and their optimized counterparts in terms of parameters and FLOPS. The results highlight substantial improvements across various architectures, with the optimized models consistently achieving reductions in both model size and computational cost. On average, the optimizations lead to a 63.6\% decrease in parameters and a 65.5\% reduction in FLOPS, demonstrating the efficiency of the proposed approach. These results indicate that our method significantly enhances scalability and is particularly effective in improving Transformer-based models.

\begin{table}[htb]
	\centering
	\caption{Comparison of model parameters and FLOPS reduction between original and optimized models using our method}
	\resizebox{\linewidth}{!}{
		\begin{tabular}{c|c|c|c|c|c|c|c|c|c|c|c}
			\toprule
			&       & Trans. & In.   & Auto.   & Cross. & Pyra. & NST   & Patch. & iTrans. & PAttn & TimeXer\\
			\midrule
			\multirow{3}[2]{*}{Paras (M)} & Original & 10.54 & 11.33 & 13.68 & 42.06 & 7.61  & 6.37  & 0.22  & 6.90 & 3.75 & 9.10\\
			& Ours  & 2.66  & 4.50  & 6.87  & 31.04 & 2.36  & 1.12  & 0.09  & 1.65 & 1.13 & 4.90\\
			& Improv. & 74.7\% & 60.2\% & 49.8\% & 26.2\% & 69.0\% & 82.4\% & 58.9\% & 76.0\% & 70.0\% & 46.1\%\\
			\midrule
			\multirow{3}[2]{*}{FLOPS (G)} & Original & 1.19  & 1.09  & 1.57  & 1.97  & 0.81  & 0.61  & 0.0025 & 0.53 & 0.27 & 0.60\\
			& Ours  & 0.31  & 0.41  & 0.81  & 1.48  & 0.15  & 0.10  & 0.0010 & 0.09 & 0.05 & 0.22 \\
			& Improv. & 74.1\% & 62.4\% & 48.2\% & 24.6\% & 82.0\% & 83.0\% & 58.6\% & 82.4\% & 81.7\% & 63.9\%\\
			\bottomrule
		\end{tabular}%
	}
	\label{tab:paras&flops}%
\end{table}%

\begin{figure}[htb]
	\centering
	\includegraphics[width=\linewidth]{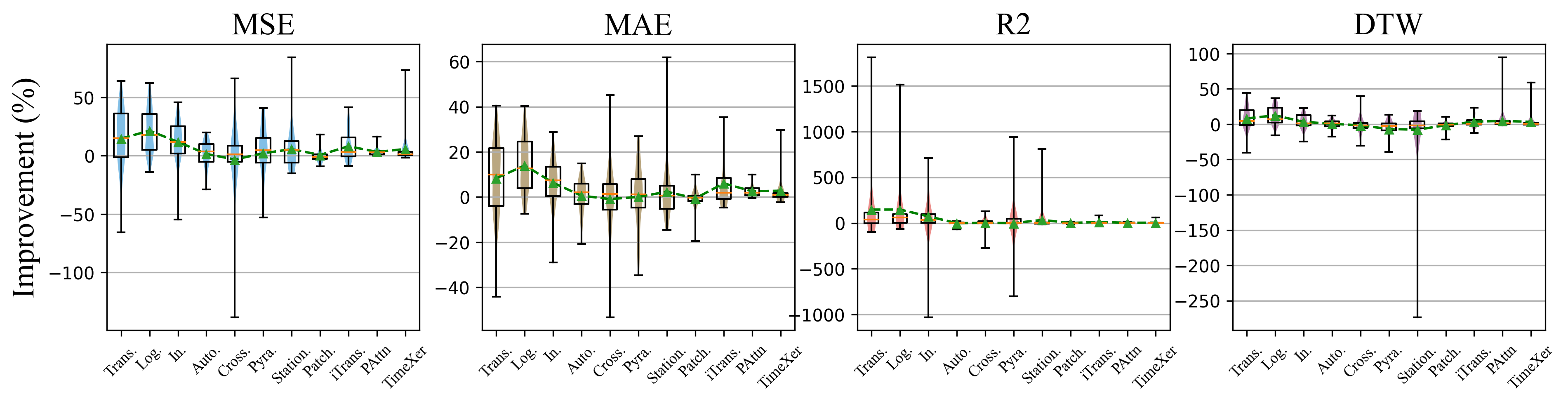}
	\caption{Box plots of statistical results on the distribution of \textbf{long-term forecasts} for each metric. The horizontal axis denotes the models and the vertical axis is the percentage of lift after using the structured matrix. The width of the color blocks in the box indicates the density of the distribution. The green dashed line is the mean and the orange dashed line is the median.}
	\label{fig:long}
\end{figure}

\textbf{Analysis:} For all 11 Transformer-based models, the number of tasks with improvement higher than -5\% exceeds $1/2$ of the total number of tasks (each model has a total of 144 tasks). For Transformer, Informer, Autoformer, Nonstationary Transformer, PatchTST, PAttn and TimeXer, this number is above $2/3$. This generally indicates that the use of structured matrix optimization for Transformer-based models has almost no significant negative impact on performance. The Transformer, Informer, Autoformer, PAttn and TimeXer have shown the best performance, with at least a moderate improvement in tasks being indicated by positive values. In addition, we conducted wilconxon significance test, the original hypothesis is that there is no significant difference in the performance of the model before and after the optimization, according to the experimental results, we calculated that the p-value is 0, which is at the set significance level of 0.05, indicating that the optimized model's performance is significantly better than that of the baseline model. This suggests that structured matrices can not only improve efficiency across a wide range of tasks, but also effectively enhance the predictive capabilities of the models. The results of the improvements on the other models did not perform well and we will analyze the reasons for this in the Section \ref{sec: ana}.
The visualization in Figure \ref{fig:long} displays the distribution of data for each model. It is evident from this that almost all the improvements in Transformer-based models have mean and median values above 0. This once again emphasizes the positive impact of structured matrices on predictive performance.

\subsubsection{Short-term Forecasting}
\textbf{Result:} For short-term forecasting, the M4 datasets\cite{M4team2018dataset} are utilized. These datasets consist of univariate marketing data collected yearly, quarterly and monthly, comprising 100,000 diverse time series gathered at various frequencies.
Table \ref{tab:full_forecasting_results_m4} presents the experimental results of short-term forecasting tasks. 

\begin{table}[h]
	\caption{Full results for the short-term forecasting task in the M4 dataset. Positive numbers indicate performance improvement, while negative numbers indicate the opposite, regardless of whether the evaluation metric is higher-the-better or lower-the-better.}
	\vskip 0.05in
	\centering
	\begin{threeparttable}
		\begin{small}
			\renewcommand{\multirowsetup}{\centering}
			\setlength{\tabcolsep}{0.7pt}
			\begin{tabular}{cc|c|cccccccccc}
				\toprule
				\multicolumn{3}{c}{\multirow{1}{*}{Models}} & 
				\multicolumn{1}{c}{\rotatebox{0}{\scalebox{0.8}{Trans.}}} &
				\multicolumn{1}{c}{\rotatebox{0}{\scalebox{0.8}{In.}}} &
				\multicolumn{1}{c}{\rotatebox{0}{\scalebox{0.8}{Auto.}}} &
				\multicolumn{1}{c}{\rotatebox{0}{\scalebox{0.8}{Cross.}}} &
				\multicolumn{1}{c}{\rotatebox{0}{\scalebox{0.8}{Pyra.}}} &
				\multicolumn{1}{c}{\rotatebox{0}{\scalebox{0.8}{NST}}} & \multicolumn{1}{c}{\rotatebox{0}{\scalebox{0.8}{Patch.}}} & \multicolumn{1}{c}{\rotatebox{0}{\scalebox{0.8}{iTrans.}}} &
				\multicolumn{1}{c}{\rotatebox{0}{\scalebox{0.8}{PAttn}}} &
				\multicolumn{1}{c}{\rotatebox{0}{\scalebox{0.8}{TimeXer}}}
				\\
				\toprule
				\multicolumn{2}{c|}{\multirow{4}{*}{\rotatebox{90}{\scalebox{0.95}{Yearly}}}}
				& SMAPE & 1.48\% & 2.22\% & 5.78\% & 12.23\% & 3.40\% & 13.19\% & -0.46\% & 4.43\% & 5.29\% & -0.27\% \\
				&       & MAPE  & 6.59\% & 1.08\% & 12.55\% & -2.28\% & 10.74\% & 16.79\% & 2.01\% & 13.19\% & 9.74\% & 0.85\% \\
				&       & MASE  & 2.18\% & 2.24\% & 8.24\% & 3.15\% & 8.48\% & 10.31\% & -1.43\% & 5.06\% & 3.77\% & 0.10\% \\
				&       & OWA   & 1.85\% & 2.20\% & 6.90\% & 7.76\% & 8.36\% & 11.79\% & -1.01\% & 5.21\% & 4.54\% & -0.13\% \\

				\midrule
				\multicolumn{2}{c|}{\multirow{4}{*}{\rotatebox{90}{\scalebox{0.95}{Quarterly}}}}
				& SMAPE & 19.36\% & 2.01\% & 6.08\% & 1.33\% & 16.04\% & 16.33\% & -0.74\% & 12.02\% & 1.18\% & -1.23\% \\
				&       & MAPE  & 33.91\% & 2.03\% & 8.68\% & -5.12\% & 18.50\% & 19.07\% & -0.46\% & 14.45\% & 2.25\% & -1.98\% \\
				&       & MASE  & 29.30\% & 3.00\% & 6.84\% & -1.05\% & 19.23\% & 19.16\% & -0.25\% & 13.54\% & 1.25\% & -0.57\% \\
				&       & OWA   & 25.23\% & 2.44\% & 6.43\% & -0.07\% & 17.57\% & 17.75\% & -0.45\% & 12.78\% & 1.22\% & -0.87\% \\

				\midrule
				\multicolumn{2}{c|}{\multirow{4}{*}{\rotatebox{90}{\scalebox{0.95}{Monthly}}}}
				& SMAPE & 3.06\% & 4.87\% & 0.87\% & 0.43\% & 10.39\% & 28.43\% & -0.67\% & 6.83\% & 3.49\% & 1.07\% \\
				&       & MAPE  & 3.45\% & 5.51\% & 0.02\% & -3.73\% & 11.42\% & 29.09\% & -1.90\% & 6.31\% & 3.61\% & 0.22\% \\
				&       & MASE  & -1.75\% & 8.25\% & 2.20\% & -0.57\% & 15.90\% & 38.58\% & -0.21\% & 12.59\% & 7.96\% & 4.25\% \\
				&       & OWA   & -0.22\% & 6.62\% & 1.59\% & -0.26\% & 13.21\% & 33.86\% & -0.45\% & 9.83\% & 5.81\% & 2.63\% \\

				\midrule
				\multicolumn{2}{c|}{\multirow{4}{*}{\rotatebox{90}{\scalebox{0.95}{Others}}}}
				& SMAPE & 0.28\% & 11.14\% & 1.48\% & 42.35\% & 10.90\% & 10.46\% & -8.13\% & 12.50\% & 0.94\% & 1.08\% \\
				&       & MAPE  & 0.20\% & 21.28\% & 5.21\% & 30.87\% & 13.54\% & 9.12\% & -28.77\% & 17.71\% & -19.90\% & 1.72\% \\
				&       & MASE  & 0.33\% & 4.58\% & 2.95\% & 31.42\% & 15.58\% & 15.55\% & 2.40\% & 17.41\% & 7.06\% & 2.53\% \\
				&       & OWA   & 0.31\% & 7.94\% & 2.25\% & 36.91\% & 13.34\% & 13.09\% & -2.81\% & 15.03\% & 3.97\% & 1.81\% \\

				\midrule
				\multirow{4}{*}{\rotatebox{90}{\scalebox{0.95}{Weighted}}}& \multirow{4}{*}{\rotatebox{90}{\scalebox{0.95}{Average}}} 
				& SMAPE & 6.48\% & 3.76\% & 3.19\% & 8.13\% & 9.84\% & 6.50\% & -0.79\% & 7.44\% & 3.45\% & 0.26\% \\
				&       & MAPE  & 11.31\% & 4.14\% & 5.38\% & -1.28\% & 10.46\% & 10.31\% & -1.20\% & 9.35\% & 4.45\% & -0.01\% \\
				&       & MASE  & 5.61\% & 4.40\% & 5.55\% & 10.19\% & 10.77\% & 37.44\% & -0.44\% & 8.88\% & 4.91\% & 1.44\% \\
				&       & OWA   & 5.95\% & 4.04\% & 4.41\% & 9.42\% & 10.26\% & 23.47\% & -0.70\% & 8.16\% & 4.12\% & 0.79\% \\
				
				\bottomrule
			\end{tabular}
		\end{small}
	\end{threeparttable}
	\label{tab:full_forecasting_results_m4}
\end{table}

\begin{figure}[htb]
	\centering
	\includegraphics[width=1.0\linewidth]{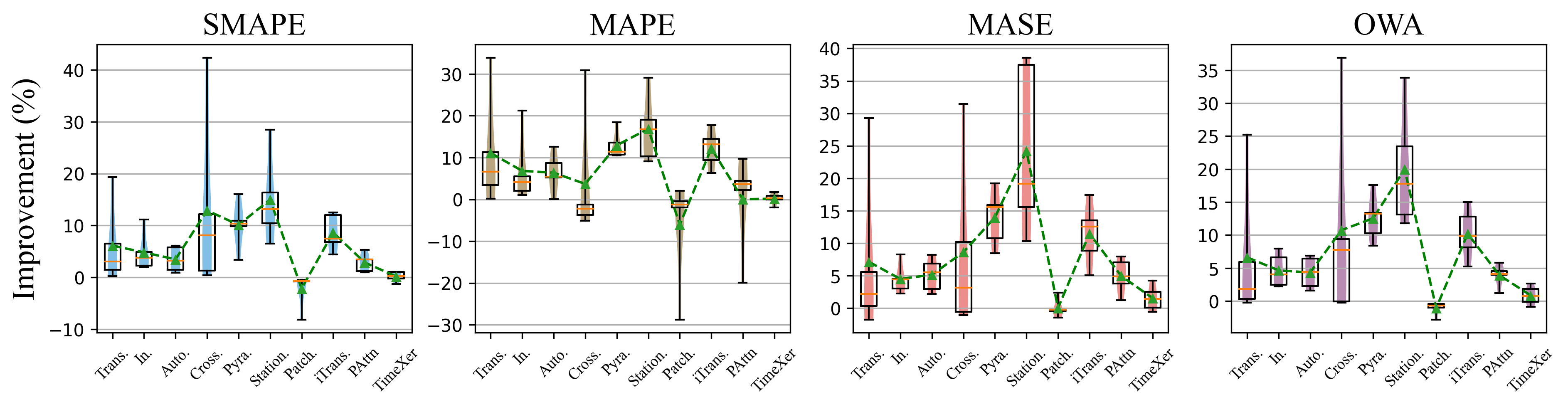}
	\caption{Box plots of statistical results on the distribution of \textbf{short-term forecasts} for each metric. The horizontal axis denotes the models and the vertical axis is the percentage of lift after using the structured matrix. The width of the color blocks in the box indicates the density of the distribution. The green dashed line is the mean and the orange dashed line is the median.}
	\label{fig:short}
\end{figure}

\textbf{Analysis:} The data in the table represents the percentage change in predictive performance of the model after applying structured matrix optimization, where positive numbers indicate improved performance. The visualization in Figure \ref{fig:short} displays the distribution of data for each model in Table \ref{tab:full_forecasting_results_m4}. From Table \ref{tab:full_forecasting_results_m4} and Figure \ref{fig:short}, it can be seen that the adaptability of the structured matrix to short-term forecasting tasks is very good, with over 66\% of tasks showing improved predictive performance, with a mean improvement of 5.35\%.

\subsubsection{Imputation}
\begin{table}[ht]
	\centering
	\caption{Full results for the imputation task. For brevity, we omit the percent signs (\%) from the data. Randomly mask 12.5\%, 25\%, 37.5\% and 50\% time points to compare the model performance under different missing degrees.}
	\resizebox{\textwidth}{!}{
		\begin{threeparttable}
			\renewcommand{\multirowsetup}{\centering}
			\begin{tabular}{c|c|ccc|ccc|ccc|ccc|ccc|ccc|ccc|ccc|ccc|ccc}
				\toprule
				\multicolumn{2}{c}{\multirow{1}{*}{Models}} & 
				\multicolumn{3}{c}{Transformer} &
				\multicolumn{3}{c}{Informer} &
				\multicolumn{3}{c}{Autoformer} &
				\multicolumn{3}{c}{Crossformer} & \multicolumn{3}{c}{Pyraformer} & \multicolumn{3}{c}{NST} & \multicolumn{3}{c}{PatchTST} &  \multicolumn{3}{c}{iTransformer} &
				\multicolumn{3}{c}{PAttn} & 
				\multicolumn{3}{c}{TimeXer}\\
				\cmidrule(lr){3-5} \cmidrule(lr){6-8}\cmidrule(lr){9-11} \cmidrule(lr){12-14}\cmidrule(lr){15-17}\cmidrule(lr){18-20}\cmidrule(lr){21-23}\cmidrule(lr){24-26}\cmidrule(lr){27-29}\cmidrule(lr){30-32}
				\multicolumn{2}{c}{Metric} & \scalebox{0.78}{MSE} & \scalebox{0.78}{MAE} & \scalebox{0.78}{R2} & \scalebox{0.78}{MSE} & \scalebox{0.78}{MAE} & \scalebox{0.78}{R2} & \scalebox{0.78}{MSE} & \scalebox{0.78}{MAE} & \scalebox{0.78}{R2} & \scalebox{0.78}{MSE} & \scalebox{0.78}{MAE} & \scalebox{0.78}{R2} & \scalebox{0.78}{MSE} & \scalebox{0.78}{MAE} & \scalebox{0.78}{R2} & \scalebox{0.78}{MSE} & \scalebox{0.78}{MAE} & \scalebox{0.78}{R2} & \scalebox{0.78}{MSE} & \scalebox{0.78}{MAE} & \scalebox{0.78}{R2} & \scalebox{0.78}{MSE} & \scalebox{0.78}{MAE} & \scalebox{0.78}{R2} & \scalebox{0.78}{MSE} & \scalebox{0.78}{MAE} & \scalebox{0.78}{R2} & \scalebox{0.78}{MSE} & \scalebox{0.78}{MAE} & \scalebox{0.78}{R2}\\
				\midrule
				\multirow{4}[2]{*}{\begin{sideways}ETTh1\end{sideways}} & 0.125 & 28.1 & 11.6 & 2.4 & 24.9 & 9.3 & 2.3 & 6.9 & 2.7 & 0.7 & -6.8 & -2.2 & -0.8 & 59.2 & 39.5 & 11.9 & 25.3 & 13.8 & 1.5 & 5.9 & 1.0 & 0.6 & 5.2 & 1.4 & 0.6 & 10.9 & 3.7 & 1.0 & 3.4 & 1.5 & 0.3 \\
				& 0.25  & 32.4 & 15.8 & 4.3 & 23.8 & 10.9 & 3.1 & 17.5 & 8.2 & 2.6 & -4.8 & -0.3 & -0.7 & 63.1 & 42.4 & 19.9 & 20.7 & 12.2 & 1.7 & 9.1 & 2.8 & 1.1 & 17.0 & 8.7 & 3.5 & 17.3 & 6.9 & 2.1 & 5.6 & 2.3 & 0.6 \\
				& 0.375 & 35.8 & 18.2 & 6.4 & 16.2 & 8.7 & 2.6 & 18.7 & 10.8 & 3.9 & -1.1 & 1.2 & -0.2 & 44.7 & 28.6 & 14.1 & 21.6 & 12.4 & 2.3 & 10.4 & 3.7 & 1.5 & 4.0 & 2.2 & 1.1 & 19.3 & 8.4 & 2.9 & 5.4 & 2.6 & 0.7 \\
				& 0.5   & 27.9 & 14.9 & 5.8 & 4.6 & 2.7 & 0.9 & 16.8 & 10.3 & 5.1 & 2.8 & 1.9 & 0.5 & 42.7 & 26.2 & 16.8 & 18.7 & 10.2 & 2.5 & 11.3 & 4.5 & 1.9 & -5.0 & -2.9 & -1.9 & 18.9 & 9.1 & 3.4 & 4.9 & 2.5 & 0.7 \\
				\midrule
				\multirow{4}[2]{*}{\begin{sideways}ETTh2\end{sideways}} & 0.125 & 51.4 & 31.9 & 15.5 & 14.6 & 7.4 & 4.2 & 76.4 & 54.0 & 114.1 & 8.0 & 2.5 & 0.8 & 66.8 & 45.3 & 29.9 & 10.1 & 6.6 & 0.3 & 0.3 & -0.1 & 0.0 & 7.6 & 4.1 & 0.6 & 1.5 & 1.1 & 0.0 & -0.3 & -0.6 & -0.01 \\
				& 0.25  & 24.7 & 14.7 & 6.4 & 50.3 & 29.3 & 31.6 & 20.0 & 7.3 & 4.1 & 16.3 & 9.2 & 2.0 & 55.5 & 37.8 & 21.7 & 12.6 & 8.8 & 0.4 & 1.9 & 0.8 & 0.1 & 6.5 & 2.9 & 0.6 & 2.6 & 1.5 & 0.1 & 0.4 & 0.1 & 0.02 \\
				& 0.375 & 21.1 & 13.3 & 5.9 & -18.3 & -9.9 & -5.0 & 30.3 & 14.8 & 9.9 & 20.9 & 11.8 & 3.1 & 53.7 & 38.0 & 36.5 & 13.0 & 8.7 & 0.5 & 2.7 & 1.4 & 0.1 & 11.5 & 5.4 & 1.4 & 2.2 & 1.5 & 0.1 & 1.3 & 0.6 & 0.1 \\
				& 0.5   & 15.9 & 12.0 & 4.9 & -35.7 & -13.3 & -10.2 & 27.5 & 14.0 & 13.2 & 26.7 & 16.5 & 4.5 & 67.1 & 48.0 & 83.8 & 11.3 & 6.9 & 0.5 & 3.0 & 1.8 & 0.2 & 32.5 & 16.6 & 7.4 & 2.2 & 1.7 & 0.1 & 1.6 & 0.8 & 0.1 \\
				\midrule
				\multirow{4}[2]{*}{\begin{sideways}ETTm1\end{sideways}} & 0.125 & 52.2 & 31.1 & 2.3 & 35.2 & 20.8 & 1.4 & 47.2 & 27.7 & 67.6 & 21.5 & 9.5 & 1.2 & 45.2 & 27.3 & 2.2 & 33.5 & 19.5 & 0.9 & 11.4 & 3.9 & 0.5 & -4.7 & -2.1 & -0.2 & 14.9 & 9.9 & 0.6 & 4.9 & 0.9 & 0.2 \\
				& 0.25  & 43.6 & 26.4 & 2.3 & 27.3 & 14.9 & 1.5 & 66.9 & 44.1 & 152.5 & 34.2 & 17.5 & 2.3 & 48.6 & 31.5 & 3.2 & 39.5 & 23.5 & 1.4 & 19.6 & 9.0 & 1.0 & 3.6 & 1.8 & 0.2 & 22.8 & 15.2 & 1.1 & 2.4 & 0.4 & 0.1 \\
				& 0.375 & 56.0 & 35.0 & 4.8 & 29.6 & 15.4 & 2.3 & 58.7 & 36.9 & 233.3 & 39.1 & 20.6 & 3.3 & 47.1 & 30.5 & 3.7 & 40.5 & 23.9 & 1.8 & 22.4 & 11.7 & 1.4 & 22.4 & 12.5 & 2.2 & 26.0 & 17.5 & 1.5 & 2.5 & 1.1 & 0.1 \\
				& 0.5   & 24.4 & 14.7 & 1.4 & 22.3 & 10.9 & 2.1 & 69.1 & 46.0 & 826.6 & 43.4 & 22.9 & 4.3 & 49.2 & 33.2 & 4.9 & 36.9 & 21.1 & 2.0 & 23.8 & 12.9 & 1.7 & 29.9 & 17.7 & 4.7 & 27.5 & 18.1 & 1.9 & 9.4 & 4.9 & 0.5 \\
				\midrule
				\multirow{4}[2]{*}{\begin{sideways}ETTm2\end{sideways}} & 0.125 & 25.1 & 7.2 & 2.7 & 58.1 & 36.1 & 11.8 & 75.6 & 49.9 & 106.2 & 19.3 & 9.4 & 1.0 & 68.4 & 51.9 & 21.8 & 19.3 & 17.4 & 0.3 & 2.6 & 0.9 & 0.0 & 13.1 & 6.1 & 0.5 & 3.4 & 4.3 & 0.1 & 1.2 & 0.4 & 0.02 \\
				& 0.25  & 19.6 & 9.1 & 2.5 & 46.6 & 26.4 & 10.3 & 46.7 & 28.0 & 110.0 & 37.3 & 20.9 & 2.4 & 60.4 & 41.4 & 20.5 & 19.1 & 16.2 & 0.3 & 6.9 & 3.8 & 0.1 & 17.6 & 7.2 & 1.0 & 6.6 & 7.5 & 0.1 & 2.1 & 0.9 & 0.04 \\
				& 0.375 & 60.9 & 39.7 & 12.3 & 39.9 & 20.4 & 11.7 & 63.2 & 40.4 & 215.5 & 41.0 & 23.3 & 3.6 & 60.8 & 43.6 & 27.1 & 19.0 & 15.0 & 0.4 & 8.0 & 4.8 & 0.2 & 22.1 & 11.3 & 1.8 & 8.2 & 8.5 & 0.2 & 2.6 & 1.5 & 0.05 \\
				& 0.5   & 57.4 & 36.0 & 11.3 & 22.3 & 12.4 & 6.2 & 70.5 & 44.4 & 203.6 & 35.0 & 20.2 & 3.3 & 64.0 & 45.0 & 38.7 & 19.4 & 14.2 & 0.4 & 9.0 & 6.0 & 0.2 & 74.7 & 47.8 & 33.2 & 8.7 & 8.4 & 0.2 & 4.3 & 3.2 & 0.1 \\
				\midrule
				\multirow{4}[2]{*}{\begin{sideways}ECL\end{sideways}} & 0.125 & 3.3 & 0.3 & 0.6 & -0.9 & 1.4 & -0.2 & -5.6 & -2.7 & -1.3 & 20.2 & 10.3 & 1.7 & 28.1 & 13.1 & 9.2 & 1.2 & 1.5 & 0.1 & 17.3 & 9.6 & 1.2 & 11.9 & 7.8 & 1.1 & 20.1 & 15.1 & 1.4 & 9.5 & 5.0 & 0.6 \\
				& 0.25  & 3.8 & 0.2 & 0.7 & 2.3 & 3.1 & 0.5 & -4.4 & -1.8 & -1.1 & 21.5 & 11.6 & 2.2 & 28.8 & 14.4 & 10.2 & 1.1 & 1.7 & 0.1 & 18.2 & 10.7 & 1.6 & 16.7 & 10.3 & 2.0 & 22.9 & 17.1 & 2.1 & 11.0 & 6.2 & 0.8 \\
				& 0.375 & 4.6 & 65.9 & 0.9 & 3.4 & 3.1 & 0.8 & -5.6 & -2.1 & -1.6 & 20.2 & 11.0 & 2.2 & 22.1 & 11.2 & 7.3 & 1.4 & 1.5 & 0.1 & 18.7 & 11.3 & 1.9 & 13.5 & 8.2 & 1.9 & 22.3 & 16.6 & 2.4 & 11.9 & 7.3 & 1.0 \\
				& 0.5   & 4.8 & 0.6 & 1.0 & 0.4 & 2.3 & 0.1 & -3.4 & -1.2 & -1.1 & 17.9 & 9.2 & 2.2 & 21.7 & 10.8 & 7.4 & 1.9 & 1.6 & 0.2 & 17.3 & 10.5 & 2.1 & 2.6 & 2.4 & 0.4 & 17.1 & 12.8 & 2.2 & 9.6 & 5.4 & 1.0 \\
				\midrule
				\multirow{4}[2]{*}{\begin{sideways}WTH\end{sideways}} & 0.125 & -7.4 & -10.3 & -0.4 & 12.2 & 9.2 & 1.0 & 98.3 & 88.9 & 105.2 & -0.8 & 0.9 & -0.1 & 24.9 & 25.9 & 1.7 & -3.7 & -3.5 & -0.2 & -1.5 & 0.7 & -0.1 & -1.1 & 0.2 & -0.1 & 2.5 & 3.2 & 0.1 & -6.7 & -8.0 & -0.3 \\
				& 0.25  & -3.0 & -9.4 & -0.2 & -0.4 & -1.3 & 0.0 & 93.0 & 76.1 & 111.5 & 7.1 & 5.7 & 0.6 & 25.0 & 24.2 & 2.1 & 4.5 & 5.1 & 0.3 & 1.8 & 3.9 & 0.1 & 32.2 & 31.9 & 4.3 & -0.2 & -2.0 & 0.0 & -1.7 & -1.9 & -0.1 \\
				& 0.375 & 6.7 & 10.7 & 0.5 & -6.5 & -2.9 & -0.5 & -38.5 & -27.4 & -6.2 & 12.7 & 9.2 & 1.2 & 16.2 & 14.4 & 1.4 & 0.0 & 3.2 & 0.0 & -0.7 & -0.1 & -0.1 & 59.5 & 48.6 & 19.2 & 4.8 & 6.6 & 0.3 & 9.2 & 11.4 & 0.6 \\
				& 0.5   & 11.1 & 15.9 & 1.0 & -14.1 & -9.0 & -1.3 & -47.1 & -29.9 & -10.1 & 14.8 & 10.8 & 1.5 & 14.4 & 13.9 & 1.3 & 2.6 & 4.3 & 0.2 & -1.3 & 46.4 & -4.2 & 74.0 & 58.5 & 64.7 & 3.3 & 2.5 & 0.2 & 2.5 & 5.6 & 0.2 \\
				
				\midrule
				\multicolumn{2}{c|}{Avg} & \multicolumn{3}{c|}{15.29\%} & \multicolumn{3}{c|}{8.94\%} & \multicolumn{3}{c|}{50.03\%} & \multicolumn{3}{c|}{10.31\%} & \multicolumn{3}{c|}{30.74\%} & \multicolumn{3}{c|}{8.80\%} & \multicolumn{3}{c|}{5.46\%} & \multicolumn{3}{c|}{12.86\%} & \multicolumn{3}{c|}{7.02\%} & \multicolumn{3}{c}{2.20\%} \\

				\bottomrule
			\end{tabular}
		\end{threeparttable}
	}
	\label{tab:full_imputation_results}
\end{table}
In real-world production scenarios, due to sensor or network failures, the collected time series data may be partially lost. Incomplete datasets can pose obstacles to downstream tasks, hence imputation is widely applied in practice. 

\textbf{Result:} Experiments are carried out on ETT, Electricity and Weather datasets, where encountering the data-missing issue is frequent. To evaluate the model's capability under varying levels of missing data, we randomly conceal time points at 12.5\%, 25\%, 37.5\% and 50\% ratios. Table \ref{tab:full_imputation_results} presents the experimental results of imputation tasks. 

\begin{figure}[htb]
	\centering
	\includegraphics[width=1.0\linewidth]{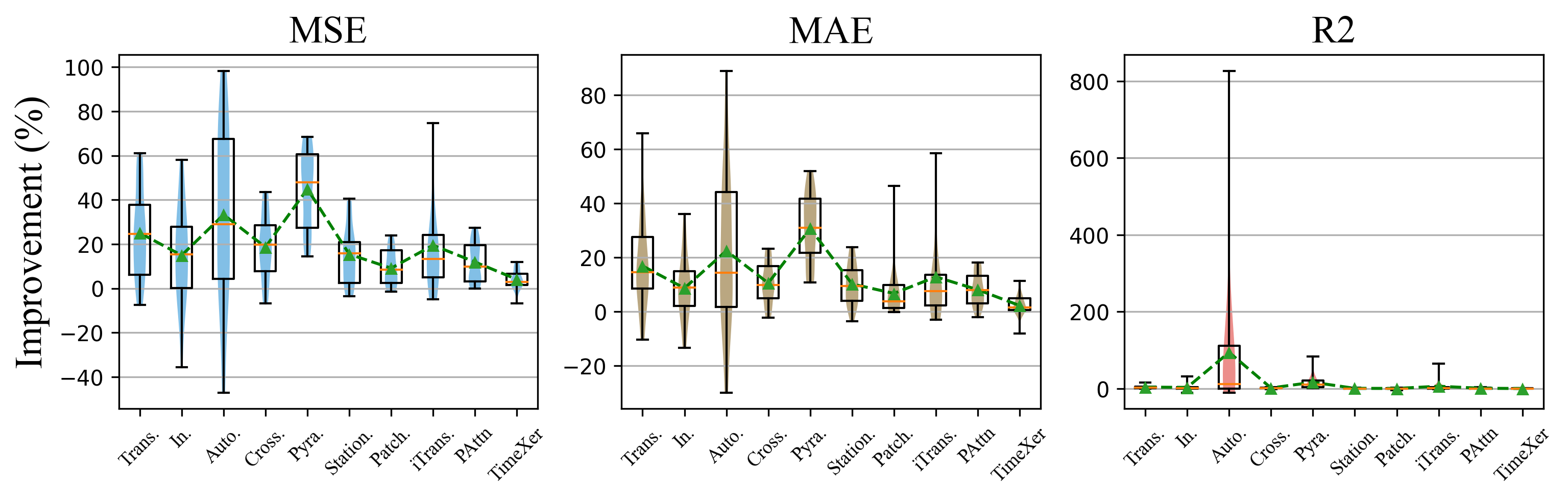}
	\caption{Box plots of statistical results for the distribution of \textbf{imputation} for each metric. The horizontal axis is the model and the vertical axis is the percentage of lift after using the structured matrix. The width of the color blocks in the box indicates the density of the distribution. The green dashed line is the mean and the orange dashed line is the median.}
	\label{fig:imputation}
\end{figure}

\textbf{Analysis:} The visualization in Figure \ref{fig:imputation} displays the distribution of data for each Transformer-based model. Positive advancements are evident across almost all 9 Transformer-based models, with Autoformer showcasing the most significant improvement, averaging 50\%. The result indicates that our method performs well on the imputation task and can capture temporal changes from incomplete time series.

\subsubsection{Classification}
\begin{table}[ht]
	\centering
	\caption{Full results for the classification task. $\ast.$ indicates the name of $\ast$former. Positive numbers indicate performance improvement, while negative numbers indicate the opposite, regardless of whether the evaluation metric is higher-the-better or lower-the-better.}
	\resizebox{\textwidth}{!}{
		\begin{threeparttable}
			\begin{small}
				\renewcommand{\multirowsetup}{\centering}
				\begin{tabular}{c|cccccccccc}
					\toprule
					\diagbox{Datasets}{Models}  & Trans. & In. & Auto. & Cross. & Pyra. & NST & PatchTST & iTrans. & PAttn & TimeXer \\
					\midrule
					EthanolConcentration & 5.3\% & 10.1\% & -1.4\% & 12.7\% & -9.6\% & 7.0\% & 4.3\% & 10.6\% & 2.7\% & -9.0\% \\
					FaceDetection & -2.4\% & 0.5\% & 24.7\% & 36.6\% & 0.8\% & -1.0\% & -3.2\% & -1.6\% & -1.8\% & -1.9\% \\
					Handwriting & -15.0\% & -9.6\% & 4.4\% & 255.8\% & -36.7\% & -15.4\% & -6.6\% & -11.9\% & 90.8\% & 168.1\% \\
					Heartbeat & 0.7\% & -1.3\% & -4.1\% & 4.1\% & -8.0\% & -2.7\% & 0.0\% & -5.9\% & -0.7\% & 0.0\% \\
					JapaneseVowels & 1.1\% & 1.4\% & 0.9\% & 0.6\% & 3.8\% & 0.8\% & 0.3\% & 0.6\% & 6.6\% & 10.4\% \\
					PEMS-SF & 2.1\% & -2.7\% & 3.6\% & 340.0\% & 0.8\% & 13.4\% & -2.1\% & -2.1\% & -3.4\% & -4.7\% \\
					SelfRegulationSCP1 & -1.1\% & 3.9\% & 46.0\% & 4.3\% & 1.5\% & 2.7\% & 3.9\% & -1.8\% & -6.0\% & 2.1\% \\
					SelfRegulationSCP2 & -6.9\% & -3.1\% & -6.2\% & 4.1\% & -6.2\% & -5.3\% & -3.2\% & -7.1\% & 6.6\% & 3.2\% \\
					SpokenArabicDigits & -0.6\% & -0.3\% & 0.2\% & -0.4\% & -0.7\% & -0.5\% & -0.3\% & 0.1\% & -0.5\% & -5.4\% \\
					UWaveGestureLibrary & -0.4\% & -1.1\% & 25.5\% & 8.3\% & -6.5\% & 1.9\% & -0.8\% & -1.5\% & 0.8\% & -1.2\% \\
					\midrule
					Average & -2.5\% & -1.4\% & 10.6\% & 72.6\% & -5.7\% & -0.6\% & -1.3\% & -3.5\% & 10.3\% & 19.0\% \\
					\bottomrule
				\end{tabular}
			\end{small}
		\end{threeparttable}
	}
	\label{tab:full_classification_results}
\end{table}

\textbf{Result:} We have selected 10 multidimensional datasets from the UEA time series classification repository\cite{Anthony2018Corr}, covering gesture, activity and audio recognition, heart rate monitoring for medical diagnosis and other real-world tasks.
Table \ref{tab:full_classification_results} presents the experimental results of classification tasks. 

\begin{figure}[htb]
	\centering
	\includegraphics[width=0.35\linewidth]{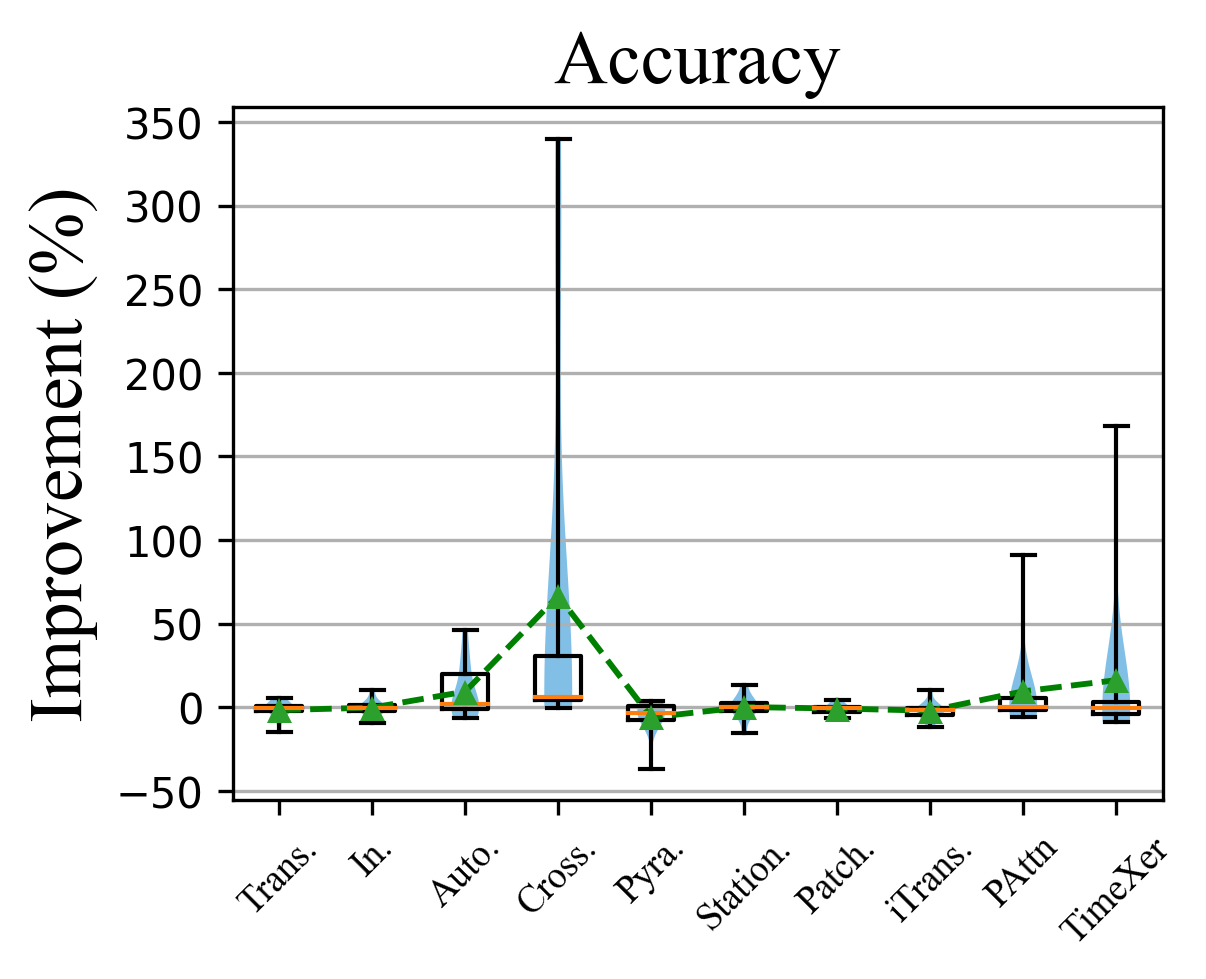}
	\caption{Box plots of statistical results on the distribution of \textbf{classification} for each metric. The horizontal axis denotes the models and the vertical axis is the percentage of lift after using the structured matrix. The width of the color blocks in the box indicates the density of the distribution. The green dashed line is the mean and the orange dashed line is the median.}
	\label{fig:classification}
\end{figure}

\textbf{Analysis:} The visualization in Figure \ref{fig:classification} displays the distribution of data for each model in Table \ref{tab:full_classification_results}. It is evident that Pyraformer's optimization performance is lacking, resulting in an average decrease of 5.7\% in classification accuracy. This decline is likely due to the sampling operations within Pyraformer's pyramid structure, which may disrupt the information organization of the structured matrix, leading to the loss of crucial information required for accurate classification. Apart from Pyraformer, the performance of the other 8 Transformer-based models is acceptable, with the worst one only decreasing the accuracy by 3.5\% (iTransformer).

\subsubsection{Anomaly Detection}

\begin{table}[htb]
	\centering
	\caption{Full results for the anomaly detection task. The P, R and F1 represent precision, recall and F1-score, respectively. F1-score is the harmonic mean of precision and recall. A higher value indicates a better performance.}
	\resizebox{\textwidth}{!}{
		\begin{threeparttable}
			\begin{small}
				\renewcommand{\multirowsetup}{\centering}
				\setlength{\tabcolsep}{1.4pt}
				\begin{tabular}{c|ccc|ccc|ccc|ccc|ccc|c}
					\toprule
					\multicolumn{1}{c}{\scalebox{0.8}{Datasets}} & 
					\multicolumn{3}{c}{\scalebox{0.8}{\rotatebox{0}{SMD}}} &
					\multicolumn{3}{c}{\scalebox{0.8}{\rotatebox{0}{MSL}}} &
					\multicolumn{3}{c}{\scalebox{0.8}{\rotatebox{0}{SMAP}}} &
					\multicolumn{3}{c}{\scalebox{0.8}{\rotatebox{0}{SWaT}}} & 
					\multicolumn{3}{c}{\scalebox{0.8}{\rotatebox{0}{PSM}}} & \scalebox{0.8}{Avg F1} \\
					\cmidrule(lr){2-4} \cmidrule(lr){5-7}\cmidrule(lr){8-10} \cmidrule(lr){11-13}\cmidrule(lr){14-16}
					\multicolumn{1}{c}{\scalebox{0.8}{Metrics}} & \scalebox{0.8}{P} & \scalebox{0.8}{R} & \scalebox{0.8}{F1} & \scalebox{0.8}{P} & \scalebox{0.8}{R} & \scalebox{0.8}{F1} & \scalebox{0.8}{P} & \scalebox{0.8}{R} & \scalebox{0.8}{F1} & \scalebox{0.8}{P} & \scalebox{0.8}{R} & \scalebox{0.8}{F1} & \scalebox{0.8}{P} & \scalebox{0.8}{R} & \scalebox{0.8}{F1} & \scalebox{0.8}{(\%)}\\
					\toprule
					Trans.                & 36.9\% & 221.0\% & 138.0\% & 0.9\%  & 0.3\%  & 0.6\%  & -2.5\% & -16.8\% & -11.4\% & -1.9\% & -5.0\%  & -3.6\% & 15.5\% & -12.2\% & -1.2\% & 23.9\% \\
					In.                   & 0.6\%  & 0.0\%   & 0.3\%   & 2.7\%  & 2.0\%  & 2.4\%  & -0.9\% & -2.4\%  & -1.8\%  & -2.3\% & -7.3\%  & -5.0\% & 0.2\%  & -4.0\%  & -2.3\% & -1.2\% \\
					Auto.                 & -0.2\% & -1.0\%  & -0.6\%  & -0.1\% & 4.2\%  & 2.3\%  & -4.4\% & -39.5\% & -26.1\% & 2.1\%  & 9.0\%   & 5.8\%  & 0.0\%  & 0.1\%   & 0.1\%  & -3.2\% \\
					Cross.               & 1.3\%  & 2.9\%   & 2.2\%   & 1.5\%  & 0.8\%  & 1.1\%  & -0.9\% & 0.5\%   & 0.0\%   & -1.2\% & -5.6\%  & -3.6\% & 0.2\%  & -1.1\%  & -0.5\% & -0.2\% \\
					Pyra.                 & 3.0\%  & 11.6\%  & 7.7\%   & 0.6\%  & -5.3\% & -2.7\% & -1.6\% & -15.3\% & -10.2\% & -3.0\% & -11.5\% & -7.6\% & 0.4\%  & -6.0\%  & -3.5\% & -2.9\% \\
					NST & 1.6\%  & 8.7\%   & 5.5\%   & 0.2\%  & -3.8\% & -1.9\% & -0.8\% & -1.3\%  & -1.1\%  & -0.7\% & -2.1\%  & -1.5\% & 1.5\%  & 0.4\%   & 1.0\%  & 0.4\%  \\
					PatchTST          & 0.3\%  & 1.0\%   & 0.7\%   & -0.2\% & 0.3\%  & 0.1\%  & -0.1\% & -1.3\%  & -0.8\%  & -0.1\% & -0.7\%  & -0.4\% & -0.5\% & -4.6\%  & -2.7\% & -0.6\% \\
					iTrans.                   & -0.3\% & -1.3\%  & -0.9\%  & 1.5\%  & 0.8\%  & 1.1\%  & -2.0\% & -17.1\% & -11.5\% & 0.1\%  & -1.8\%  & -0.9\% & 0.9\%  & 9.7\%   & 5.3\%  & -1.1\% \\
					PAttn & 0.5\% & 1.1\% & 1.0\% & -0.1\% & 2.3\% & 1.1\% & 0.1\% & 0.7\% & 0.5\% & 0.2\% & 0.8\% & 0.5\% & 0.2\% & 1.9\% & 1.0\% & 0.8\% \\
					TimeXer & 0.0\% & 0.0\% & 0.0\% & 0.0\% & 0.1\% & 0.1\% & 0.1\% & -0.2\% & -0.1\% & 0.2\% & 1.9\% & 1.1\% & 0.1\% & 0.0\% & 0.0\% & 0.2\% \\
					\bottomrule
				\end{tabular}
			\end{small}
		\end{threeparttable}}
	\label{tab:full_anomaly_results}
\end{table}
\begin{figure}[htb]
	\centering
	\includegraphics[width=1.0\linewidth]{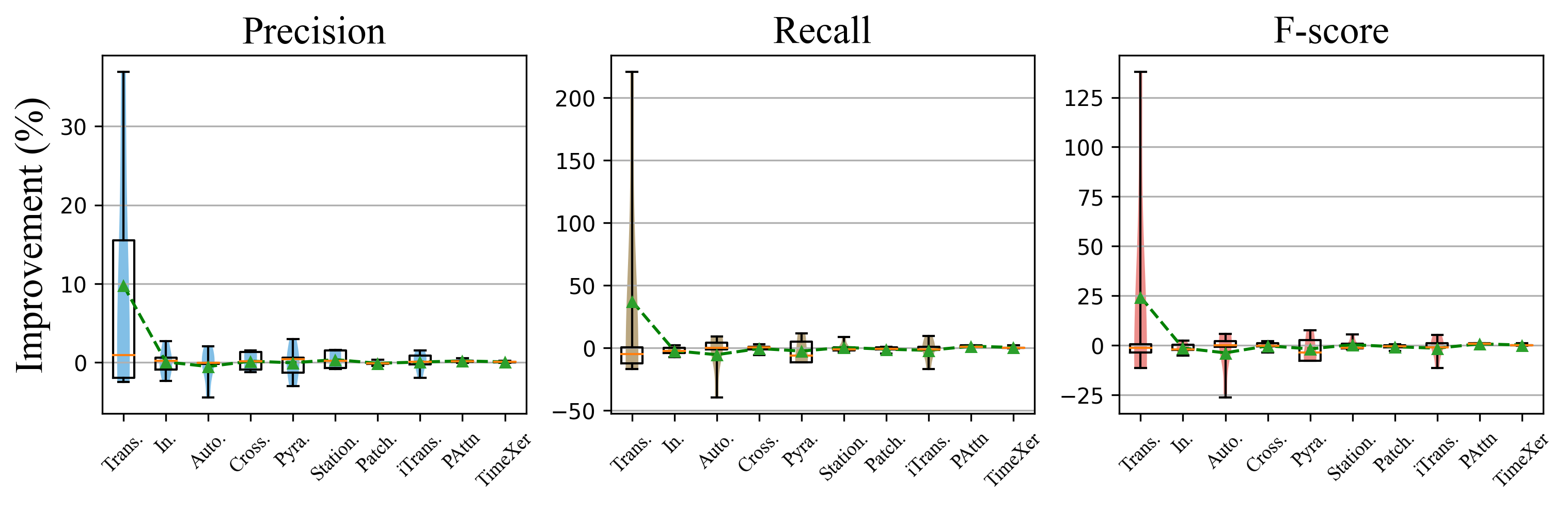}
	\caption{Box plots of statistical results on the distribution of \textbf{anomaly detection} for each metric. The horizontal axis denotes the models and the vertical axis is the percentage of lift after using the structured matrix. The width of the color blocks in the box indicates the density of the distribution. The green dashed line is the mean and the orange dashed line is the median.}
	\label{fig:anomaly}
\end{figure}

Similar to imputation tasks, the motivation for anomaly detection also arises from sensor failures, except that the failures do not result in data loss but rather in data anomalies. Anomalous data deviates significantly from the true values, which can severely impact the performance of prediction or classification tasks. Therefore, anomaly detection is crucial for industrial maintenance. 

\textbf{Result:} We focus on unsupervised time series anomaly detection, specifically detecting anomalous time points. We compare models from five widely used anomaly detection benchmarks: SMD\cite{Su2019kdd}, MSL\cite{Hundman2018kdd}, SMAP\cite{Hundman2018kdd}, SWaT\cite{Mathur2016cpsweek} and PSM\cite{Abdulaal2021kdd}, covering applications in service monitoring, space and earth exploration and water treatment. Following the preprocessing method in Anomaly Transformer\cite{Xu2022iclr}, we segment the dataset into contiguous, non-overlapping segments using a sliding window approach.
Table \ref{tab:full_anomaly_results} presents the experimental results of anomaly detection tasks. 

\textbf{Analysis:} The visualization in Figure \ref{fig:anomaly} displays the distribution of data for each model in Table \ref{tab:full_anomaly_results}. Except Transformer, the performance changes of other Transformer-based models are within 4\%. The worst-performing one is Autoformer, with an average performance decrease of only 3.2\%. This may come from the need for Transformer-based models in anomaly detection to identify rare abnormal time patterns. The attention mechanism calculates the similarity between each pair of time points, which may be dominated by the normal time points and dilute the focus of the attention mechanism, thus replacing the attention mechanism has little effect on the results.

\subsection{Comparative Analysis}\label{sec: ana}
\begin{figure}[ht]
	\centering
	\subfloat[] {
		\label{fig:radar}
		\includegraphics[width=0.33\linewidth]{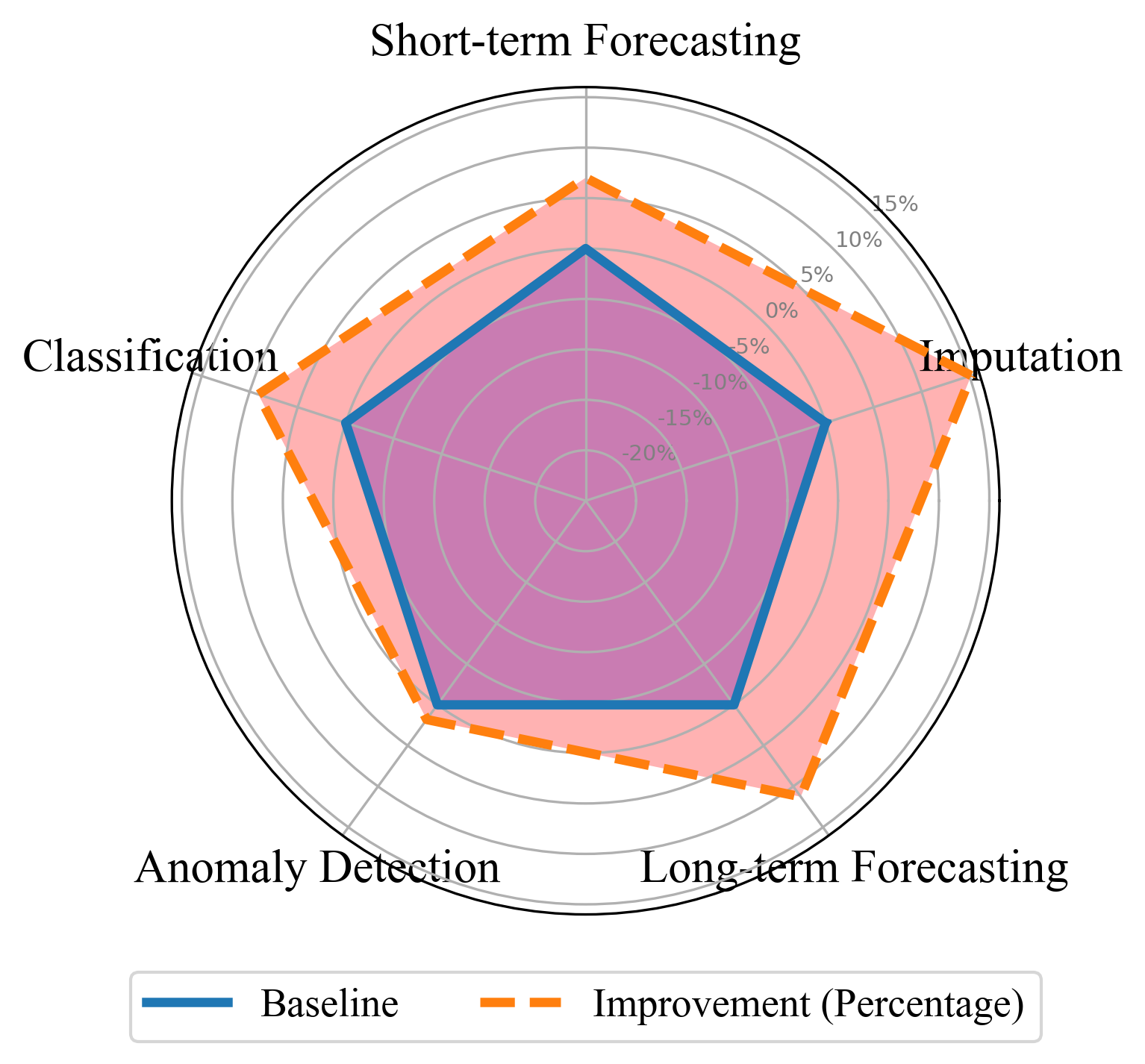}
	}
	\subfloat[long-term forecasting] {
		\label{fig:polygon_long}
		\includegraphics[width=0.33\linewidth]{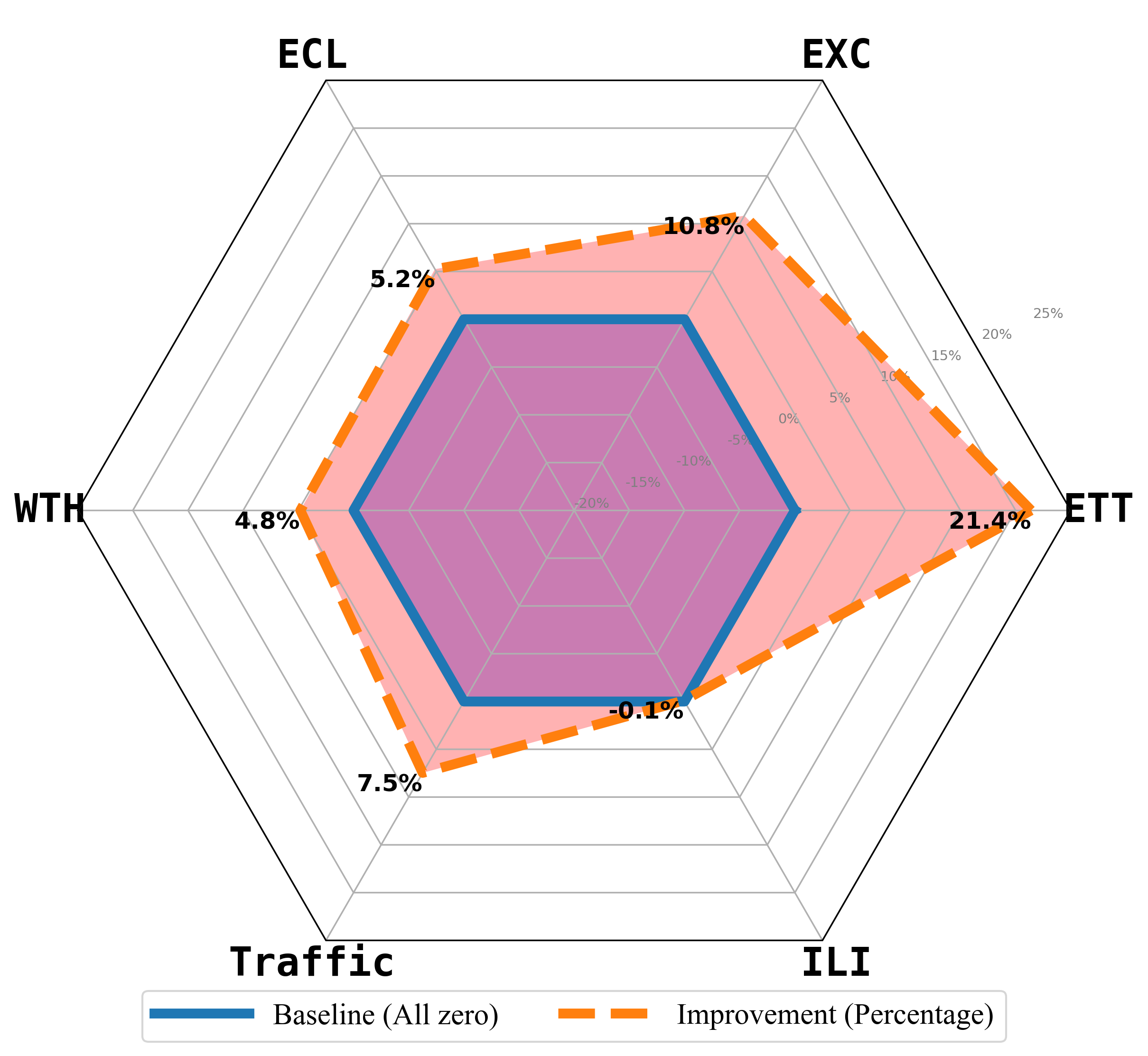}
	}
	\subfloat[short-term forecasting] {
		\label{fig:polygon_short}
		\includegraphics[width=0.33\linewidth]{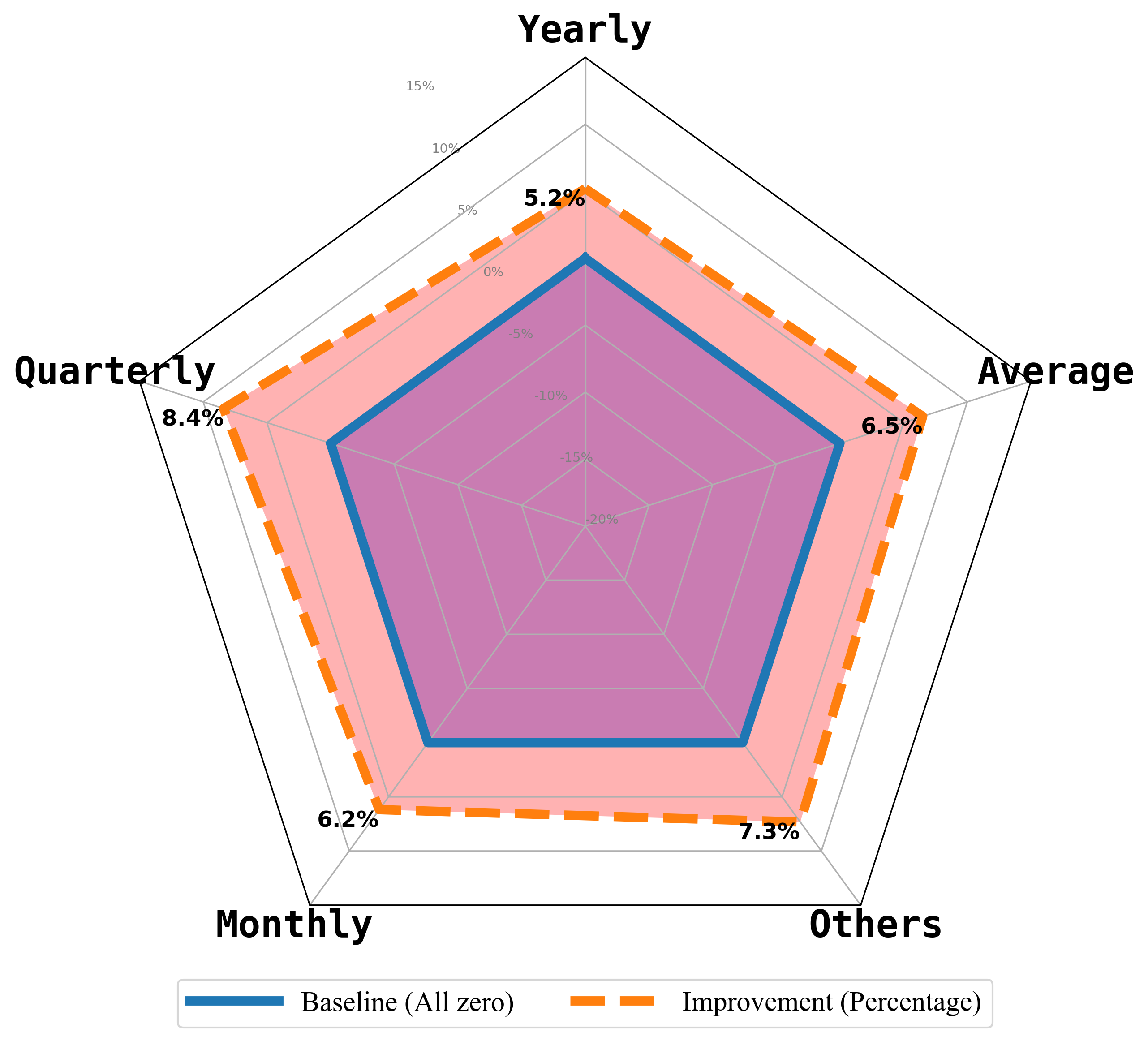}
	}\\
	\subfloat[imputation] {
		\label{fig:polygon_imputation}
		\includegraphics[width=0.33\linewidth]{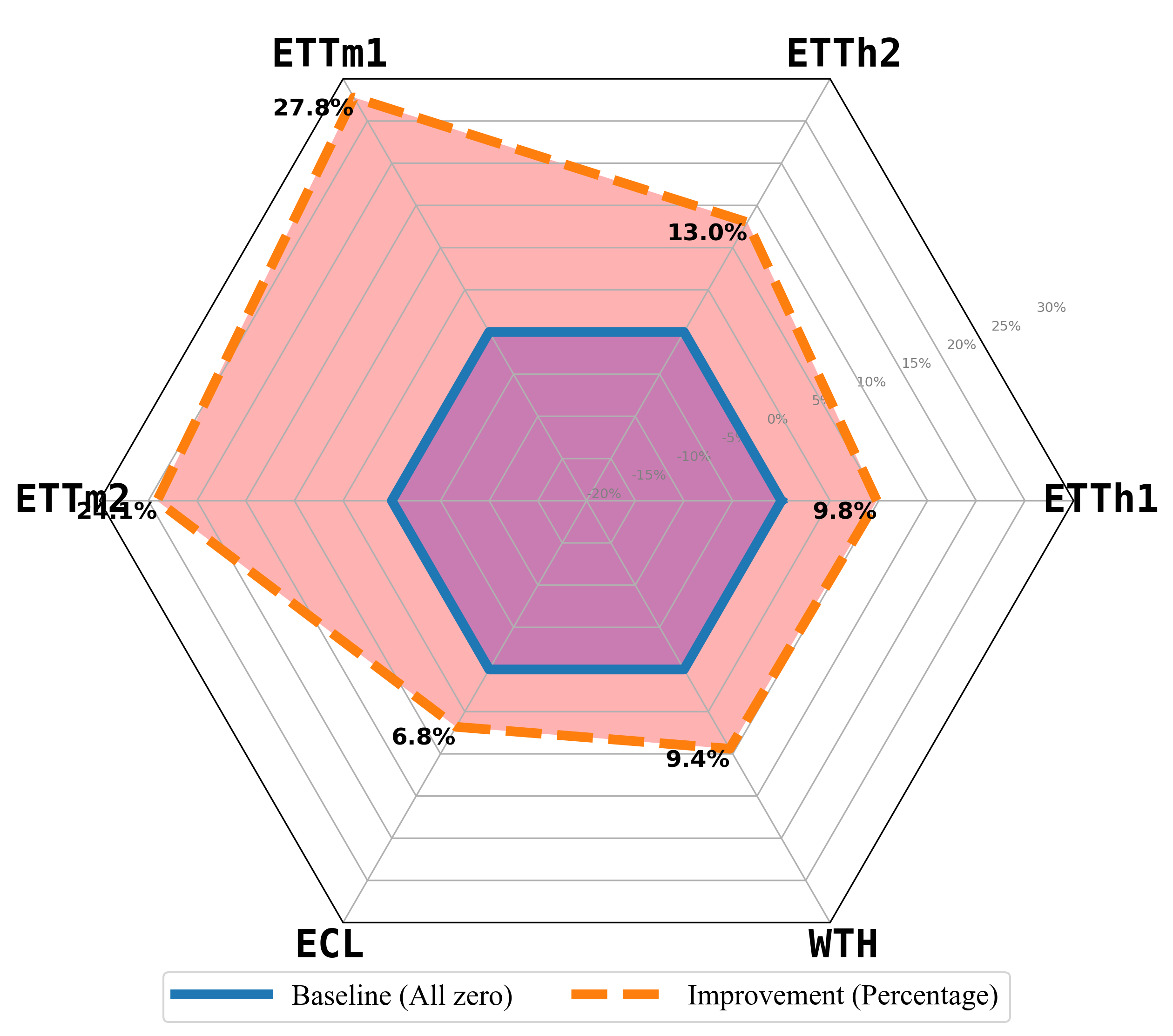}
	}
	\subfloat[anomaly detection] {
		\label{fig:polygon_anomaly}
		\includegraphics[width=0.33\linewidth]{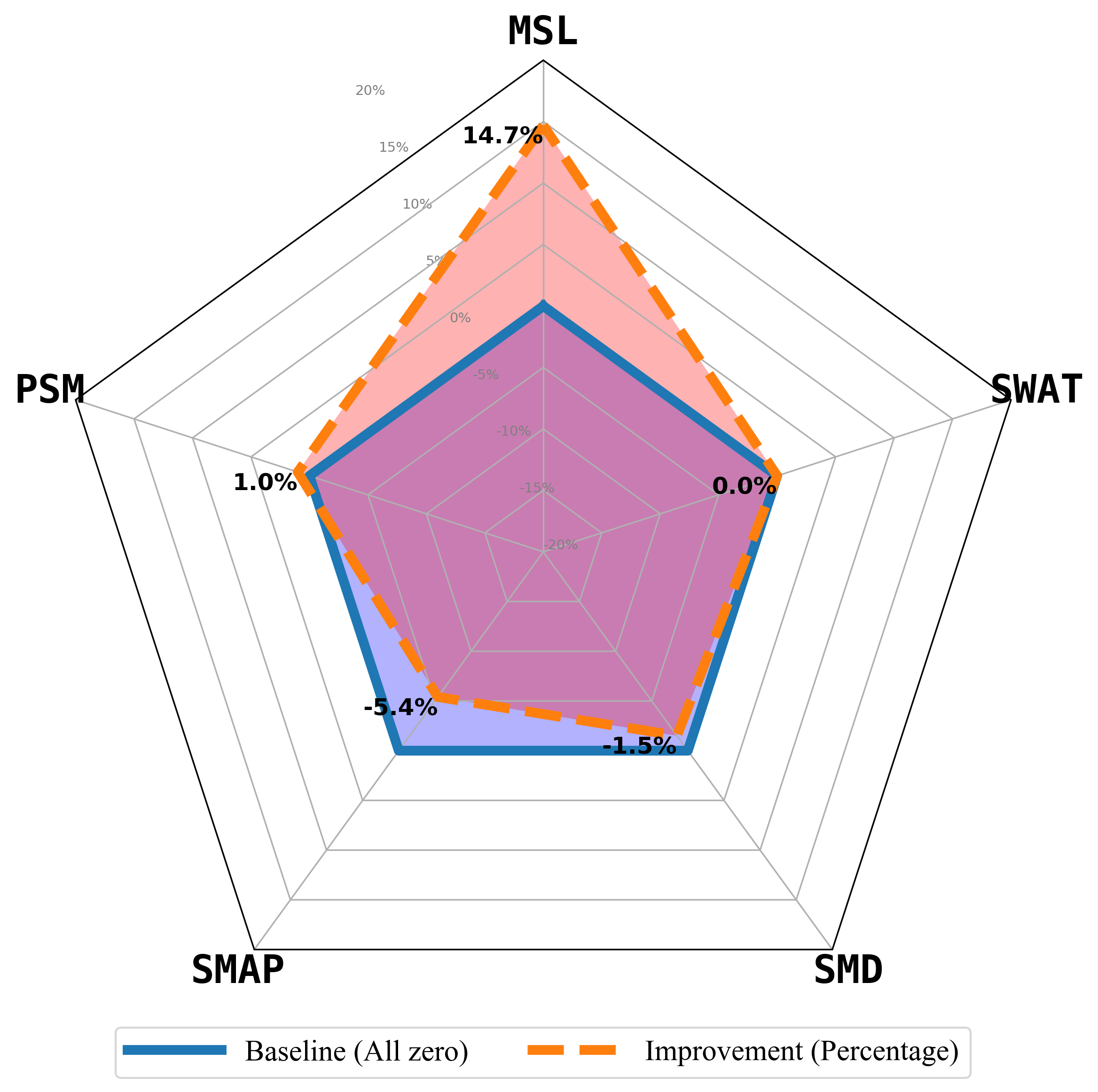}
	}
	\subfloat[classification] {
		\label{fig:polygon_classification}
		\includegraphics[width=0.33\linewidth]{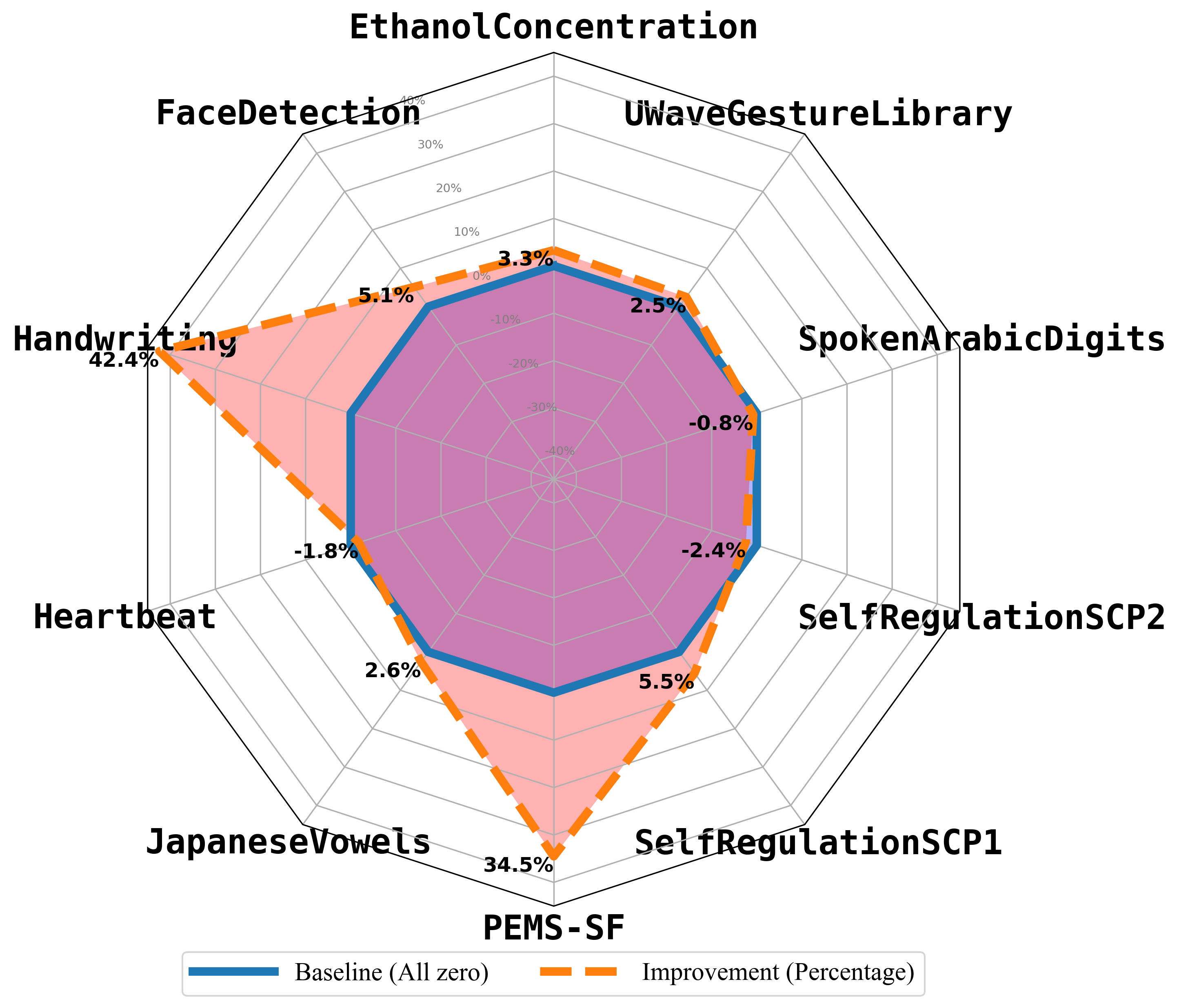}
	}
	
	\caption{(a) Radar chart depicting performance improvement ratios across 5 tasks. (b)-(f) Comparative analysis of performance improvement ratios achieved by structured matrices on specific datasets for each tasks.}
	
\end{figure}

In Section \ref{subsec:main result}, our experimental results demonstrate the effectiveness of our approach, with significant achievements in model efficiency and performance. We successfully reduced the parameters of the original model by 61.3\% and FLOPS by 66.1\% on average. Moreover, through a comprehensive experimental evaluation encompassing 2,769 performance tests, 1,955 experiments (72.4\%) showcased the optimization framework's ability to improve model performance, with an average performance enhancement of 12.4\%.
Our detailed analysis across five distinct tasks using Transformer-based models, as visualized in Figure \ref{fig:radar}, reveals consistent performance improvements compared to baseline models. By delving deeper into individual datasets within each task category (illustrated in Figures \ref{fig:polygon_long}-\ref{fig:polygon_classification}), the radar plots underscore the robustness and generalizability of our approach. These numerous experiments collectively demonstrate that our proposed optimization framework not only maintains but often enhances model performance while simultaneously improving computational efficiency. Specifically, our approach performs exceptionally well in optimizing forecasting and imputation tasks but exhibits limited performance on anomaly detection and classification tasks. This is primarily because forecasting and imputation tasks rely heavily on the model's ability to capture long-term dependencies and sequential patterns, which align closely with the strengths of our method.  
For the classification task, while our method performs well on the PEMS dataset due to its strong temporal characteristics, other datasets lack traditional time-series properties and do not exhibit clear long- and short-term dependency features, making it challenging for our method to be effective. Similarly, although the heartbeat dataset is a time-series dataset, its non-smooth nature further hinders the effectiveness of our approach.  
Moreover, anomaly detection and classification tasks often require the model to focus on fine-grained details or outlier patterns in the data, which are not fully captured by our current optimization strategy. These inherent differences in task requirements underscore the need for task-specific adaptations. Moving forward, we plan to explore enhancements to our method to improve its sensitivity to local variations and anomalies, thereby broadening its applicability across a wider range of time-series tasks.

However, the performance improvements are not uniform across all models. To systematically investigate the underlying factors contributing to these variations, we conducted a comprehensive analysis focusing on two key aspects: model architecture and feature processing mechanisms.
\begin{itemize}
	\item \textbf{Architectural Differences:} We observed that models with less consistent improvements (Crossformer and Pyraformer) often employ specialized mechanisms that deviate significantly from traditional transformer architectures:
	\begin{itemize}
		\item Crossformer utilizes cross-dimensional attention, which may interact differently with structured matrices compared to standard attention mechanisms.
		\item Pyraformer employs a pyramid-shaped hierarchical structure and multi-scale temporal convolutions, with its unique multi-resolution representation method potentially altering the effectiveness of structured matrix optimization. Specifically, its hierarchical attention mechanism and multi-scale feature extraction may introduce non-linear effects on matrix optimization.
	\end{itemize}
	
	\item \textbf{Feature Processing Mechanisms:} Different models employ unique approaches to extracting and processing temporal features:
	\begin{itemize}
		\item Crossformer leverages cross-dimensional attention mechanisms with dynamic dimension decomposition, enabling complex feature interactions that may fundamentally alter the traditional impact of structured matrix optimization by simultaneously capturing intricate temporal and feature-level patterns.
		\item Pyraformer utilizes multi-scale pyramid-shaped convolution networks with hierarchical attention mechanisms, allowing simultaneous processing of short-term and long-term dependencies through layered feature extraction that may diminish or fundamentally transform the performance gains from structured matrix optimization.
	\end{itemize}
\end{itemize}

We also evaluated FEDformer but excluded it from the main results due to its poor performance and architectural similarity to Autoformer. Our method faces challenges in optimizing FEDformer due to its use of attention in the frequency domain, a context where our approach is not currently applicable. Addressing this limitation will be a focus of our future work.
The presence of these advanced features may limit the additional benefits that can be gained from our structured matrix optimization approach. These architectures might already inherently capture some of the computational efficiencies that structured matrices provide in more standard models. Moreover, the complex interactions between these specialized components and structured matrices could potentially introduce unexpected behaviors or diminish the effectiveness of our optimization technique. 
It's important to note that our structured matrix optimization was primarily designed with traditional transformer architectures in mind. The more complex and specialized nature of these newer models may require a different approach to optimization, one that is more closely aligned with their unique operational principles. 

\subsection{Ablation study}
To thoroughly evaluate the effectiveness and efficiency of our proposed approach, we conducted a comprehensive ablation study. This study aims to isolate and analyze the impact of various components and design choices in our model, providing insights into their individual and collective contributions to the overall performance.
Our ablation experiments focus on three key aspects:Impact of Surrogate Blocks, Convergence and Trainability and Layer-wise Analysis.

Through these experiments, we aim to validate our design choices, quantify the improvements brought by each component and provide a deeper understanding of our model's behavior under various conditions. The following subsections detail each part of our ablation study, presenting the experimental setup, results and key findings.
\subsubsection{Study of Surrogate Blocks}
To validate the effectiveness of the surrogate blocks we proposed, ablation experiments are conducted on the ETTh1 (series length = 96) dataset by only replacing the linear projection layer, feed-forward network layer, or self-attention layer on Transformer. Experimental results can be found in Table \ref{tab:ablation}. Besides, we simultaneously counted the effects of the three surrogate blocks on the model size and the results are also shown in Table \ref{tab:ablation}.
From Table \ref{tab:ablation} we can conclude that our proposed surrogate blocks have a positive impact on the prediction results while effectively reducing the size of the model and no block is redundant. 

\begin{table}[H]
	\centering
	\caption{The impact of the surrogate blocks on model performance and size.}
	\begin{tabular}{l|c|c|c|c}
		\toprule
		& \multicolumn{2}{c|}{MSE} & \multicolumn{2}{c}{Model Size (MB)} \\
		\cmidrule(lr){2-3} \cmidrule(lr){4-5}
		& \multicolumn{1}{c|}{Value} & \multicolumn{1}{c|}{Impact} & \multicolumn{1}{c|}{Value} & \multicolumn{1}{c}{Impact} \\
		\midrule
		Transformer & 0.749 &   -    & 59.8  & - \\
		
		$\leftrightarrows$ SAB & 0.6401 & -14.54\% & 55.7  & -6.86\% \\
		$\leftrightarrows$ SFB & 0.7336 & -2.06\% & 36.3  & -39.30\% \\
		\bottomrule 
	\end{tabular}%
	
	\label{tab:ablation}
\end{table}%

In the SAB, the Monarch matrix can be substituted for the projection matrix. Since the original attention layer involves projecting query, key and value (QKV) embeddings, we explored replacing their respective projection matrices. To evaluate the effectiveness of these replacements, we conducted experiments across five distinct datasets: ETTh1, ETTh2, EXC, WTH and ILI.
Table \ref{tab:ablation-qkv} summarizes the average performance of various projection matrix replacement combinations. The results reveal that replacing all QKV projection matrices with the Monarch matrix yields the most significant improvements, as formalized in Equation (\ref{Eq:projection-m2}).
\begin{table}[htbp]
	\centering
	\caption{Average performance of different combinations of projection matrix substitutions on five datasets. Each column represents a combination of replacements, e.g., column QK indicates that only the projections of the query and keys are replaced, not the projections of the values. A downward arrow indicates that a lower value for the metric is better, while an upward arrow indicates that a higher value is better.}
	\begin{tabular}{l|ccccccc}
		\toprule
		& QKV   & QK    & QV    & KV    & Q     & K     & V \\
		\midrule
		\textdownarrow MSE   & \textbf{1.2332} & 1.4384 & 1.3978 & 1.3921 & 1.4248 & 1.4370 & 1.4624 \\
		\textdownarrow MAE   & \textbf{0.7127} & 0.7706 & 0.7443 & 0.7413 & 0.7499 & 0.7568 & 0.7682 \\
		\textuparrow R2    & \textbf{0.4183} & 0.3566 & 0.3906 & 0.3953 & 0.3781 & 0.3717 & 0.3589 \\
		\textdownarrow DTW   & \textbf{3.6511} & 3.7779 & 3.7019 & 3.7025 & 3.6196 & 3.6549 & 3.7924 \\
		\bottomrule
	\end{tabular}%
	\label{tab:ablation-qkv}%
\end{table}%

\subsubsection{Convergence and Trainability Analysis}
\begin{figure}[h]
	\centering
	\subfloat[]{
		\includegraphics[width=0.5\linewidth]{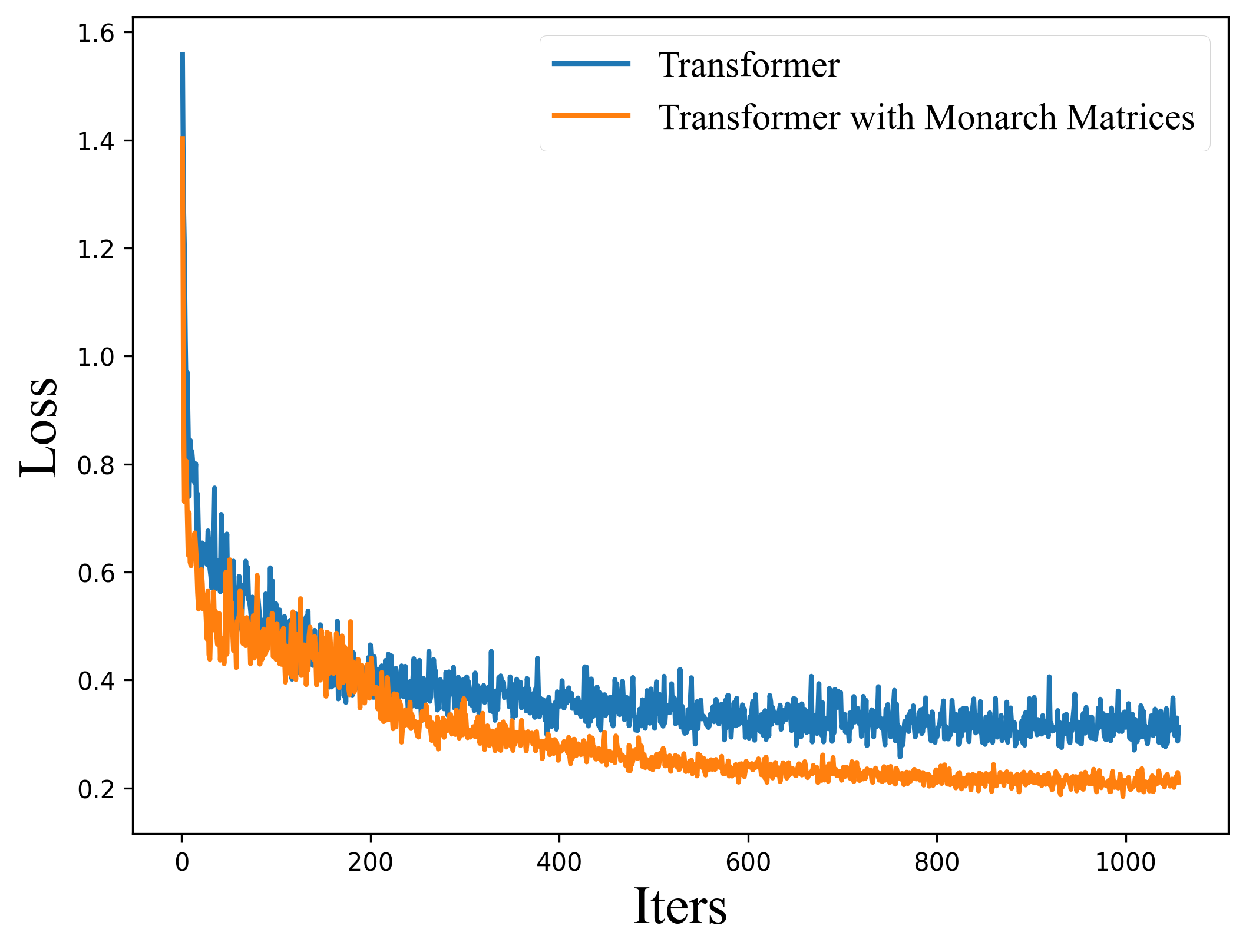}
		\label{fig:iters}	}
	\subfloat[]{
		\includegraphics[width=0.5\linewidth]{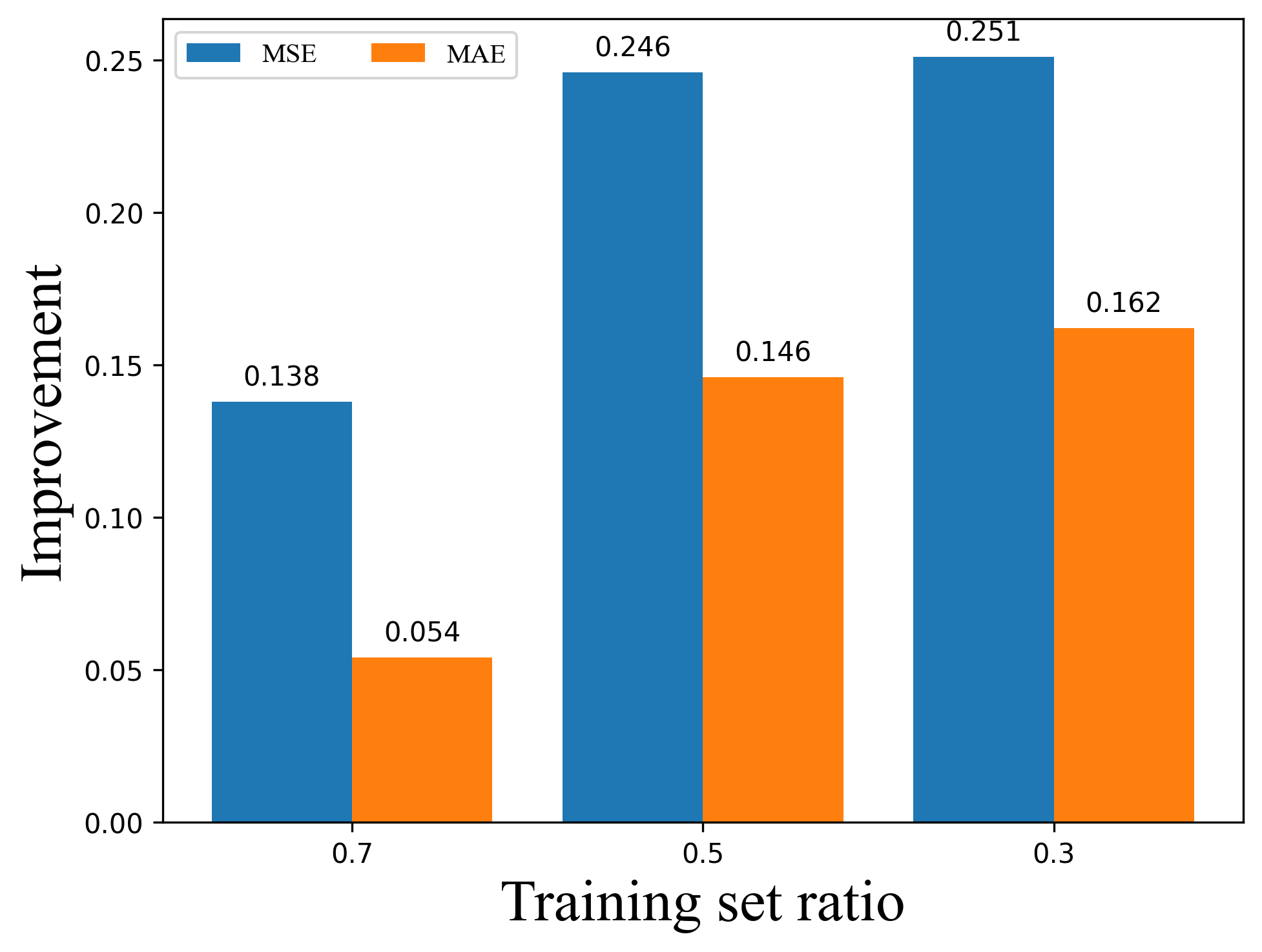}
		\label{fig:trainset}	}
	\caption{(a): Variation of loss with training iters. (b): Bar chart of experimental results with different training set sizes. The horizontal axis represents the proportion of training set size to the total dataset, while the vertical axis shows the performance improvement of the model using the structured matrix.}
\end{figure}

We also compared the convergence and trainability of the model before and after using the structured matrix. Figure \ref{fig:iters} shows the variation of losses of the model during the training process before and after the improvement. It is clear that the improved model achieves lower losses more quickly and with smaller fluctuations, indicating better convergence. Additionally, to validate our approach for making Transformer-based models easier to train, we continuously reduced the size of the training set and repeated comparative experiments. We chose the Transformer model and conducted experiments on the electricity, exchange rate and weather datasets. The experimental results are shown in Figure \ref{fig:trainset}, it can be seen that as the training set size decreases, the performance improvement ratio of the model optimized using structured matrices becomes increasingly significant. Therefore, it can be concluded that using structured matrices will make model training easier. In order to examine the inference time and GPU memory utilization of the model, we conducted a statistical analysis of the results of Transformer at various step lengths on the ETTh1 dataset. According to the results in Table \ref{tab:speed}, it is apparent that the model modified with SAB has faster inference speed and lower memory usage.
\begin{table}[htbp]
	\centering
	\caption{Comparison of inference time and GPU memory utilization with Attention blocks. \textbf{Bold} indicates better performance.}
	\begin{tabular}{c|c|cccc}
		\toprule
		\multicolumn{2}{c|}{Series Length} & 96    & 192   & 336   & 720 \\
		\midrule
		\multirow{2}[4]{*}{Attn} & Inf.(s/batch) & 3.39  & 3.50  & 4.98  & 6.44 \\
		\cmidrule{2-6}          & Mem.(GB) & 1.11  & 2.43  & 2.73  & 3.93 \\
		\midrule
		\multirow{2}[4]{*}{SAB} & Inf.(s/batch) & \textbf{3.23} & \textbf{3.26} & \textbf{4.77} & \textbf{5.82} \\
		\cmidrule{2-6}          & Mem.(GB) & \textbf{1.04} & \textbf{1.99} & \textbf{2.43} & \textbf{2.87} \\
		\bottomrule
	\end{tabular}%
	
	\label{tab:speed}%
\end{table}%

\subsubsection{Layer-wise Analysis of Surrogate Attention Block}
\begin{figure}
	\centering
	\includegraphics[width=1.0\linewidth]{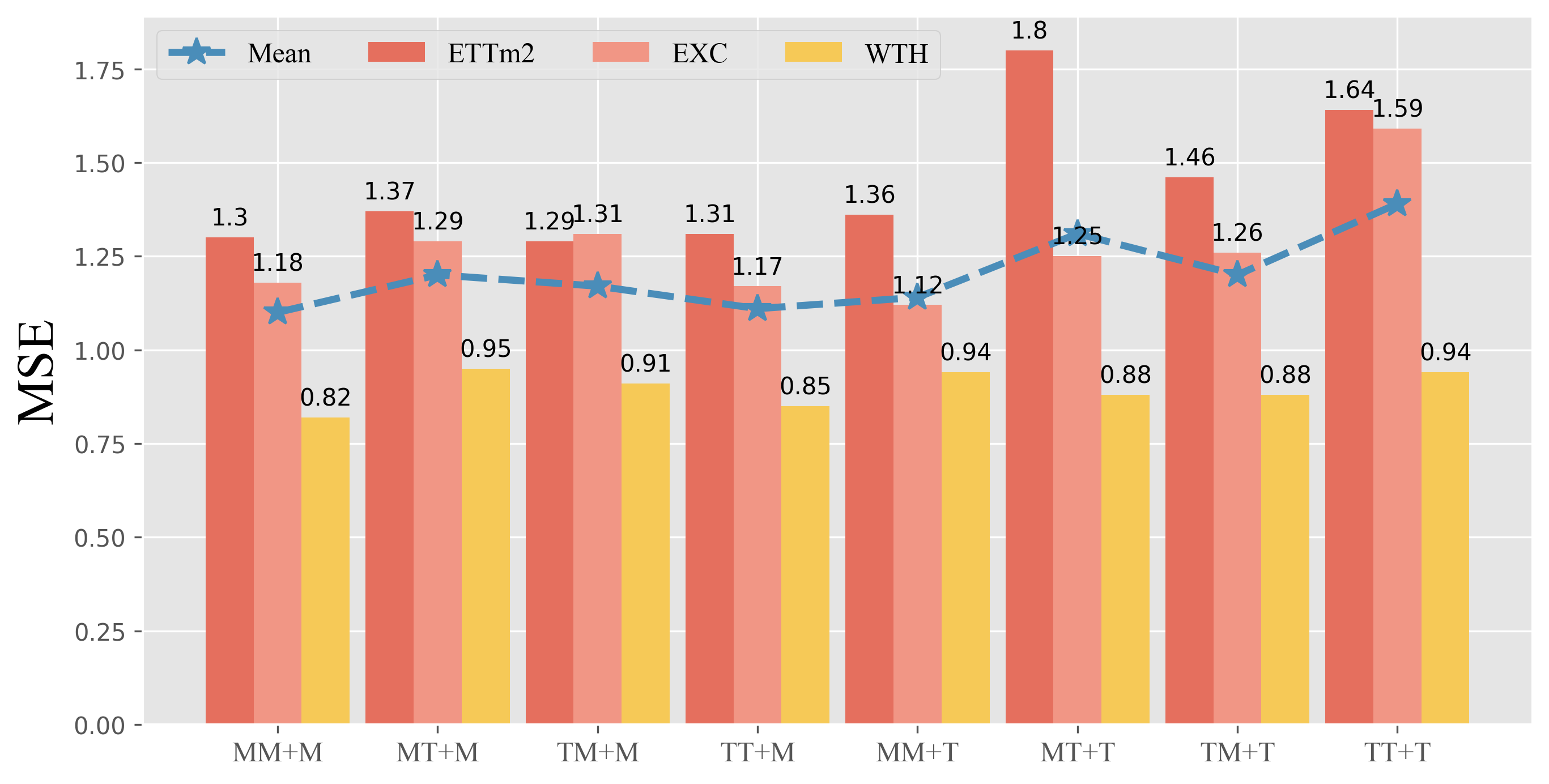}
	\caption{Comparison experiment results of replacing each layer with the Surrogate Attention Block. \emph{M} represents the use of the Surrogate Attention Block, while \emph{T} represents not using it. For example, \emph{MT+M} denotes replacing the first layer of the encoder with the Surrogate Attention Block, while keeping the second layer of the encoder unchanged.	}
	\label{fig:layers}
\end{figure}

In addition, we also attempted to verify whether the surrogate attention block is useful at each layer. For the forecasting task, in a common experimental setup, two encoder layers and one decoder layer are used. Therefore, we chose Informer with three datasets and conducted control experiments for each layer, the results are shown in Figure  \ref{fig:layers}. From the results, it is still the case that replacing the model in all layers performs the best.

\subsection{Comparison experiment}
We evaluate multiple optimize approaches on Vanilla Transformer, including sparsity methods like LogTrans and Informer, locality methods such as Autoformer (without decomposition) and CNNformer and our proposed novel optimization approach. The key metrics of interest are Mean Squared Error (MSE), number of parameters (G) and FLOPS (M), which comprehensively assess both predictive accuracy and computational efficiency. The results presented in Table \ref{tab:opt-comp} systematically illustrate the performance of each optimization method. Notably, our proposed approach achieves the lowest MSE of 0.666, significantly outperforming other methods. Moreover, it demonstrates exceptional computational efficiency, with the lowest parameter count and the smallest FLOPS, highlighting its superior performance and potential for practical implementation.

\begin{table}[H]
	\centering
	\caption{Performance comparison of optimization methods on Vanilla Transformer}
	\resizebox{\textwidth}{!}{
		\begin{tabular}{c|c|cc|cc|c}
			\toprule
			\multirow{3}[6]{*}{} & \multirow{3}[6]{*}{Vanilla Transformer} & \multicolumn{5}{c}{Optimize approaches} \\
			\cmidrule{3-7}          &       & \multicolumn{2}{c|}{Sparsity} & \multicolumn{2}{c|}{Locality} & \multirow{2}[4]{*}{Ours} \\
			\cmidrule{3-6}          &       & LogTrans & Informer & Autoformer w/o decomp & CNNformer &  \\
			\midrule
			MSE   & 0.749 & 0.762 & 0.860 & 0.740 & 0.727 & \textbf{0.666} \\
			Paras & 10.540 & 10.540 & 11.328 & 10.536 & 15.262 & \textbf{2.665} \\
			Flops & 1.189 & 1.189 & 1.093 & 1.169 & 1.718 & \textbf{0.308} \\
			\bottomrule
		\end{tabular}%
	}
	\label{tab:opt-comp}%
\end{table}%

Given the computational complexity of $O(n^{3/2})$ brought by structured matrices, we extended our experiments to ultra-long prediction horizons beyond 720. Table \ref{Tab:1024} presents the prediction results of Informer, TimeXer and their enhanced version on the ETTh1 and ETTh2 datasets with horizons of 1024 and 2048. On a horizon of 1024, the enhanced Informer and TimeXer models consistently outperformed their original versions. For a horizon of 2048, the original Informer encountered a memory overflow issue, while the enhanced Informer successfully generated predictions. Similarly, enhanced TimeXer exhibited better results than the original TimeXer, demonstrating the advantages of structured matrix enhancements for ultra-long sequence prediction tasks.
\begin{table}[h]
	\centering
	\caption{The prediction results on the length of the ultra-long sequence. \emph{Informer$^\star$} and \emph{TimeXer$^\star$} represents models enhanced with structured matrices. \textbf{Bold} indicates a better result. "-" indicates the result cannot be obtained due to memory overflow.}
	\begin{tabular}{c|c|cc|cc|cc|cc}
		\toprule
		\multicolumn{2}{c}{} & \multicolumn{2}{c}{Informer} & \multicolumn{2}{c}{Informer$^\star$} & \multicolumn{2}{c}{TimeXer} & \multicolumn{2}{c}{TimeXer$^\star$} \\
		\cmidrule(lr){3-4} \cmidrule(lr){5-6} \cmidrule(lr){7-8} \cmidrule(lr){9-10}
		Datasets & Metrics & MSE & MAE & MSE & MAE & MSE & MAE & MSE & MAE \\
		\midrule
		\multirow{2}[1]{*}{ETTh1} & 1024 & 1.284 & 0.907 & \textbf{1.269} & \textbf{0.931} & 0.504 & 0.490 & \textbf{0.495} & \textbf{0.477} \\
		& 2048 & - & - & \textbf{1.278} & \textbf{0.909} & 0.535 & 0.507 & \textbf{0.515} & \textbf{0.487} \\
		\midrule
		\multirow{2}[1]{*}{ETTh2} & 1024 & 3.408 & 1.584 & \textbf{3.395} & \textbf{1.581} & 0.440 & 0.450 & \textbf{0.425} & \textbf{0.446} \\
		& 2048 & - & - & \textbf{2.689} & \textbf{1.285} & 0.515 & 0.487 & \textbf{0.450} & \textbf{0.455} \\
		\bottomrule
	\end{tabular}
	
	\label{Tab:1024}
\end{table}

\section{Conclusion, Discussion and Future Works}\label{sec:conclusions}
\subsection{Conclusion}
In this paper, our study introduces a groundbreaking structural innovation for Transformer-based models tailored specifically for time series tasks. By designing the Surrogate Attention Block and Surrogate FFN Block built upon structured matrices, we endeavor to elevate efficiency without compromising on model performance. Crucially, we establish the Surrogate Attention Block's equivalence to the conventional self-attention mechanism in terms of expressiveness and trainability, underscoring its suitability for LSTF tasks. Through comprehensive experimentation across nine Transformer-based models spanning five distinct time series forecasting tasks, our findings showcase a remarkable average performance enhancement of 12.4\%. Simultaneously, our framework achieves a substantial reduction in model size, slashing it by an impressive 61.3\%, achieving the objective defined in Section \ref{subsec:obj}.

\subsection{Discussion}
Current Transformer-based time series forecasting models have undergone substantial optimization to address the computational complexities inherent in self-attention mechanisms. Researchers have pursued diverse strategies to mitigate the quadratic computational complexity that constrains Transformer models' applicability to long sequence forecasting tasks. Early approaches, exemplified by LogTrans and Informer, focused on exploiting the inherent sparsity and locality within attention scoring matrices, pioneering techniques that reduced computational complexity to $O(N\log N)$. This initial wave of optimizations demonstrated the potential for more efficient sequence modeling by strategically sampling and approximating attention computations. More radical approaches, such as Autoformer and TCCT, proposed replacing self-attention layers entirely with convolutional layers, aiming at increasing efficiency.

Our proposed optimization method demonstrates significant versatility across various Transformer-based time series forecasting models. The method shows particular compatibility with models that maintain the core attention mechanism's fundamental structure, including Nonstationary Transformer, iTransformer, PAttn and TimeXer. These models, which preserve the original attention computation paradigm, can directly incorporate our optimization approach with minimal architectural modifications. Models that introduce refined attention mechanisms, such as Informer and Autoformer, which enhance attention through techniques like sparsity and auto-correlation, also prove highly compatible with our approach. These models retain the essential matrix multiplication operations of the query, key and value matrices, allowing seamless integration of our optimization strategy. This compatibility stems from our method's ability to preserve the fundamental computational logic while introducing more efficient matrix operations.

However, our approach encounters meaningful limitations when confronted with models that fundamentally reimagine the attention mechanism. Architectures that entirely replace attention layers with alternative computational paradigms—such as pure convolutional or graph-based approaches—present significant challenges for our optimization method. Particularly notable is the case of Pyraformer, whose pyramid-like structural complexity renders our approach less effective. Experimental results consistently demonstrate suboptimal performance when attempting to apply our method to such fundamentally different architectural designs.

These observations underscore the nuanced landscape of Transformer optimization in time series forecasting. While our method offers a powerful and broadly applicable approach, it is not a universal solution. The effectiveness of our optimization strategy depends critically on maintaining the core computational logic of the original Transformer attention mechanism. As the field continues to evolve, future research may explore more generalized optimization techniques that can bridge these architectural differences more comprehensively. For non-Transformer-based models, our method is not applicable since these models lack an attention layer. Although we recognize the potential for broader applicability, our current work remains focused on Transformer-based architectures.

As indicated by the experimental results in Section \ref{sec:experiments}, Surrogate Attention Block exhibits remarkable efficiency without compromising accuracy, enabling it to handle large-scale time series datasets efficiently. This ability stems from its ability to capture both short-term and long-term dependencies simultaneously and its equivalence with attention mechanisms. As the core component of Surrogate Attention Block, Monarch matrices allow it to capture local dependencies akin to convolutional operations\cite{Fu2023iclr}. However, the Surrogate Attention Block performs better on time series tasks compared to convolutional layers. This is because convolution can only capture local dependencies, whereas the Surrogate Attention Block can also capture global dependencies (i.e. long-term dependencies, as proved in Section \ref{subsec:expresivity}) that are also important in time series tasks.

Through inspecting the experimental results, compared to the other 8 Transformer-based models, the Surrogate Attention Block shows a significant improvement in performance on the Vanilla Transformer. Furthermore, some works replacing attention mechanisms with simpler operators such as convolution, like Autoformer\cite{Wu2021NIPS}, have achieved better performance as well. These observations suggest that the pure self-attention mechanism in the Vanilla Transformer has limited effectiveness in time series compared to the Surrogate Attention Block and other special attention blocks in Transformer-based models. This aligns with the conclusion in \cite{Dong2021icml}. 

\subsection{Future Works}
The intuition derived from the discussion suggests a promising direction for future research that involves delving into the causative capabilities of attention mechanisms in time series forecasting. Examining how these mechanisms perceive causal connections within temporal data could unveil fresh perspectives and enhancements in forecasting precision. Subsequently, efforts can be made to decrease the dependence on self-attention mechanisms in time series forecasting, while crafting more straightforward yet efficient forecasting models that could offer advantages like decreased computational intricacy and enhanced scalability.






\bibliographystyle{elsarticle-num} 
\bibliography{references.bib}
\clearpage

\appendix
\section{Theoretical Proof}\label{appendix:theoretical proof}
\subsection{Proof of Theorem \ref{Theo: diagonal attn=conv}}\label{appendix:diagonal proof}
First, since the heads in the diagonal MHSA layer reach the maximum diagonal pattern measurement value, the attention weights for non-local regions around queries are fixed to $0$. There is
\begin{equation}
	\begin{aligned}
		A^{(h)}_{q,k} = \left\lbrace 
		\begin{array}{ll}
			f^{(q,h)}(q-k) & \left( q-k\right)\in \Delta\\
			0 & otherwise
		\end{array}\right. 
	\end{aligned}
	\nonumber
\end{equation}
where $\Delta = \{-\lfloor \lambda/2 \rfloor, \cdots,\lfloor \lambda/2 \rfloor\}$ contains all the corresponding shifts in the local region with size $\lambda$. $f^{(q,h)}$ is a set of bijective mappings: $f^{(q,h)}: \Delta \rightarrow (0,1)$. For fixed $q$ and $h$, $\sum_{\delta\in \Delta} f^{(q, h)}(\delta)=1$. There is:
\begin{equation}
	\begin{aligned}
		\begin{array}{ll}
			MHSA(\mathbf{X})^D_{q, :} &= \sum_{h\in [H]} \sum_{k=1}^{N}A^{(h)}_{q,k}\mathbf{X}_{k,:}W^{(h)}\\
			&= \sum_{h\in [H]} \sum_{\delta\in \Delta}f^{(q,h)}(\delta)\mathbf{X}_{q-\delta,:}W^{(h)}\\
			&= \sum_{h\in [H]} \left( \sum_{\delta\in \Delta}\mathbf{X}_{q-\delta,:}W_{\delta}^{(q,h)}\right)
		\end{array}
	\end{aligned}
	\nonumber
\end{equation}
where $MHSA(\mathbf{X})^D$ is utilized to represent the output of the diagonal MHSA layer and $W_{\delta}^{(q,h)}=diag\left(f^{(q,h)}(\delta)\right) W^{(h)}$ . We can observe that the expression inside the parentheses has the same form as the expression for convolution. Therefore, 
\begin{equation}
	\begin{aligned}
		MHSA(\mathbf{X})^D_{q, :} = \sum_{h\in [H]} Conv^{h}(\mathbf{X})_{q,:}
	\end{aligned}
	\nonumber
\end{equation}
\subsection{Proof of Theorem \ref{Theo: vertical attn=conv}}\label{appendix:vertical proof}

For the $h$-th head, let $\widetilde{K}^{(h)} = \left\lbrace \widetilde{k}_1^{(h)}, \cdots, \widetilde{k}_\lambda^{(h)}\right\rbrace $. For any $q\in [N]$, let $\phi^{(q,h)}$ be a mapping that $\phi^{(q,h)}: \widetilde{K}^{(h)}\rightarrow (0,1]$,  which satisfies $\sum_{k\in\widetilde{K}^{(h)}}\phi^{(q,h)}(k)=1$. The attention scoring matrix of the $h$-th head in a vertical MHSA layer that reaches the maximum vertical pattern measurement value can be denoted as:
\begin{equation}
	\label{Eq:vertical locality condition}
	\begin{aligned}
		A_{q,k}^{(h)} =
		\left\{\begin{array}{ll}
			\phi^{(q,h)}(k)   & k\in \widetilde{K}^{(h)}
			\\ 0 &  otherwise
		\end{array}\right.
	\end{aligned}
	\nonumber
\end{equation}

Let $f^{(h)}$ be a mapping that $f^{(h)}: \widetilde{K}^{(h)}\rightarrow \Delta$, where $\Delta = \{-\lfloor \lambda/2 \rfloor, \cdots,\lfloor \lambda/2 \rfloor\}$. For any $q\in [N]$, let $\varphi^{(q,h)}$ be a mapping that $\varphi^{(q,h)}: \Delta\rightarrow (0,1]$,  which satisfies $\sum_{\delta\in\Delta}\varphi^{(q,h)}(\delta)=1$. Eq.(\ref{Eq:vertical locality condition}) can be rewritten as:
\begin{equation}
	\begin{aligned}
		A_{q,k}^{(h)} =
		\left\{\begin{array}{ll}
			\varphi^{(q,h)}(q-k)   & q-k\in f\left( \widetilde{K}^{(h)}\right) = \Delta
			\\ 0 &  otherwise
		\end{array}\right.
	\end{aligned}
	\nonumber
\end{equation}

Thus,
\begin{equation}
	\label{Eq:vertical multi-head self-attention h}
	\begin{aligned}
		\begin{array}{ll}
			MHSA(\mathbf{X})_{q,:}^V &= \sum_{h \in [H]} \left( \sum_{k \in [N]}A_{q,k}^{(h)} \mathbf{X}_{k,:}\right) W^{(h)} \\
			&= \sum_{h \in [H]} \left( \sum_{\delta \in \Delta} \varphi^{(q,h)}(\delta) \mathbf{X}_{q-\delta,:}\right) W^{(h)} \\
			&= \sum_{h \in [H]} \left( \sum_{\delta \in \Delta} \mathbf{X}_{q-\delta,:}W^{(q,h)}_{\delta}\right) \\
		\end{array}
	\end{aligned}
	\nonumber
\end{equation}
where $MHSA(\mathbf{X})^V$ is utilized to represent the output of the vertical MHSA layer and $W_{\delta}^{(q,h)}=diag\left(\varphi^{(q,h)}(\delta)\right) W^{(h)}$ . We can observe that the expression inside the parentheses has the same form as the expression for convolution. Therefore, 
\begin{equation}
	\begin{aligned}
		MHSA(\mathbf{X})^V_{q, :} = \sum_{h\in [H]} Conv^{h}(\mathbf{X})_{q,:}
	\end{aligned}
	\nonumber
\end{equation}


\subsection{Stationary Analysis}\label{appendix:LTI}
In this section, we will prove that the Surrogate Attention Block is a linear time-invariant system, which can be described by the following equations:
\begin{equation}
	\begin{aligned}
		x_{t+1} &= \mathbf{A}x_t + \mathbf{B}u_{t+1} \\
		y_{t+1} &= \mathbf{C}x_{t+1} + \mathbf{D}u_{t+1}
	\end{aligned}
	\nonumber
\end{equation}

For a time series task, $x_t$ is the input at time step $t$, which is mapped to $q_t,k_t,v_t$ by linear projection. From Eq.(\ref{Eq:monarch-attn}), we can derive the output $y_t$ as:
\begin{equation}
	\begin{aligned}
		y_t =  \begin{bmatrix}m^{2}_{t,1} & m^{2}_{t,2} & \dots & m^{2}_{t,N}\end{bmatrix}\times \left( \begin{bmatrix}
			k_1 m^{1}_{1,1} & k_1 m^{1}_{1,2} & \dots & k_1 m^{1}_{1,N} \\
			k_2 m^{1}_{2,1} & k_2 m^{1}_{2,2} & \dots & k_2 m^{1}_{2,N} \\
			\vdots & \vdots & \ddots & \vdots \\
			k_N m^{1}_{N,1} & k_N m^{1}_{N,2} & \dots & k_N m^{1}_{N,N} \\
		\end{bmatrix}\times
		\begin{bmatrix}q_1\\q_2\\ \vdots\\ q_{N}\end{bmatrix}\right) \odot v_t
	\end{aligned}
	\nonumber
\end{equation}

This is clearly a time-varying system and to relate our approach to time-invariant systems, we treat $\mathbf{M}^2_t=\begin{bmatrix}m^{2}_{t,1} & m^{2}_{t,2} & \dots & m^{2}_{t,N}\end{bmatrix}$ as a post-processing step, facilitating the identification of the linear time-invariant components.

By redefining matrices \(\mathbf{A}\), \(\mathbf{B}\), \(\mathbf{C}\) and \(\mathbf{D}\), we observe the system in a new light:
\begin{equation}
	\begin{aligned}
		\mathbf{A}=0, \mathbf{B}=\mathbf{I},\mathbf{D}=0 
	\end{aligned}
	\nonumber
\end{equation}
\begin{equation}
	\begin{aligned}
		\mathbf{C}=\begin{bmatrix}
			k_1 m^{1}_{1,1} & k_1 m^{1}_{1,2} & \dots & k_1 m^{1}_{1,N} \\
			k_2 m^{1}_{2,1} & k_2 m^{1}_{2,2} & \dots & k_2 m^{1}_{2,N} \\
			\vdots & \vdots & \ddots & \vdots \\
			k_N m^{1}_{N,1} & k_N m^{1}_{N,2} & \dots & k_N m^{1}_{N,N} \\
		\end{bmatrix}\times
		\begin{bmatrix}q_1\\q_2\\ \vdots\\ q_{N}\end{bmatrix}
	\end{aligned}
	\nonumber
\end{equation}

This transformation yields a simplified LTI form:
\begin{equation}
	\begin{aligned}
		x_{t+1} &= \mathbf{B}v_{t+1}, \\
		y_{t+1}' &= \mathbf{C}x_{t+1}.
	\end{aligned}
	\nonumber
\end{equation}

In this context, \(y_{t+1}'\) represents the transformed output without the influence of $\mathbf{M}^2_t$ and it serves as a basis for subsequent post-processing. This separation allows for a comprehensive analysis of the LTI properties of the system. We proceed to apply the post-processing step by multiplying \(y_{t+1}'\) with $\mathbf{M}^2_t$, facilitating the extraction of temporal features and providing insights into the system's behavior. This decomposition and reformulation underscore the linear time-invariant characteristics of the system, enhancing our understanding of its stability and behavior over time.

\subsection{Expressiveness Analysis}\label{appendix:expresivity}

Let $\mathbf{L}^1$ , $\mathbf{R}^1$ and $A$ be as follows:
$$
\mathbf{L}^1 = \begin{bmatrix}
	l^1_{[0,0]}, l^1_{[0,1]} , \cdots , l^1_{[0,\sqrt{N}-1 ]} , \cdots \\
	l^1_{[1,0]}, l^1_{[1,1]} , \cdots , l^1_{[1,\sqrt{N}-1 ]} , \cdots \\
	\vdots \\
	l^1_{[\sqrt{N}-1,0]}, l^1_{[\sqrt{N}-1,1]} , \cdots , l^1_{[\sqrt{N}-1,\sqrt{N}-1 ]} , \cdots\\
	\cdots,l^1_{[\sqrt{N},\sqrt{N}]}, l^1_{[\sqrt{N},\sqrt{N}+1]} , \cdots , l^1_{[\sqrt{N},2\sqrt{N}-1 ]} , \cdots\\
	\cdots,l^1_{[\sqrt{N}+1,\sqrt{N}]}, l^1_{[\sqrt{N}+1,\sqrt{N}+1]} , \cdots , l^1_{[\sqrt{N}+1,2\sqrt{N}-1 ]} , \cdots\\
	\vdots \\
	\cdots,l^1_{[2\sqrt{N}-1,\sqrt{N}]}, l^1_{[2\sqrt{N}-1,\sqrt{N}+1]} , \cdots , l^1_{[2\sqrt{N}-1,2\sqrt{N}-1 ]} , \cdots\\
	\vdots \\
	\cdots,l^1_{[N-1-\sqrt{N},N-1-\sqrt{N}]}, l^1_{[N-1-\sqrt{N},N+\sqrt{N}]} , \cdots , l^1_{[N-1-\sqrt{N},N-1 ]}\\
	\cdots,l^1_{[N-\sqrt{N},N-1-\sqrt{N}]}, l^1_{[N-\sqrt{N},N+\sqrt{N}]} , \cdots , l^1_{[N-\sqrt{N},N-1 ]}\\
	\vdots \\
	\cdots,l^1_{[N-1,N-1-\sqrt{N}]}, l^1_{[N-1,N+\sqrt{N}]} , \cdots , l^1_{[N-1,N-1 ]}\\
\end{bmatrix}
$$

$$
\mathbf{R}^1 = \begin{bmatrix}
	r^1_{[0,0]}, r^1_{[0,1]} , \cdots , r^1_{[0,\sqrt{N}-1 ]} , \cdots \\
	r^1_{[1,0]}, r^1_{[1,1]} , \cdots , r^1_{[1,\sqrt{N}-1 ]} , \cdots \\
	\vdots \\
	r^1_{[\sqrt{N}-1,0]}, r^1_{[\sqrt{N}-1,1]} , \cdots , r^1_{[\sqrt{N}-1,\sqrt{N}-1 ]} , \cdots\\
	\cdots,r^1_{[\sqrt{N},\sqrt{N}]}, r^1_{[\sqrt{N},\sqrt{N}+1]} , \cdots , r^1_{[\sqrt{N},2\sqrt{N}-1 ]} , \cdots\\
	\cdots,r^1_{[\sqrt{N}+1,\sqrt{N}]}, r^1_{[\sqrt{N}+1,\sqrt{N}+1]} , \cdots , r^1_{[\sqrt{N}+1,2\sqrt{N}-1 ]} , \cdots\\
	\vdots \\
	\cdots,r^1_{[2\sqrt{N}-1,\sqrt{N}]}, r^1_{[2\sqrt{N}-1,\sqrt{N}+1]} , \cdots , r^1_{[2\sqrt{N}-1,2\sqrt{N}-1 ]} , \cdots\\
	\vdots \\
	\cdots,r^1_{[N-1-\sqrt{N},N-1-\sqrt{N}]}, r^1_{[N-1-\sqrt{N},N+\sqrt{N}]} , \cdots , r^1_{[N-1-\sqrt{N},N-1 ]}\\
	\cdots,r^1_{[N-\sqrt{N},N-1-\sqrt{N}]}, r^1_{[N-\sqrt{N},N+\sqrt{N}]} , \cdots , r^1_{[N-\sqrt{N},N-1 ]}\\
	\vdots \\
	\cdots,r^1_{[N-1,N-1-\sqrt{N}]}, r^1_{[N-1,N+\sqrt{N}]} , \cdots , r^1_{[N-1,N-1 ]}\\
\end{bmatrix}
$$

$$
A=\mathbf{M}^1Q\odot K=\begin{bmatrix}
	a_0\\
	a_1\\
	\vdots \\
	a_{N-1}
\end{bmatrix}
$$
where $\mathbf{R}^1$ and $\mathbf{L}^1$ are block diagonal matrices of size $N \times N$, with each block having dimensions $\sqrt{N} \times \sqrt{N}$.  Let $b$ and $c$ represent the abscissa and ordinate of the elements in the diagonal blocks of $\mathbf{R}^1$ and $\mathbf{L}^1$. If $\left \lfloor b/\sqrt{N}  \right \rfloor \ne \left \lfloor c/\sqrt{N}  \right \rfloor$, then $\mathbf{R}^1_{[b, c]} = \mathbf{L}^1_{[b, c]}=0$. 

It is important to note that the superscript in the upper right corner of the matrix or element denotes its identification, while the subscript in the lower right corner indicates the element's index within the matrix.  For example, $m^2_{[0,0]}$ signifies the element located at coordinates$[0, 0]$ within $\mathbf{M}^2$ , rather than the square of $m_{[0, 0]}$.

Given the following matrix definitions:
$$\mathbf{M}^1=P\mathbf{L}^1P\mathbf{R}^1P, \mathbf{M}^2=P\mathbf{L}^2P\mathbf{R}^2P$$
where $P$ is a permutation matrix. Let $h(i) =\left \lfloor i/\sqrt{N}  \right \rfloor  + \sqrt{N} (i \% \sqrt{N} )$, The subsequent calculation process follows this redefinition, illustrating the transformation of these matrices based on these block structures. The detailed process is as follows:
$$
P\mathbf{L}^1 = \begin{bmatrix}
	l^1_{[0,0]}, l^1_{[0,1]} , \cdots , l^1_{[0,\sqrt{N}-1 ]} , \cdots \\
	l^1_{[\sqrt{N},0]}, l^1_{[\sqrt{N},1]} , \cdots , l^1_{[\sqrt{N},\sqrt{N}-1 ]} , \cdots \\
	\vdots \\
	l^1_{[\sqrt{N}(\sqrt{N}-1),0]}, l^1_{[\sqrt{N}(\sqrt{N}-1),1]} , \cdots , l^1_{[\sqrt{N}(\sqrt{N}-1),\sqrt{N}-1 ]} , \cdots\\
	\cdots,l^1_{[1,\sqrt{N}]}, l^1_{[1,\sqrt{N}+1]} , \cdots , l^1_{[1,2\sqrt{N}-1 ]} , \cdots\\
	\cdots,l^1_{[1+\sqrt{N},\sqrt{N}]}, l^1_{[1+\sqrt{N},\sqrt{N}+1]} , \cdots , l^1_{[1+\sqrt{N},2\sqrt{N}-1 ]} , \cdots\\
	\vdots \\
	\cdots,l^1_{[1+\sqrt{N}(\sqrt{N}-1),\sqrt{N}]}, l^1_{[1+\sqrt{N}(\sqrt{N}-1),\sqrt{N}+1]} , \cdots , l^1_{[1+\sqrt{N}(\sqrt{N}-1),2\sqrt{N}-1 ]} , \cdots\\
	\vdots \\
	\cdots,l^1_{[\sqrt{N}-1,N-1-\sqrt{N}]}, l^1_{[\sqrt{N}-1,N+\sqrt{N}]} , \cdots , l^1_{[\sqrt{N}-1,N-1 ]}\\
	\cdots,l^1_{[2\sqrt{N}-1,N-1-\sqrt{N}]}, l^1_{[2\sqrt{N}-1,N+\sqrt{N}]} , \cdots , l^1_{[2\sqrt{N}-1,N-1 ]}\\
	\vdots \\
	\cdots,l^1_{[N-1,N-1-\sqrt{N}]}, l^1_{[N-1,N+\sqrt{N}]} , \cdots , l^1_{[N-1,N-1 ]}\\
\end{bmatrix}
$$

$$
P\mathbf{L}^1P = \begin{bmatrix}
	l^1_{[0,0]}, l^1_{[0,\sqrt{N}]} , \cdots ,l^1_{[0,\sqrt{N}(\sqrt{N}-1)]},\cdots \\
	l^1_{[\sqrt{N},0]}, l^1_{[\sqrt{N},\sqrt{N}]} , \cdots , l^1_{[\sqrt{N},\sqrt{N}(\sqrt{N}-1) ]} \cdots \\
	\vdots \\
	l^1_{[\sqrt{N}(\sqrt{N}-1),0]}, l^1_{[\sqrt{N}(\sqrt{N}-1),\sqrt{N}]} , \cdots , l^1_{[\sqrt{N}(\sqrt{N}-1),\sqrt{N}(\sqrt{N}-1) ]} , \cdots\\
	\cdots,l^1_{[1,1]}, l^1_{[1,1+\sqrt{N}]} , \cdots , l^1_{[1,1+\sqrt{N}(\sqrt{N}-1) ]}  ,\cdots\\
	\cdots,l^1_{[1+\sqrt{N},1]}, l^1_{[1+\sqrt{N},1+\sqrt{N}]} , \cdots , l^1_{[1+\sqrt{N},1+\sqrt{N}(\sqrt{N}-1) ]} ,  \cdots\\
	\vdots \\
	\cdots,l^1_{[1+\sqrt{N}(\sqrt{N}-1),1]}, l^1_{[1+\sqrt{N}(\sqrt{N}-1),1+\sqrt{N}]},  \cdots ,l^1_{[1+\sqrt{N}(\sqrt{N}-1),1+\sqrt{N}(\sqrt{N}-1)]}, \cdots\\
	\vdots \\
	\cdots,l^1_{[\sqrt{N}-1,\sqrt{N}-1]}, l^1_{[\sqrt{N}-1,2\sqrt{N}-1]} , \cdots , l^1_{[\sqrt{N}-1,N-1 ]}\\
	\cdots,l^1_{[2\sqrt{N}-1,\sqrt{N}-1]}, l^1_{[2\sqrt{N}-01,2\sqrt{N}-1]]} , \cdots , l^1_{[2\sqrt{N}-1,N-1 ]}\\
	\vdots \\
	\cdots,l^1_{[N-1,\sqrt{N}-1]}, l^1_{[N-1,2\sqrt{N}-1]]} , \cdots , l^1_{[N-1,N-1 ]}\\
\end{bmatrix}
$$

$$
P\mathbf{L}^1P\mathbf{R}^1 = \begin{bmatrix}
	\sum_{i=0}^{N-1}l^1_{[0,h(i)]}r^1_{[i,0]},\sum_{i=0}^{N-1}l^1_{[0,h(i)]}r^1_{[i,1]},\cdots   \\
	\sum_{i=0}^{N-1}l^1_{[\sqrt{N},h(i)]}r^1_{[i,0]},\sum_{i=0}^{N-1}l^1_{[\sqrt{N},h(i)]}r^1_{[i,1]},  \cdots \\
	\vdots \\
	\sum_{i=0}^{N-1}l^1_{[\sqrt{N}(\sqrt{N}-1),h(i) ]}r^1_{[i,0]},\sum_{i=0}^{N-1}l^1_{[\sqrt{N}(\sqrt{N}-1),h(i)]}r^1_{[i,1]},  \cdots \\
	\vdots \\
\end{bmatrix}
$$

$$
\mathbf{M}^1=P\mathbf{L}^1P\mathbf{R}^1P = \begin{bmatrix}
	\sum_{i=0}^{N-1}l^1_{[0,h(i) ]}r^1_{[i,0]},\sum_{i=0}^{N-1}l^1_{[0,h(i)]}r^1_{[i,\sqrt{N}]},\cdots   \\
	\sum_{i=0}^{N-1}l^1_{[\sqrt{N},h(i)]}r^1_{[i,0]},\sum_{i=0}^{N-1}l^1_{[\sqrt{N},h(i)]}r^1_{[i,\sqrt{N}]},  \cdots \\
	\vdots \\
	\sum_{i=0}^{N-1}l^1_{[\sqrt{N}(\sqrt{N}-1),h(i)]}r^1_{[i,0]},\sum_{i=0}^{N-1}l^1_{[\sqrt{N}(\sqrt{N}-1),h(i)]}r^1_{[i,\sqrt{N}]},  \cdots \\
	\vdots \\
\end{bmatrix}
$$

$\mathbf{M}^2$ can be obtained in the same way:
$$
\mathbf{M}^2 = \begin{bmatrix}
	\sum_{i=0}^{N-1}l^2_{[0,h(i)]}r^2_{[i,0]},\sum_{i=0}^{N-1}l^2_{[0,h(i)]}r^2_{[i,\sqrt{N}]},\cdots   \\
	\sum_{i=0}^{N-1}l^2_{[\sqrt{N},h(i)]}r^2_{[i,0]},\sum_{i=0}^{N-1}l^2_{[\sqrt{N},h(i)]}r^2_{[i,\sqrt{N}]},  \cdots \\
	\vdots \\
	\sum_{i=0}^{N-1}l^2_{[\sqrt{N}(\sqrt{N}-1),h(i)]}r^2_{[i,0]},\sum_{i=0}^{N-1}l^2_{[\sqrt{N}(\sqrt{N}-1),h(i)]}r^2_{[i,\sqrt{N}]},  \cdots \\
	\vdots \\
\end{bmatrix}
=\begin{bmatrix}
	m^2_{[0,0]},m^2_{[0,1]}\cdots   \\
	m^2_{[1,0]},m^2_{[1,1]}\cdots\\
	\vdots \\
	m^2_{[\sqrt{N}-1,0]},m^2_{[\sqrt{N}-1,1]}\cdots \\
	\vdots \\
\end{bmatrix}
$$

$$
\mathbf{M}^1Q\odot K = \begin{bmatrix}
	(\sum_{i=0}^{N-1}l^1_{[0,h(i) ]}r^1_{[i,0]}q_0+\sum_{i=0}^{N-1}l^1_{[0,h(i)]}r^1_{[i,\sqrt{N}]}q_1+\cdots)k_0  \\
	(\sum_{i=0}^{N-1}l^1_{[\sqrt{N},h(i)]}r^1_{[i,0]}q_0+\sum_{i=0}^{N-1}l^1_{[\sqrt{N},h(i)]}r^1_{[i,\sqrt{N}]}q_1+\cdots)k_1   \\
	\vdots \\
	(\sum_{i=0}^{N-1}l^1_{[\sqrt{N}(\sqrt{N}-1),h(i)]}r^1_{[i,0]}q_0+\sum_{i=0}^{N-1}l^1_{[\sqrt{N}(\sqrt{N}-1),h(i)]}r^1_{[i,\sqrt{N}]}q_1+\cdots)k_{\sqrt{N}-1}   \\
	\vdots \\
\end{bmatrix}
$$

$$
\mathbf{M}^2(\mathbf{M}^1Q\odot K)=\mathbf{M}^2 A=\begin{bmatrix}
	\sum_{i=0}^{N-1}m^2_{[0,i]}a_i \\
	\sum_{i=0}^{N-1}m^2_{[1,i]}a_i \\
	\vdots \\
	\sum_{i=0}^{N-1}m^2_{[\sqrt{N}-1 ,i]}a_i \\
	\vdots \\
\end{bmatrix}
$$

$$
\mathbf{Y} = \mathbf{M}^2(\mathbf{M}^1Q\odot K)\odot V=\begin{bmatrix}
	\sum_{j=0}^{N-1}m^2_{[0,j]}a_jv_0 \\
	\sum_{j=0}^{N-1}m^2_{[1,j]}a_jv_1 \\
	\vdots \\
	\sum_{j=0}^{N-1}m^2_{[\sqrt{N}-1 ,j]}a_jv_{\sqrt{N}-1} \\
	\vdots \\
\end{bmatrix}
=
\begin{bmatrix}
	y_0 \\
	y_1 \\
	\vdots \\
	y_{\sqrt{N}-1} \\
	\vdots \\
\end{bmatrix}
$$

So we can get
\begin{equation}
	\begin{aligned}
		& y_{k} = \sum_{j=0}^{N-1}m^2_{[k ,j]}a_jv_{k}=v_k(m^2_{[k ,0]}a_0+\cdots+m^2_{[k ,k-1]}a_{k-1}+\cdots) \\
		& =v_k \{    [(\sum_{i=0}^{N-1}l^2_{[h(k),h(i)]}r^2_{[i,0]})(\sum_{i=0}^{N-1}l^1_{[0,h(i)]}r^1_{[i,0]}q_0+\cdots)k_0]+\cdots \\
		& +[(\sum_{i=0}^{N-1}l^2_{[h(k),h(i)]}r^2_{[i,h(k-1)]})
		(\sum_{i=0}^{N-1}l^1_{[h(k-1),h(i)]}r^1_{[i,0]}q_0+\cdots+\sum_{i=0}^{N-1}l^1_{[h(k-1),h(i)]}r^1_{[i,h(k-1)]}q_{k-1}+\cdots)k_{k-1}]
		+\cdots  \}
	\end{aligned}
	\nonumber
\end{equation}

To evaluate the behavior of the Surrogate Attention Block in terms of capturing long-term and short-term dependencies, we can ignore the projections of queries, keys and values for simplicity.

Let’s consider a case where specific elements within matrices are set to 1, while others are set to 0, allowing us to isolate specific correlations. Define these conditions for the Surrogate Attention Block:

\begin{enumerate}
	\item \textit{Long-Term Dependencies}: Let $l^2_{[h(k),h(i)]}=r^2_{[i,0]}=l^1_{[0,h(i)]}=r^1_{[i,0]}=1$, with all other elements set to 0. Under these conditions, if we calculate:
	$$
	\sum_{i=0}^{N-1}l^2_{[h(k),h(i)]}r^2_{[i,h(k-1)]}=\sum_{i=0}^{N-1}l^1_{[0,h(i)]}r^1_{[i,0]}=1
	$$
	
	It yields that the output at time step $k$, denoted by $y_k$,  is equivalent to $x_kx^2_{0}$. This indicates that the attention mechanism strongly correlates time step $k$ with time step $0$. thereby validating that the Surrogate Attention Block can effectively capture long-term dependencies.
	\item \textit{Short-Term Dependencies}: Similarly, let $l^2_{[h(k),h(i)]}=r^2_{[i,h(k-1)]}=l^1_{[h(k-1),h(i)]}=r^1_{[i,h(k-1)]}=1$, with other elements set to $0$. If we calculate:
	$$
	\sum_{i=0}^{N-1}l^2_{[h(k),h(i)]}r^2_{[i,h(k-1)]}=\sum_{i=0}^{N-1}l^1_{[h(k-1),h(i)]}r^1_{[i,h(k-1)]}=1
	$$
	
	It leads to the result that $y_k=x_kx^2_{k-1}$. This configuration confirms that the attention mechanism creates a strong correlation between time step $k$ and time step $k-1$, suggesting that the Surrogate Attention Block is also capable of learning short-term dependencies.
\end{enumerate}
\clearpage

\section{Experiment details}\label{appendix:exp}
\subsection{Metrics}\label{appendix:exp-metrics}
\begin{itemize}
	\item \textbf{MAE (Mean Absolute Error)} measures the average absolute difference between the actual values and predicted values. It provides a straightforward and interpretable measure of prediction accuracy. A lower MAE indicates that the model tends to make predictions that are closer to the actual values in magnitude. Formula:
	\begin{equation}
		\begin{aligned}
			MAE = \frac{1}{n}\sum_{i=1}^{n}|y_i-\hat{y}_i|
		\end{aligned}
		\nonumber
	\end{equation}
	
	\item \textbf{MSE (Mean Squared Error)} calculates the average of the squared differences between actual and predicted values. MSE is sensitive to large errors and penalizes them more than MAE. It is commonly used for its mathematical tractability and suitability for optimization algorithms. Formula:
	\begin{equation}
		\begin{aligned}
			MSE = \frac{1}{n}\sum_{i=1}^{n}(y_i-\hat{y}_i)^2
		\end{aligned}
		\nonumber
	\end{equation}
	
	\item \textbf{R-Square (Coefficient of Determination)} evaluates the goodness of fit of a model to the data. It quantifies the proportion of the variance in the dependent variable that is explained by the model. R-Square values range from 0 to 1, with higher values indicating that the model captures a larger portion of the variation in the data. It helps assess how well the model represents the underlying data patterns. Formula:
	\begin{equation}
		\begin{aligned}
			R2 = 1 - \frac{\sum_{i=1}^{n}(y_i - \hat{y}_i)^2}{\sum_{i=1}^{n}(y_i - \bar{y})^2}
		\end{aligned}
		\nonumber
	\end{equation}
	
	\item \textbf{DTW (Dynamic Time Wraping)} is a method used for comparing two time series with potentially different lengths and time axes. It determines the optimal alignment of elements in the two series, minimizing their paired distances. Therefore, DTW can be used for measuring the waveform similarity between two time series. DTW calculates the alignment between two time series by finding the optimal path through a cost matrix. The optimal path $P$ is determined by minimizing the accumulated cost:
	\begin{equation} 
		\begin{aligned}
			DTW = \min_{P} \sqrt{\sum_{(i, j) \in P} |y_i-\hat{y}_i|}
		\end{aligned}
		\nonumber
	\end{equation}
	
	\item \textbf{SMAPE (Symmetric Mean Absolute Percentage Error)} is a symmetric percentage-based error metric widely used in short-term time series forecasting. It measures the percentage difference between the actual and predicted values, accommodating situations where the scale of the data varies. Formula:
	\begin{equation} 
		\begin{aligned}
			SMAPE = \frac{1}{n}\sum_{i=1}^{n}\frac{|y_i - \hat{y}_i|}{(|y_i| + |\hat{y}_i|)/2} \times 100
		\end{aligned}
		\nonumber
	\end{equation}
	
	\item \textbf{MASE (Mean Absolute Scaled Error)} is a scale-independent metric that evaluates the accuracy of a forecasting model. It compares the mean absolute error of the model to the mean absolute error of a naïve forecast, providing a standardized measure of performance. Formula:
	\begin{equation} 
		\begin{aligned}
			MASE = \frac{1}{n}\sum_{i=1}^{n}\frac{|y_i - \hat{y}_i|}{\frac{1}{n-1}\sum_{i=2}^{n}|y_i - y_{i-1}|}
		\end{aligned}
		\nonumber
	\end{equation}

	\item \textbf{OWA (Overall Weighted Average)} is a special metric used in M4 competition. Formula:
	\begin{equation} 
		\begin{aligned}
			OWA =\frac{1}{2}\left( \frac{SMAPE}{SMAPE_{Naive}} + \frac{MASE}{MASE_{Naive}} \right) 
		\end{aligned}
		\nonumber
	\end{equation}
	
	\item \textbf{Accuracy} is a classification metric that measures the proportion of correct predictions out of the total predictions made by a model. Formula:
	\begin{equation} 
		\begin{aligned}
			Accuracy = \frac{\text{Number of Correct Predictions}}{\text{Total Number of Predictions}} 
		\end{aligned}
		\nonumber
	\end{equation}
	
	\item \textbf{Precision} assesses the accuracy of positive predictions. It is the ratio of true positive predictions to the total number of positive predictions made by the model. Formula:
	\begin{equation} 
		\begin{aligned}
			Precision = \frac{\text{True Positives}}{\text{True Positives + False Positives}}
		\end{aligned}
		\nonumber
	\end{equation}
	
	\item \textbf{Recall (Sensitivity or True Positive Rate)} measures the ability of a model to identify all relevant instances. It is the ratio of true positive predictions to the total number of actual positive instances. Formula:
	\begin{equation} 
		\begin{aligned}
			Recall = \frac{\text{True Positives}}{\text{True Positives + False Negatives}}
		\end{aligned}
		\nonumber
	\end{equation}
	
	\item \textbf{F1-Score} is the harmonic mean of precision and recall. It provides a balance between precision and recall, making it suitable for situations where there is an uneven class distribution. Formula:
	\begin{equation} 
		\begin{aligned}
			F1-Score = 2 \times \frac{\text{Precision} \times \text{Recall}}{\text{Precision} + \text{Recall}}
		\end{aligned}
		\nonumber
	\end{equation}
	
\end{itemize}
\clearpage

\subsection{Model Details}\label{appendix:detail}
This section will show the details of all 9 models of Transformer-based models improved with structured matrices. \textbf{Bold} indicates sections that have been replaced.
\definecolor{mypink}{RGB}{248, 206, 204}
\definecolor{myred}{RGB}{184, 84, 80}

\begin{table}[!h]
	\centering
	\caption{Details of the improved \underline{Vanilla Transformer} with structured matrices.}
	{\tiny \begin{tabular}{lccc|lccccc|c}
		\toprule
		\textbf{Encoder:} &       &       & \multicolumn{1}{c}{} & \multicolumn{1}{c}{} &       &       &       &       & \multicolumn{1}{c}{} & \multicolumn{1}{c}{N} \\
		\midrule
		\multicolumn{4}{c|}{Inputs} & \multicolumn{2}{p{8.08em}|}{1 × 3 Conv1d} & \multicolumn{4}{c|}{Embedding(d = 512)} &  \\
		\midrule
		\multicolumn{4}{c|}{\multirow{4}[8]{*}{Self-attention Block}} & \multicolumn{6}{c|}{\textbf{Surrogate Attention Block} (h = 8, d = 64)} & \multirow{4}[8]{*}{2} \\
		\cmidrule{5-10}    \multicolumn{4}{c|}{}         & \multicolumn{6}{c|}{Add, LayerNorm, Dropout (p = 0.05)} &  \\
		\cmidrule{5-10}    \multicolumn{4}{c|}{}         & \multicolumn{6}{c|}{\textbf{Surrogate FFN Block}, GELU} &  \\
		\cmidrule{5-10}    \multicolumn{4}{c|}{}         & \multicolumn{6}{c|}{Add, LayerNorm, Dropout (p = 0.05)} &  \\
		\midrule    
		\textbf{Decoder:} &       &       & \multicolumn{1}{c}{} & \multicolumn{1}{c}{} &       &       &       &       & \multicolumn{1}{c}{} & \multicolumn{1}{c}{N} \\
		\midrule
		\multicolumn{4}{c|}{Inputs} & \multicolumn{2}{p{8.08em}|}{1 × 3 Conv1d} & \multicolumn{4}{c|}{Embedding(d = 512)} &  \\
		\midrule
		\multicolumn{4}{c|}{Masked PSB} & \multicolumn{6}{c|}{add Mask on Attention Block} & \multirow{7}[14]{*}{1} \\
		\cmidrule{1-10}    \multicolumn{4}{c|}{\multirow{6}[12]{*}{Self-attention Block}} & \multicolumn{6}{c|}{\textbf{Surrogate Attention Block} (h = 8, d = 64)} &  \\
		\cmidrule{5-10}    \multicolumn{4}{c|}{}         & \multicolumn{6}{c|}{Add, LayerNorm, Dropout (p = 0.05)} &  \\
		\cmidrule{5-10}    \multicolumn{4}{c|}{}         & \multicolumn{6}{c|}{MultiHeadedAttention (h = 8, d = 64)} &  \\
		\cmidrule{5-10}    \multicolumn{4}{c|}{}         & \multicolumn{6}{c|}{Add, LayerNorm, Dropout (p = 0.05)} &  \\
		\cmidrule{5-10}    \multicolumn{4}{c|}{}         & \multicolumn{6}{c|}{\textbf{Surrogate FFN Block}, GELU} &  \\
		\cmidrule{5-10}    \multicolumn{4}{c|}{}         & \multicolumn{6}{c|}{Add, LayerNorm, Dropout (p = 0.05)} &  \\
		\midrule
		\textbf{Final:} &       &       & \multicolumn{1}{c}{} & \multicolumn{1}{c}{} &       &       &       &       & \multicolumn{1}{c}{} &  \\
		\midrule
		\multicolumn{4}{c|}{Outputs} & \multicolumn{6}{c|}{FCN}             &  \\
		\bottomrule
	\end{tabular}}
	
\end{table}%
\begin{figure}[h]
	\centering
	\includegraphics[width=0.5\linewidth]{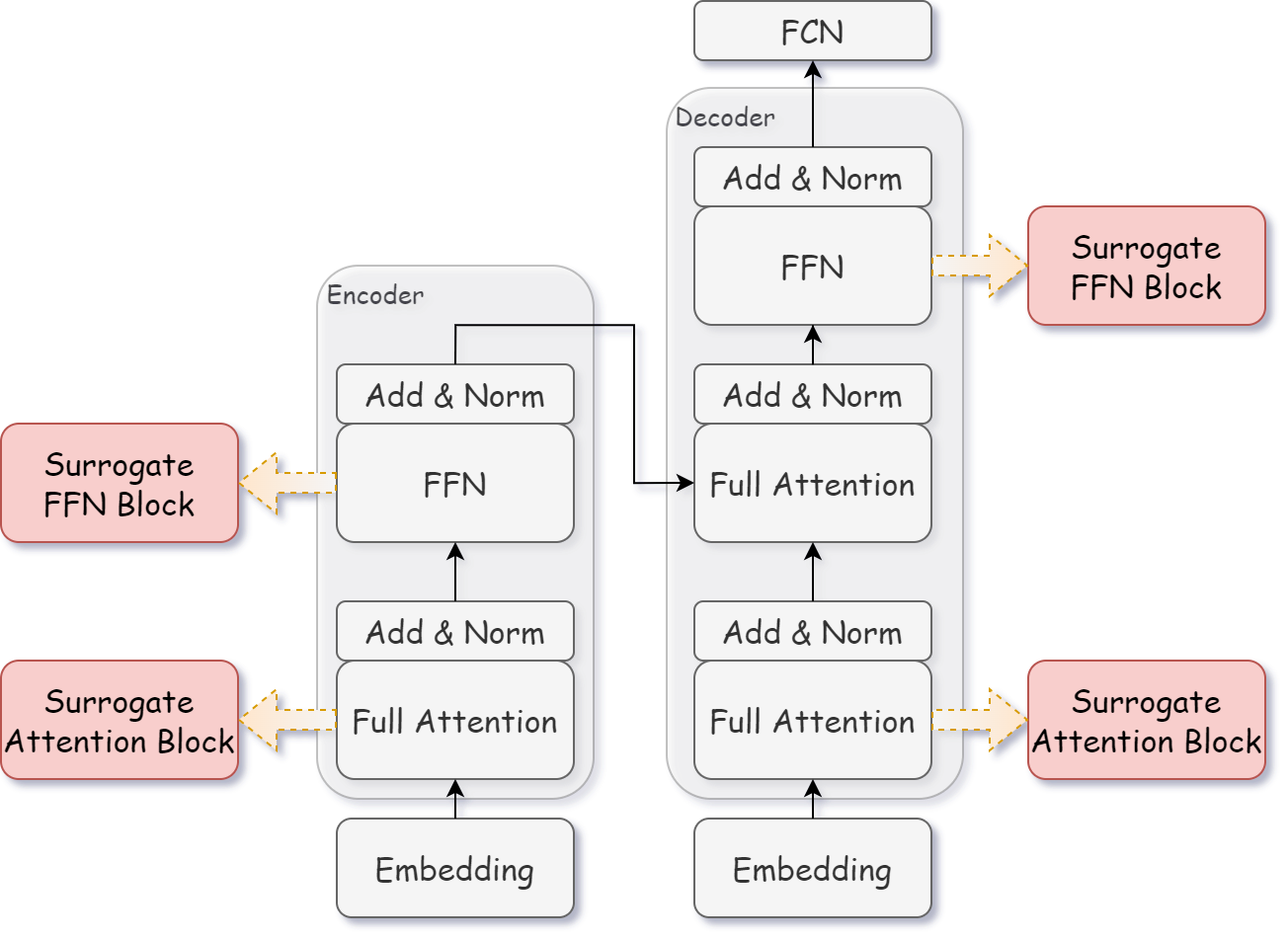}
	\caption{Architecture of the improved \underline{Vanilla Transformer} with structured matrices. The \fcolorbox{myred}{mypink}{red blocks} represent the replaced blocks.}
	\label{fig:xforemr-transformer}
\end{figure}
\clearpage

\begin{table}[htbp]
	\centering
	\caption{Details of the improved \underline{Informer} with structured matrices.}
	{\tiny \begin{tabular}{lccc|lccccc|c}
		\toprule
		\textbf{Encoder:} &       &       & \multicolumn{1}{c}{} & \multicolumn{1}{c}{} &       &       &       &       & \multicolumn{1}{c}{} & \multicolumn{1}{c}{N} \\
		\midrule
		\multicolumn{4}{c|}{Inputs} & \multicolumn{2}{c|}{1 × 3 Conv1d} & \multicolumn{4}{c|}{Embedding(d = 512)} &  \\
		\midrule
		\multicolumn{4}{c|}{\multirow{5}[10]{*}{ProbSparse Self-attention Block}} & \multicolumn{6}{c|}{\textbf{Surrogate Attention Block} (h = 8, d = 64)} & \multirow{7}[14]{*}{2} \\
		\cmidrule{5-10}    \multicolumn{4}{c|}{}         & \multicolumn{6}{c|}{Add, LayerNorm, Dropout (p = 0.05)} &  \\
		\cmidrule{5-10}    \multicolumn{4}{c|}{}         & \multicolumn{6}{c|}{\textbf{Surrogate FFN Block}, GELU} &  \\
		\cmidrule{5-10}    \multicolumn{4}{c|}{}         & \multicolumn{6}{c|}{Add, LayerNorm, Dropout (p = 0.05)} &  \\
		\cmidrule{5-10}    \multicolumn{4}{c|}{}         & \multicolumn{6}{c|}{Series decomposition (moving avg = 25)} &  \\
		\cmidrule{1-10}    \multicolumn{4}{c|}{\multirow{2}[4]{*}{Distilling}} & \multicolumn{6}{c|}{1 × 3 Conv1d, GELU} &  \\
		\cmidrule{5-10}    \multicolumn{4}{c|}{}         & \multicolumn{6}{c|}{Max pooling (stride = 2)} &  \\
		\midrule
		\textbf{Decoder:} &       &       & \multicolumn{1}{c}{} & \multicolumn{1}{c}{} &       &       &       &       & \multicolumn{1}{c}{} & \multicolumn{1}{c}{N} \\
		\midrule
		\multicolumn{4}{c|}{Inputs} & \multicolumn{2}{p{8.08em}|}{1 × 3 Conv1d} & \multicolumn{4}{c|}{Embedding(d = 512)} &  \\
		\midrule
		\multicolumn{4}{c|}{Masked PSB} & \multicolumn{6}{c|}{add Mask on Attention Block} & \multirow{7}[14]{*}{1} \\
		\cmidrule{1-10}    \multicolumn{4}{c|}{\multirow{6}[12]{*}{ProbSparse Self-attention Block}} & \multicolumn{6}{c|}{\textbf{Surrogate Attention Block} (h = 8, d = 64)} &  \\
		\cmidrule{5-10}    \multicolumn{4}{c|}{}         & \multicolumn{6}{c|}{Add, LayerNorm, Dropout (p = 0.05)} &  \\
		\cmidrule{5-10}    \multicolumn{4}{c|}{}         & \multicolumn{6}{c|}{Multi-head ProbSparse Attention (h = 8, d = 64)} &  \\
		\cmidrule{5-10}    \multicolumn{4}{c|}{}         & \multicolumn{6}{c|}{Add, LayerNorm, Dropout (p = 0.05)} &  \\
		\cmidrule{5-10}    \multicolumn{4}{c|}{}         & \multicolumn{6}{c|}{\textbf{Surrogate FFN Block}, GELU} &  \\
		\cmidrule{5-10}    \multicolumn{4}{c|}{}         & \multicolumn{6}{c|}{Add, LayerNorm, Dropout (p = 0.05)} &  \\
		\midrule
		\textbf{Final:} &       &       & \multicolumn{1}{c}{} & \multicolumn{1}{c}{} &       &       &       &       & \multicolumn{1}{c}{} &  \\
		\midrule
		\multicolumn{4}{c|}{Outputs} & \multicolumn{6}{c|}{FCN}             &  \\
		\bottomrule
	\end{tabular}}%
	
\end{table}%
\begin{figure}[!h]
	\centering
	\includegraphics[width=0.7\linewidth]{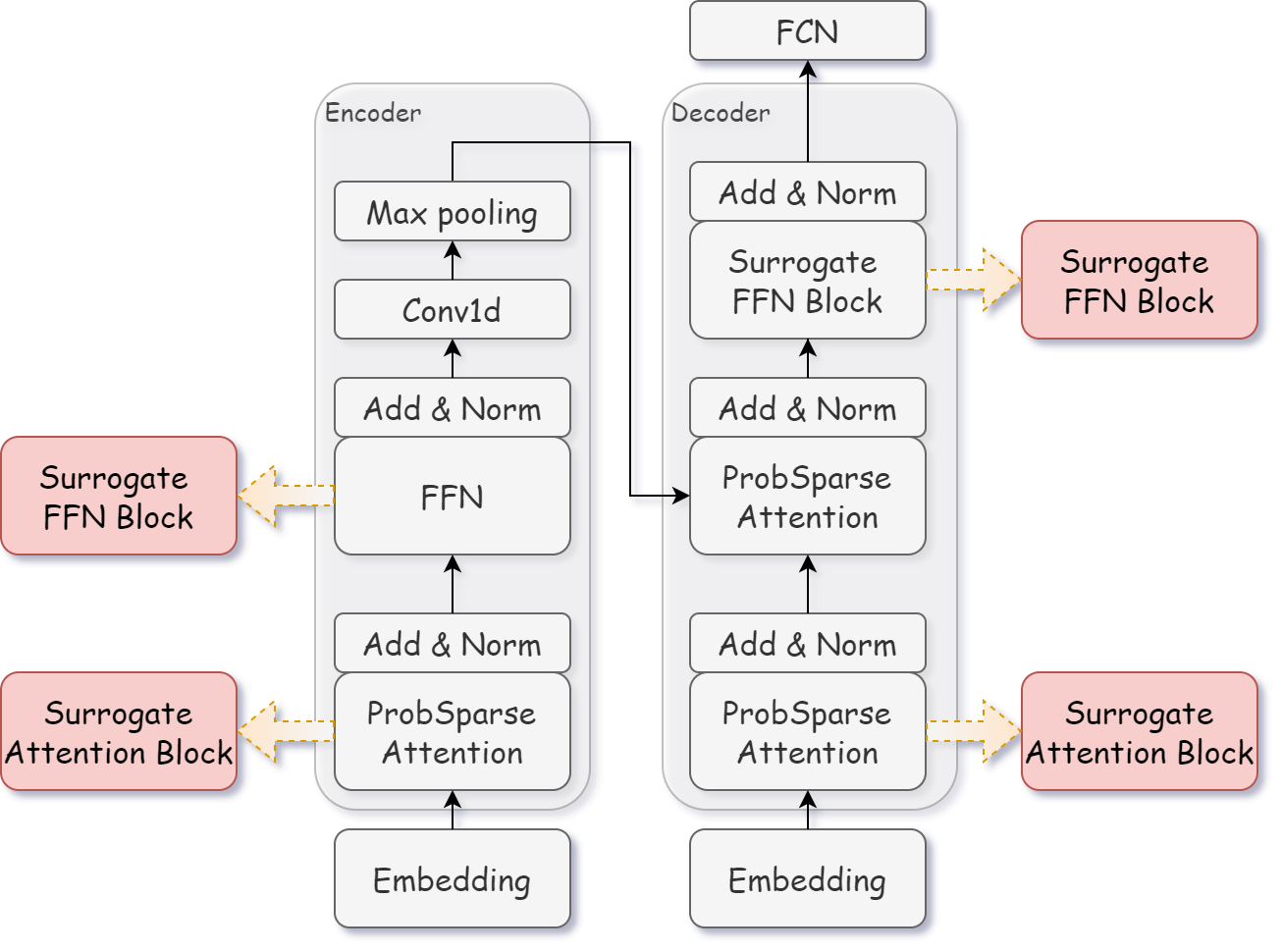}
	\caption{Architecture of the improved \underline{Informer} with structured matrices. The \fcolorbox{myred}{mypink}{red blocks} represent the replaced blocks.}
	\label{fig:xforemr-informer}
\end{figure}
\clearpage

\begin{table}[h]
	\centering
	\caption{Details of the improved \underline{Autoformer} with structured matrices.}
	{\tiny \begin{tabular}{lccc|lccccc|c}
		\toprule
		\textbf{Encoder:} &       &       & \multicolumn{1}{c}{} & \multicolumn{1}{c}{} &       &       &       &       & \multicolumn{1}{c}{} & \multicolumn{1}{c}{N} \\
		\midrule
		\multicolumn{4}{c|}{Inputs} & \multicolumn{2}{c|}{1 × 3 Conv1d} & \multicolumn{4}{c|}{Embedding(d = 512)} &  \\
		\midrule
		\multicolumn{4}{c|}{\multirow{6}[12]{*}{Auto-Correlation Block}} & \multicolumn{6}{c|}{\textbf{Surrogate Attention Block} (h = 8, d = 64)} & \multirow{6}[12]{*}{2} \\ 
		\cmidrule{5-10}    \multicolumn{4}{c|}{}         &
		\multicolumn{6}{c|}{Add, Dropout (p = 0.05)} &  \\
		\cmidrule{5-10}    \multicolumn{4}{c|}{}         & \multicolumn{6}{c|}{Series decomposition (moving avg = 25)} &  \\
		\cmidrule{5-10}    \multicolumn{4}{c|}{}         & \multicolumn{6}{c|}{\textbf{Surrogate FFN Block}, GELU} &  \\
		\cmidrule{5-10}    \multicolumn{4}{c|}{}         & \multicolumn{6}{c|}{Add, Dropout (p = 0.05)} &  \\
		\cmidrule{5-10}    \multicolumn{4}{c|}{}         & \multicolumn{6}{c|}{Series decomposition (moving avg = 25)} &  \\
		\midrule
		\textbf{Decoder:} &       &       & \multicolumn{1}{c}{} & \multicolumn{1}{c}{} &       &       &       &       & \multicolumn{1}{c}{} & \multicolumn{1}{c}{N} \\
		\midrule
		\multicolumn{4}{c|}{\multirow{2}[4]{*}{Inputs}} & \multicolumn{6}{c|}{Series decomposition (moving avg = 25)} & \multirow{2}[4]{*}{} \\
		\cmidrule{5-10}    \multicolumn{4}{c|}{}         & \multicolumn{2}{c|}{1 × 3 Conv1d} & \multicolumn{4}{c|}{Embedding(d = 512)} &  \\
		\midrule
		\multicolumn{4}{c|}{Masked Auto-Correlation Block} & \multicolumn{6}{c|}{add Mask on Auto-Correlation Block} & \multirow{10}[20]{*}{1} \\
		\cmidrule{1-10}    \multicolumn{4}{c|}{\multirow{9}[18]{*}{Auto-Correlation Block}} & \multicolumn{6}{c|}{\textbf{Surrogate Attention Block} (h = 8, d = 64)} &  \\
		\cmidrule{5-10}    \multicolumn{4}{c|}{}         & \multicolumn{6}{c|}{Add, Dropout (p = 0.05)} &  \\
		\cmidrule{5-10}    \multicolumn{4}{c|}{}         & \multicolumn{6}{c|}{Series decomposition (moving avg = 25)} &  \\
		\cmidrule{5-10}    \multicolumn{4}{c|}{}         & \multicolumn{6}{c|}{Auto-Correlation (h = 8, d = 64)} &  \\
		\cmidrule{5-10}    \multicolumn{4}{c|}{}         & \multicolumn{6}{c|}{Add, Dropout (p = 0.05)} &  \\
		\cmidrule{5-10}    \multicolumn{4}{c|}{}         & \multicolumn{6}{c|}{Series decomposition (moving avg = 25)} &  \\
		\cmidrule{5-10}    \multicolumn{4}{c|}{}         & \multicolumn{6}{c|}{\textbf{Surrogate FFN Block}, GELU} &  \\
		\cmidrule{5-10}    \multicolumn{4}{c|}{}         & \multicolumn{6}{c|}{Add, Dropout (p = 0.05)} &  \\
		\cmidrule{5-10}    \multicolumn{4}{c|}{}         & \multicolumn{6}{c|}{Series decomposition (moving avg = 25)} &  \\
		\midrule
		Final: &       &       & \multicolumn{1}{c}{} & \multicolumn{1}{c}{} &       &       &       &       & \multicolumn{1}{c}{} &  \\
		\midrule
		\multicolumn{4}{c|}{Outputs} & \multicolumn{6}{c|}{FCN}             &  \\
		\bottomrule
	\end{tabular}}%
	
\end{table}%
\begin{figure}[H]
	\centering
	\includegraphics[width=0.5\linewidth]{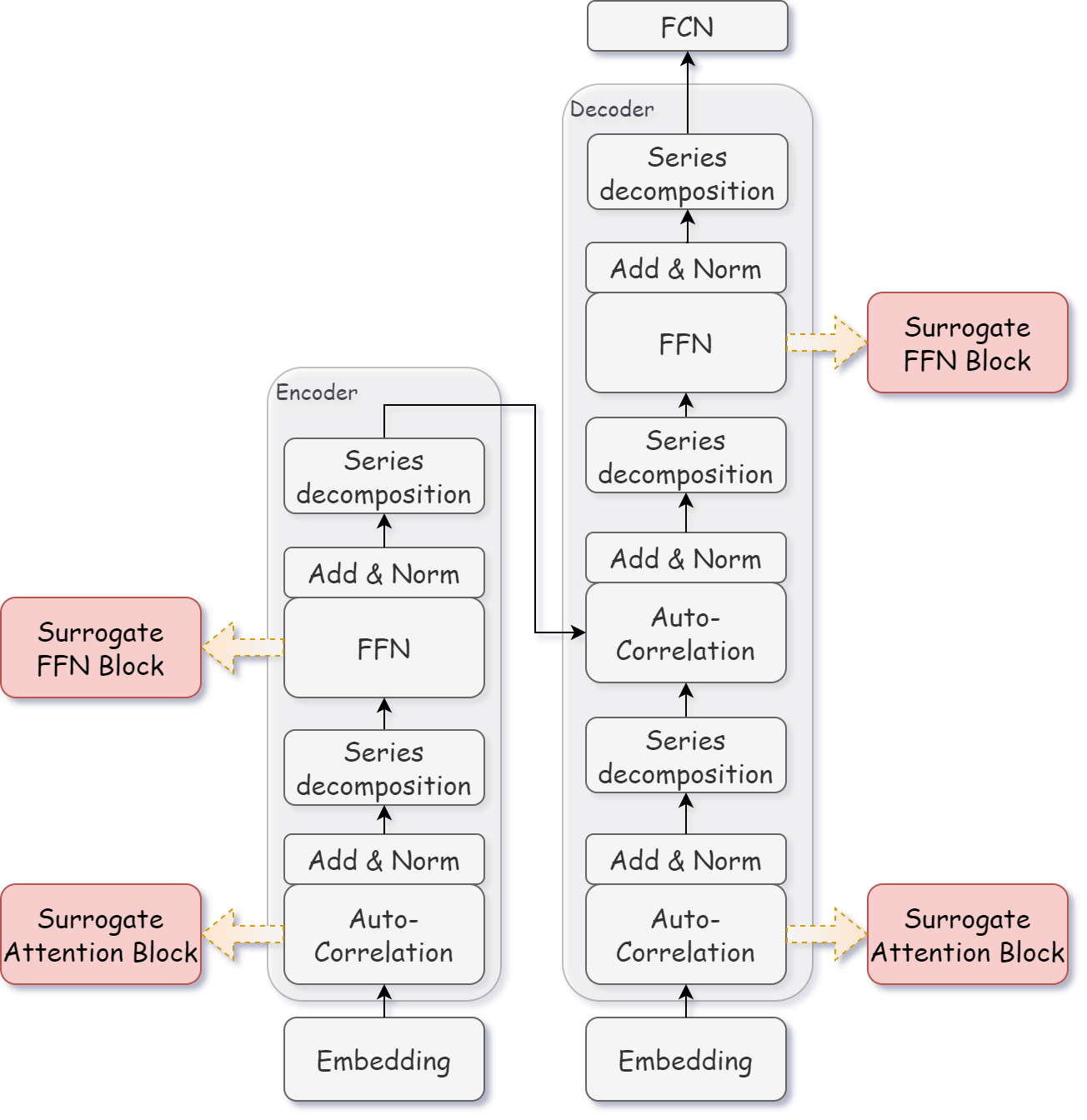}
	\caption{Architecture of the improved \underline{Autoformer} with structured matrices. The \fcolorbox{myred}{mypink}{red blocks} represent the replaced blocks.}
	\label{fig:xforemr-autoformer}
\end{figure}
\clearpage

\begin{table}[h]
	\centering
	\caption{Details of the improved \underline{FEDformer} with structured matrices.}
	{\tiny \begin{tabular}{lccc|lccccc|c}
		\toprule
		\textbf{Encoder:} &       &       & \multicolumn{1}{c}{} & \multicolumn{1}{c}{} &       &       &       &       & \multicolumn{1}{c}{} & \multicolumn{1}{c}{N} \\
		\midrule
		\multicolumn{4}{c|}{\multirow{3}[6]{*}{Inputs}} & \multicolumn{6}{c|}{Series\_decomp(kernel\_size=24)} & \multirow{3}[6]{*}{} \\
		\cmidrule{5-10}    \multicolumn{4}{c|}{}         & \multicolumn{6}{c|}{TokenEmbedding(d\_model=512)} &  \\
		\cmidrule{5-10}    \multicolumn{4}{c|}{}         & \multicolumn{6}{c|}{TemporalEmbedding(d\_model=512)} &  \\
		\midrule
		\multicolumn{4}{c|}{\multirow{6}[12]{*}{Frequency Enhanced Block}} & \multicolumn{6}{c|}{\textbf{Surrogate Attention Block} (h = 8, d = 64)}      & \multirow{6}[12]{*}{2} \\
		\cmidrule{5-10}    \multicolumn{4}{c|}{}         & \multicolumn{6}{c|}{Add, Dropout (p = 0.05)} &  \\
		\cmidrule{5-10}    \multicolumn{4}{c|}{}         & \multicolumn{6}{c|}{Series\_decomp(kernel\_size=24)} &  \\
		\cmidrule{5-10}    \multicolumn{4}{c|}{}         & \multicolumn{6}{c|}{\textbf{Surrogate FFN Block}}    &  \\
		\cmidrule{5-10}    \multicolumn{4}{c|}{}         & \multicolumn{6}{c|}{Add, Dropout (p = 0.05)} &  \\
		\cmidrule{5-10}    \multicolumn{4}{c|}{}         & \multicolumn{6}{c|}{Series\_decomp (kernel\_size=24)} &  \\
		\midrule
		\textbf{Decoder:} &       &       & \multicolumn{1}{c}{} & \multicolumn{1}{c}{} &       &       &       &       & \multicolumn{1}{c}{} & \multicolumn{1}{c}{N} \\
		\midrule
		\multicolumn{4}{c|}{\multirow{4}[8]{*}{Inputs}} & \multicolumn{6}{c|}{Series\_decomp(kernel\_size=24)} & \multirow{3}[6]{*}{} \\
		\cmidrule{5-10}    \multicolumn{4}{c|}{}         & \multicolumn{6}{c|}{TokenEmbedding(d\_model=512)} &  \\
		\cmidrule{5-10}    \multicolumn{4}{c|}{}         & \multicolumn{6}{c|}{TemporalEmbedding(d\_model=512)} &  \\
		\cmidrule{5-10}    \multicolumn{4}{c|}{}         & \multicolumn{6}{c|}{The output of Encoder} &  \\
		\midrule
		\multicolumn{4}{c|}{\multirow{3}[6]{*}{Frequency Enhanced Block}} & \multicolumn{6}{c|}{\textbf{Surrogate Attention Block} (h = 8, d = 64)}      & \multirow{11}[22]{*}{1} \\
		\cmidrule{5-10}    \multicolumn{4}{c|}{}         & \multicolumn{6}{c|}{Add, Dropout (p = 0.05)} &  \\
		\cmidrule{5-10}    \multicolumn{4}{c|}{}         & \multicolumn{6}{c|}{Series\_decomp (kernel\_size=24)} &  \\
		\cmidrule{1-10}    \multicolumn{4}{c|}{\multirow{8}[16]{*}{Frequency Enhanced Attention}} & \multicolumn{6}{c|}{Projection (d = 512)}      &  \\
		\cmidrule{5-10}    \multicolumn{4}{c|}{}         & \multicolumn{6}{c|}{MultiWaveletCross/ FourierCrossAttention  (h = 8, d = 64)} &  \\
		\cmidrule{5-10}    \multicolumn{4}{c|}{}         & \multicolumn{6}{c|}{Projection (d = 512)}      &  \\
		\cmidrule{5-10}    \multicolumn{4}{c|}{}         & \multicolumn{6}{c|}{Add, Dropout (p = 0.05)} &  \\
		\cmidrule{5-10}    \multicolumn{4}{c|}{}         & \multicolumn{6}{c|}{Series\_decomp (kernel\_size=24)} &  \\
		\cmidrule{5-10}    \multicolumn{4}{c|}{}         & \multicolumn{6}{c|}{\textbf{Surrogate FFN Block}}    &  \\
		\cmidrule{5-10}    \multicolumn{4}{c|}{}         & \multicolumn{6}{c|}{Add, Dropout (p = 0.05)} &  \\
		\cmidrule{5-10}    \multicolumn{4}{c|}{}         & \multicolumn{6}{c|}{Series\_decomp (kernel\_size=24)} &  \\
		\midrule
		\textbf{Final:} &       &       & \multicolumn{1}{c}{} & \multicolumn{1}{c}{} &       &       &       &       & \multicolumn{1}{c}{} &  \\
		\midrule
		\multicolumn{4}{c|}{Outputs} & \multicolumn{6}{c|}{FCN}             &  \\
		\bottomrule
	\end{tabular}}%
	
\end{table}%
\begin{figure}[H]
	\centering
	\includegraphics[width=1.0\linewidth]{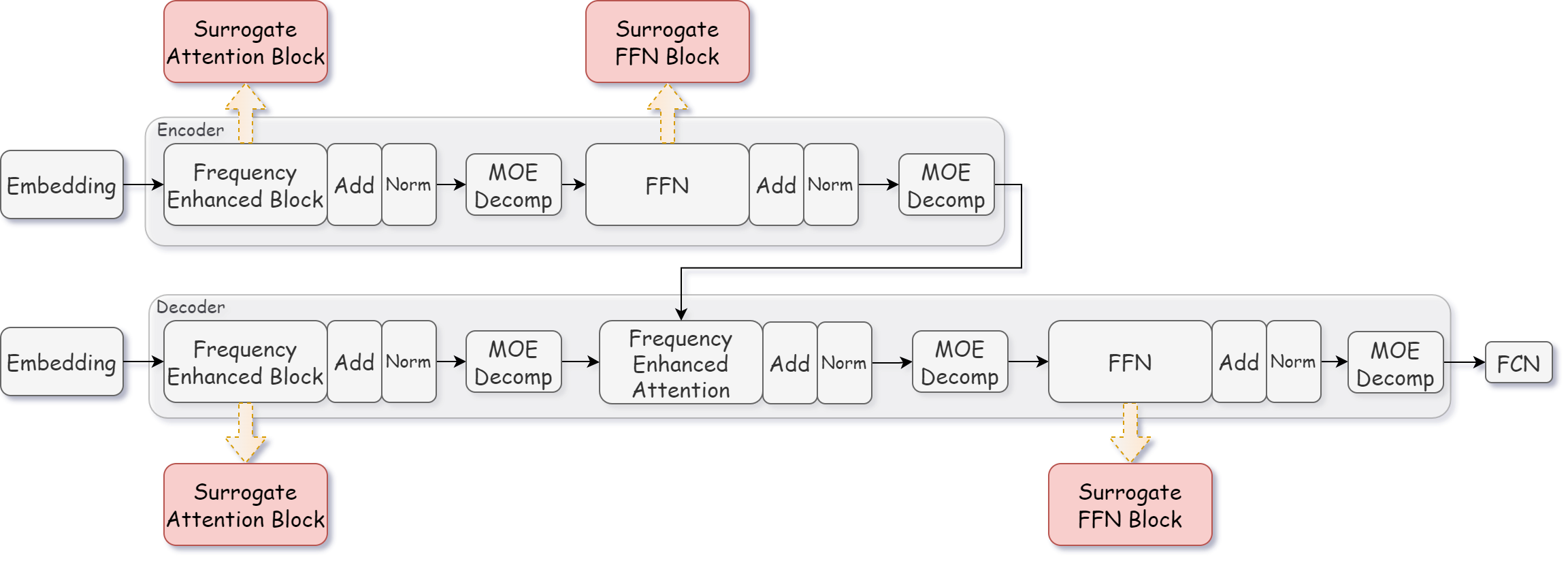}
	\caption{Architecture of the improved \underline{FEDformer} with structured matrices. The \fcolorbox{myred}{mypink}{red blocks} represent the replaced blocks.}
	\label{fig:xforemr-fedformer}
\end{figure}
\clearpage

\begin{table}[htbp]
	\centering
	\caption{Details of the improved \underline{Crossformer} with structured matrices.}
	\resizebox{\textwidth}{!}{
	\begin{tabular}{cccc|cccccc|c}
		\toprule
		\textbf{Encoder:} &       &       & \multicolumn{1}{c}{} & \multicolumn{1}{c}{} &       &       &       &       & \multicolumn{1}{c}{} & \multicolumn{1}{c}{N} \\
		\midrule
		\multicolumn{4}{c|}{\multirow{2}[4]{*}{Inputs}} & \multicolumn{6}{c|}{DSW\_embedding(seg\_len=6, d\_model=256)} & \multirow{2}[4]{*}{} \\
		\cmidrule{5-10}    \multicolumn{4}{c|}{}         & \multicolumn{6}{c|}{LayerNorm(d\_model=256)} &  \\
		\midrule
		\multicolumn{4}{c|}{\multirow{2}[4]{*}{Scale Block}} & \multicolumn{6}{c|}{SegMerging(d\_model=265, win\_size=2, nn.LayerNorm)} & \multirow{2}[4]{*}{3} \\
		\cmidrule{5-10}    \multicolumn{4}{c|}{}         & \multicolumn{6}{c|}{TwoStageAttentionLayer(seg\_num=6, factor=10, d\_model=256, n\_heads=4,  d\_ff=512, dropout=0.2)} &  \\
		\midrule
		\textbf{Decoder:} &       &       & \multicolumn{1}{c}{} & \multicolumn{1}{c}{} &       &       &       &       & \multicolumn{1}{c}{} & \multicolumn{1}{c}{N} \\
		\midrule
		\multicolumn{4}{c|}{Inputs} & \multicolumn{6}{c|}{The output of Encoder , The position of embedding} &  \\
		\midrule
		\multicolumn{4}{c|}{\multirow{7}[14]{*}{TwoStageAttentionLayer}} & \multicolumn{6}{c|}{\textbf{Surrogate Attention Block}(d\_model=256, n\_heads=4, dropout = 0.2)} & \multirow{13}[26]{*}{4} \\
		\cmidrule{5-10}    \multicolumn{4}{c|}{}         & \multicolumn{6}{c|}{Add,Dropout(p=0.2) LayerNorm(d =256)} &  \\
		\cmidrule{5-10}    \multicolumn{4}{c|}{}         & \multicolumn{6}{c|}{AttentionLayer(d\_model=256, n\_heads=4, dropout = 0.2)} &  \\
		\cmidrule{5-10}    \multicolumn{4}{c|}{}         & \multicolumn{6}{c|}{AttentionLayer(d\_model=256, n\_heads=4, dropout = 0.2)} &  \\
		\cmidrule{5-10}    \multicolumn{4}{c|}{}         & \multicolumn{6}{c|}{Add,Dropout(p=0.2),LayerNorm(d =256)} &  \\
		\cmidrule{5-10}    \multicolumn{4}{c|}{}         & \multicolumn{6}{c|}{\textbf{Surrogate FFN Block}, GELU} &  \\
		\cmidrule{5-10}    \multicolumn{4}{c|}{}         & \multicolumn{6}{c|}{Add,Dropout(p=0.2),LayerNorm(d =256)} &  \\
		\cmidrule{1-10}    \multicolumn{4}{c|}{\multirow{6}[12]{*}{}} & \multicolumn{6}{c|}{AttentionLayer(d\_model=256, n\_heads=4, dropout = 0.2)} &  \\
		\cmidrule{5-10}    \multicolumn{4}{c|}{}         & \multicolumn{6}{c|}{Dropout(p=0.2)}  &  \\
		\cmidrule{5-10}    \multicolumn{4}{c|}{}         & \multicolumn{6}{c|}{LayerNorm(d\_mode=256)} &  \\
		\cmidrule{5-10}    \multicolumn{4}{c|}{}         & \multicolumn{6}{c|}{\textbf{Surrogate FFN Block}, GELU} &  \\
		\cmidrule{5-10}    \multicolumn{4}{c|}{}         & \multicolumn{6}{c|}{LayerNorm(d\_model=256)} &  \\
		\cmidrule{5-10}    \multicolumn{4}{c|}{}         & \multicolumn{6}{c|}{Linear(d\_model=256, seg\_len=6)} &  \\
		\midrule
		\textbf{Final}: &       &       & \multicolumn{1}{c}{} & \multicolumn{1}{c}{} &       &       &       &       & \multicolumn{1}{c}{} &  \\
		\midrule
		\multicolumn{4}{c|}{Outputs} & \multicolumn{6}{c|}{FCN}             &  \\
		\bottomrule
	\end{tabular}%
	}
	
\end{table}%
\begin{figure}[H]
	\centering
	\includegraphics[width=0.45\linewidth]{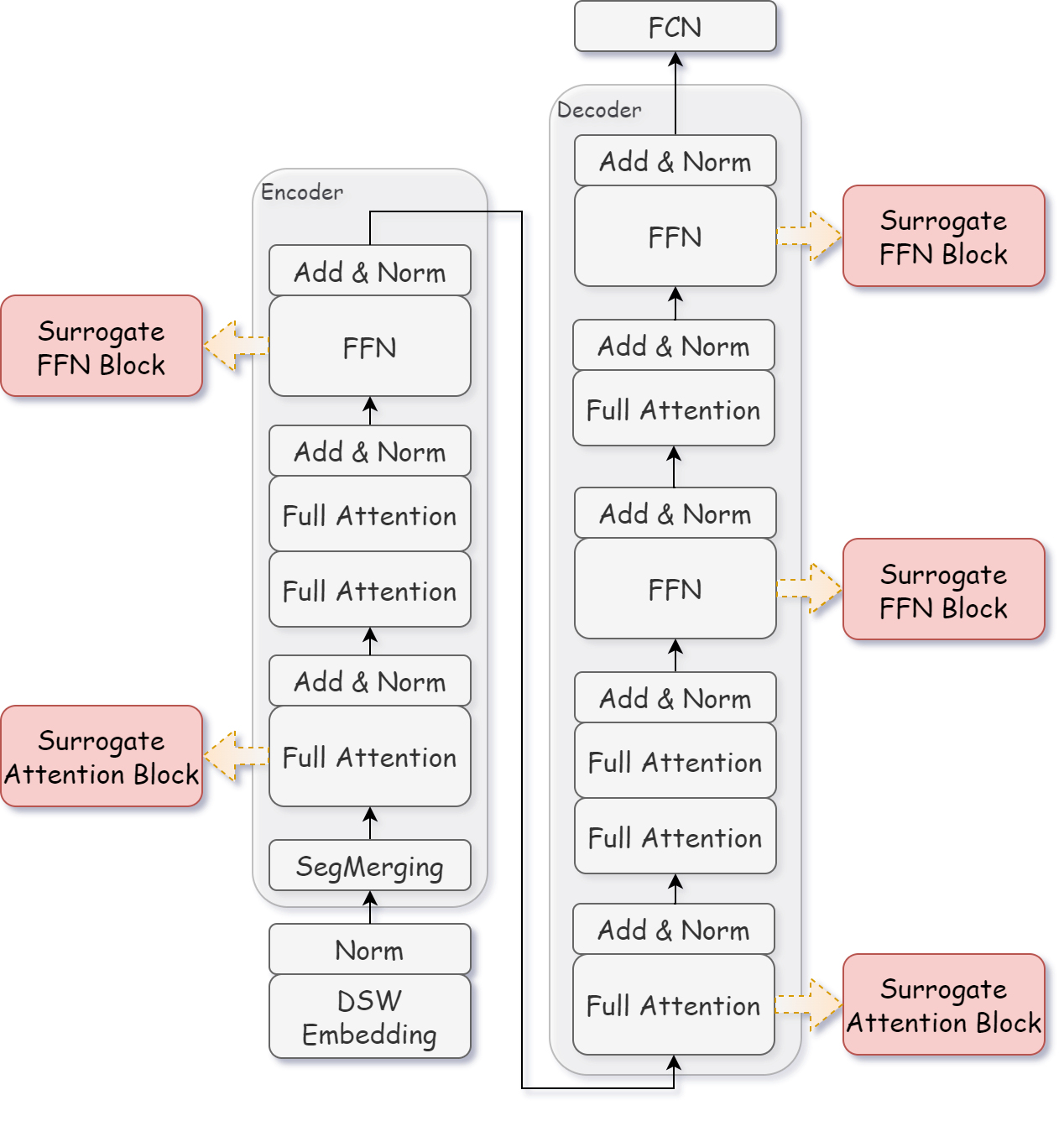}
	\caption{Architecture of the improved \underline{Crossformer} with structured matrices. The \fcolorbox{myred}{mypink}{red blocks} represent the replaced blocks.}
	\label{fig:xforemr-crossformer}
\end{figure}
\clearpage

\begin{table}[htbp]
	\centering
	\caption{Details of the improved \underline{Pyraformer} with structured matrices.}
	\resizebox{\textwidth}{!}{
	\begin{tabular}{cccc|cccccc|c}
		\toprule
		\textbf{Encoder:} &       &       & \multicolumn{1}{c}{} & \multicolumn{1}{c}{} &       &       &       &       & \multicolumn{1}{c}{} & \multicolumn{1}{c}{N} \\
		\midrule
		\multicolumn{4}{c|}{\multirow{3}[6]{*}{Inputs}} & \multicolumn{6}{c|}{Embedding(d=512)} & \multirow{3}[6]{*}{} \\
		\cmidrule{5-10}    \multicolumn{4}{c|}{}         & \multicolumn{6}{c|}{Mask(input\_size=168/169, window\_size= '[4, 4, 4]',inner\_size=3)} &  \\
		\cmidrule{5-10}    \multicolumn{4}{c|}{}         & \multicolumn{6}{c|}{Bottleneck\_Construct(d\_model=512, d\_inner=512, window\_size= '[4, 4, 4]')} &  \\
		\midrule
		\multicolumn{4}{c|}{Attention} & \multicolumn{6}{c|}{\textbf{Surrogate Attention Block}(n\_head=4, d\_model=512,d)} & \multirow{4}[8]{*}{} \\
		\cmidrule{1-10}    \multicolumn{4}{c|}{\multirow{3}[6]{*}{PositionwiseFeedForward}} & \multicolumn{6}{c|}{LayerNorm(d\_in=512, eps=1e-6)} &  \\
		\cmidrule{5-10}    \multicolumn{4}{c|}{}         & \multicolumn{6}{c|}{\textbf{Surrogate FFN Block}, GELU} &  \\
		\cmidrule{5-10}    \multicolumn{4}{c|}{}         & \multicolumn{6}{c|}{LayerNorm(d\_in=512, eps=1e-6)} &  \\
		\midrule
		\textbf{Decoder:} &       &       & \multicolumn{1}{c}{} & \multicolumn{1}{c}{} &       &       &       &       & \multicolumn{1}{c}{} & \multicolumn{1}{c}{N} \\
		\midrule
		\multicolumn{4}{c|}{\multirow{2}[4]{*}{Inputs}} & \multicolumn{6}{c|}{Embedding(d=512)} & \multirow{2}[4]{*}{} \\
		\cmidrule{5-10}    \multicolumn{4}{c|}{}         & \multicolumn{6}{c|}{Mask(input\_size=168/169, window\_size= '[4, 4, 4]',inner\_size=3) } &  \\
		\midrule
		\multicolumn{4}{c|}{Attention} & \multicolumn{6}{c|}{\textbf{Surrogate Attention Block}(n\_head=4, d\_model=512)} & \multirow{4}[8]{*}{2} \\
		\cmidrule{1-10}    \multicolumn{4}{c|}{\multirow{3}[6]{*}{PositionwiseFeedForward}} & \multicolumn{6}{c|}{LayerNorm(d\_in=512, eps=1e-6)} &  \\
		\cmidrule{5-10}    \multicolumn{4}{c|}{}         & \multicolumn{6}{c|}{\textbf{Surrogate FFN Block}, GELU} &  \\
		\cmidrule{5-10}    \multicolumn{4}{c|}{}         & \multicolumn{6}{c|}{LayerNorm(d\_in=512, eps=1e-6)} &  \\
		\midrule
		\textbf{Final}: &       &       & \multicolumn{1}{c}{} & \multicolumn{1}{c}{} &       &       &       &       & \multicolumn{1}{c}{} &  \\
		\midrule
		\multicolumn{4}{c|}{Outputs} & \multicolumn{6}{c|}{Linear(dim=512/2048)} &  \\
		\bottomrule
	\end{tabular}%
	}
	
\end{table}%
\begin{figure}[H]
	\centering
	\includegraphics[width=0.3\linewidth]{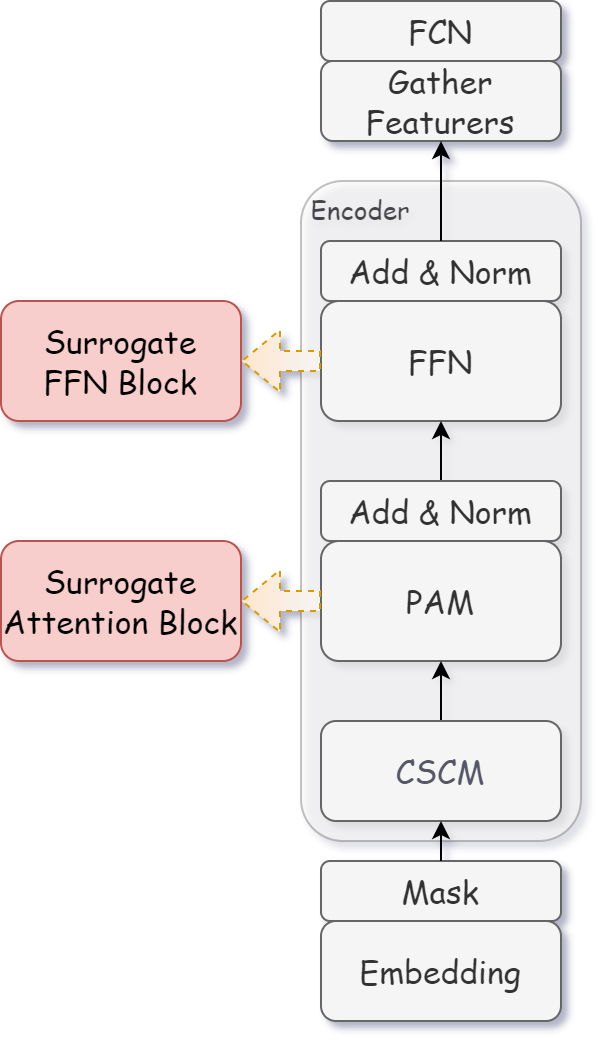}
	\caption{Architecture of the improved \underline{Pyraformer} with structured matrices. The \fcolorbox{myred}{mypink}{red blocks} represent the replaced blocks.}
	\label{fig:xforemr-pyraformer}
\end{figure}
\clearpage

\begin{table}[htbp]
	\centering
	\caption{Details of the improved \underline{Non-stationary Transformers} with structured matrices.}
	{\tiny \begin{tabular}{lccc|lccccc|c}
		\toprule
		\textbf{Encoder:} &       &       & \multicolumn{1}{c}{} & \multicolumn{1}{c}{} &       &       &       &       & \multicolumn{1}{c}{} & \multicolumn{1}{c}{N} \\
		\midrule
		\multicolumn{4}{c|}{\multirow{2}[4]{*}{Inputs}} & \multicolumn{6}{c|}{Normaliztion (dim = 0)} & \multirow{6}[12]{*}{2} \\
		\cmidrule{5-10}    \multicolumn{4}{c|}{}         & \multicolumn{6}{c|}{Embedding(d = 512)} &  \\
		\cmidrule{1-10}    \multicolumn{4}{c|}{\multirow{4}[8]{*}{}} & \multicolumn{6}{c|}{\textbf{Surrogate Attention Block} (h = 8, d = 64)} &  \\
		\cmidrule{5-10}    \multicolumn{4}{c|}{}         & \multicolumn{6}{c|}{Add \& Norm}     &  \\
		\cmidrule{5-10}    \multicolumn{4}{c|}{}         & \multicolumn{6}{c|}{\textbf{Surrogate FFN Block}, GELU} &  \\
		\cmidrule{5-10}    \multicolumn{4}{c|}{}         & \multicolumn{6}{c|}{Add \& Norm}     &  \\
		\midrule
		\textbf{Decoder:} &       &       & \multicolumn{1}{c}{} & \multicolumn{1}{c}{} &       &       &       &       & \multicolumn{1}{c}{} & \multicolumn{1}{c}{N} \\
		\midrule
		\multicolumn{4}{c|}{Inputs} & \multicolumn{2}{c|}{ConCat} & \multicolumn{4}{c|}{Embedding(d = 512)} &  \\
		\midrule
		\multicolumn{4}{c|}{\multirow{6}[12]{*}{}} & \multicolumn{6}{c|}{\textbf{Surrogate Attention Block} (h = 8, d = 64)} & \multirow{6}[12]{*}{1} \\
		\cmidrule{5-10}    \multicolumn{4}{c|}{}         & \multicolumn{6}{c|}{Add \& Norm}     &  \\
		\cmidrule{5-10}    \multicolumn{4}{c|}{}         & \multicolumn{6}{c|}{De-stationary Attention ($\tau, \varDelta = 0$)} &  \\
		\cmidrule{5-10}    \multicolumn{4}{c|}{}         & \multicolumn{6}{c|}{Add \& Norm}     &  \\
		\cmidrule{5-10}    \multicolumn{4}{c|}{}         & \multicolumn{6}{c|}{\textbf{Surrogate FFN Block}, GELU} &  \\
		\cmidrule{5-10}    \multicolumn{4}{c|}{}         & \multicolumn{6}{c|}{Add \& Norm}     &  \\
		\midrule
		\textbf{Final:} &       &       & \multicolumn{1}{c}{} & \multicolumn{1}{c}{} &       &       &       &       & \multicolumn{1}{c}{} &  \\
		\midrule
		\multicolumn{4}{c|}{Outputs} & \multicolumn{6}{c|}{De-normaliztion ($\mu_x, \sigma_x$)} &  \\
		\bottomrule
	\end{tabular}}%
	
\end{table}%
\begin{figure}[H]
	\centering
	\includegraphics[width=0.7\linewidth]{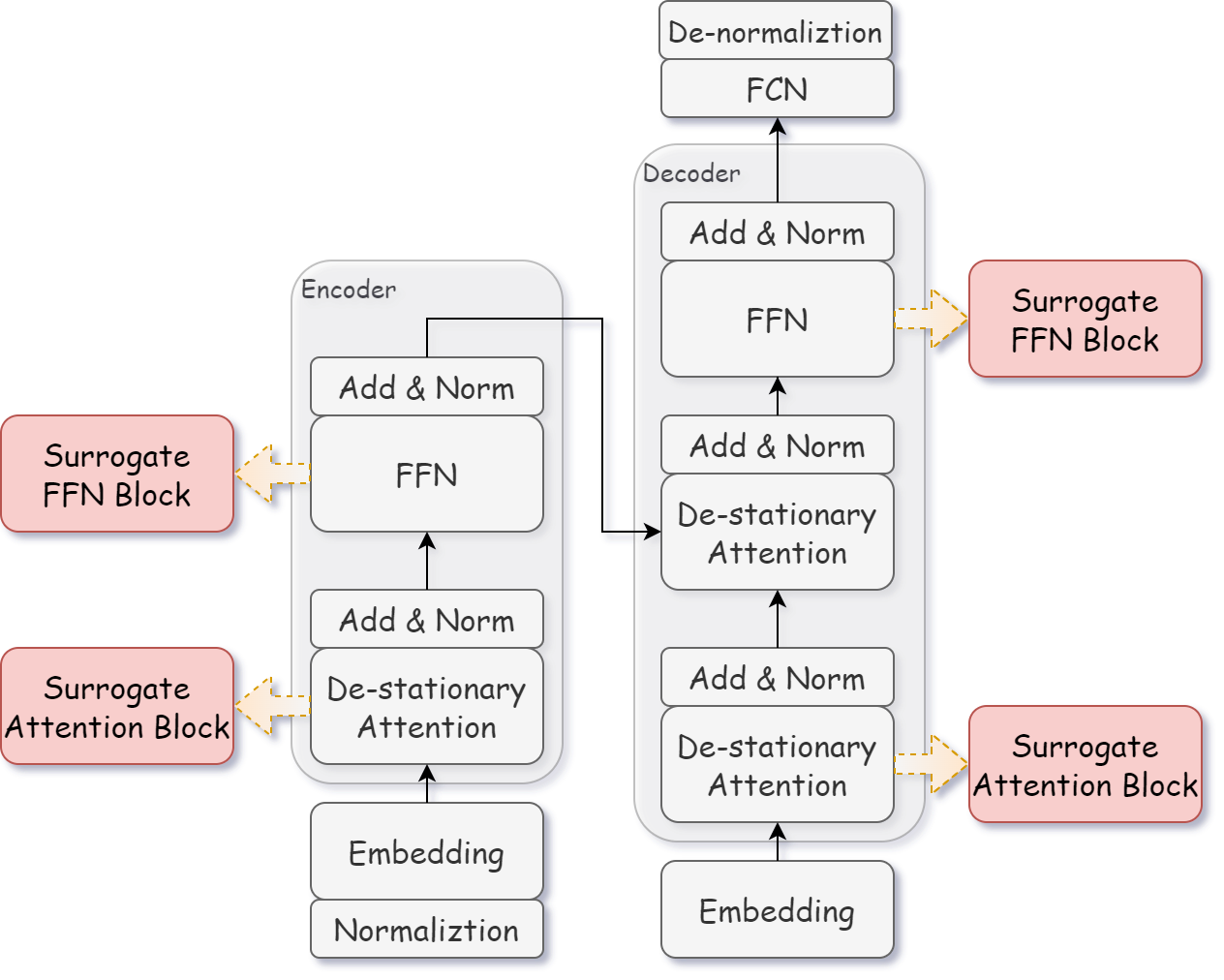}
	\caption{Architecture of the improved \underline{Non-stationary Transformers} with structured matrices. The \fcolorbox{myred}{mypink}{red blocks} represent the replaced blocks.}
	\label{fig:xforemr-nst}
\end{figure}
\clearpage

\begin{table}[htbp]
	\centering
	\caption{Details of the improved \underline{PatchTST} with structured matrices.}
	{\tiny \begin{tabular}{lccc|lccccc|c}
		\toprule
		\textbf{Encoder:} &       &       & \multicolumn{1}{c}{} & \multicolumn{1}{c}{} &       &       &       &       & \multicolumn{1}{c}{} & \multicolumn{1}{c}{N} \\
		\midrule
		\multicolumn{4}{c|}{\multirow{3}[6]{*}{Inputs}} & \multicolumn{6}{c|}{Normaliztion (dim = 0)} & \multirow{8}[16]{*}{2} \\
		\cmidrule{5-10}    \multicolumn{4}{c|}{}         & \multicolumn{6}{c|}{Patching (patch\_len=16, stride=8)}        &  \\
		\cmidrule{5-10}    \multicolumn{4}{c|}{}         & \multicolumn{6}{c|}{Embedding (d = 512)} &  \\
		\cmidrule{1-10}    \multicolumn{4}{c|}{\multirow{5}[10]{*}{}} & \multicolumn{6}{c|}{\textbf{Surrogate Attention Block} (h = 8, d = 64)} &  \\
		\cmidrule{5-10}    \multicolumn{4}{c|}{}         & \multicolumn{6}{c|}{Add \& Norm}     &  \\
		\cmidrule{5-10}    \multicolumn{4}{c|}{}         & \multicolumn{6}{c|}{\textbf{Surrogate FFN Block}} &  \\
		\cmidrule{5-10}    \multicolumn{4}{c|}{}         & \multicolumn{6}{c|}{Add \& Norm}     &  \\
		\cmidrule{5-10}    \multicolumn{4}{c|}{}         & \multicolumn{6}{c|}{Reshape (n = Number of variables)}         &  \\
		\midrule
		\textbf{Decoder:} &       &       & \multicolumn{1}{c}{} & \multicolumn{1}{c}{} &       &       &       &       & \multicolumn{1}{c}{} & \multicolumn{1}{c}{N} \\
		\midrule
		\multicolumn{4}{c|}{\multirow{3}[5]{*}{}} & \multicolumn{6}{c|}{Flatten (start\_dim=-2)}         & \multirow{3}[5]{*}{1} \\
		\cmidrule{5-10}    \multicolumn{4}{c|}{}         & \multicolumn{6}{c|}{Linear Projection (d = pred\_len)} &  \\
		\cmidrule{5-10}    \multicolumn{4}{c|}{}         & \multicolumn{6}{c|}{Dropout (p = 0.05)}         &  \\
		\midrule
		\textbf{Final:} &       &       & \multicolumn{1}{c}{} & \multicolumn{1}{c}{} &       &       &       &       & \multicolumn{1}{c}{} &  \\
		\midrule
		\multicolumn{4}{c|}{Outputs} & \multicolumn{6}{c|}{De-normaliztion ($\mu_x, \sigma_x$)} &  \\
		\bottomrule
	\end{tabular}}%
	
\end{table}%
\begin{figure}[H]
	\centering
	\includegraphics[width=0.3\linewidth]{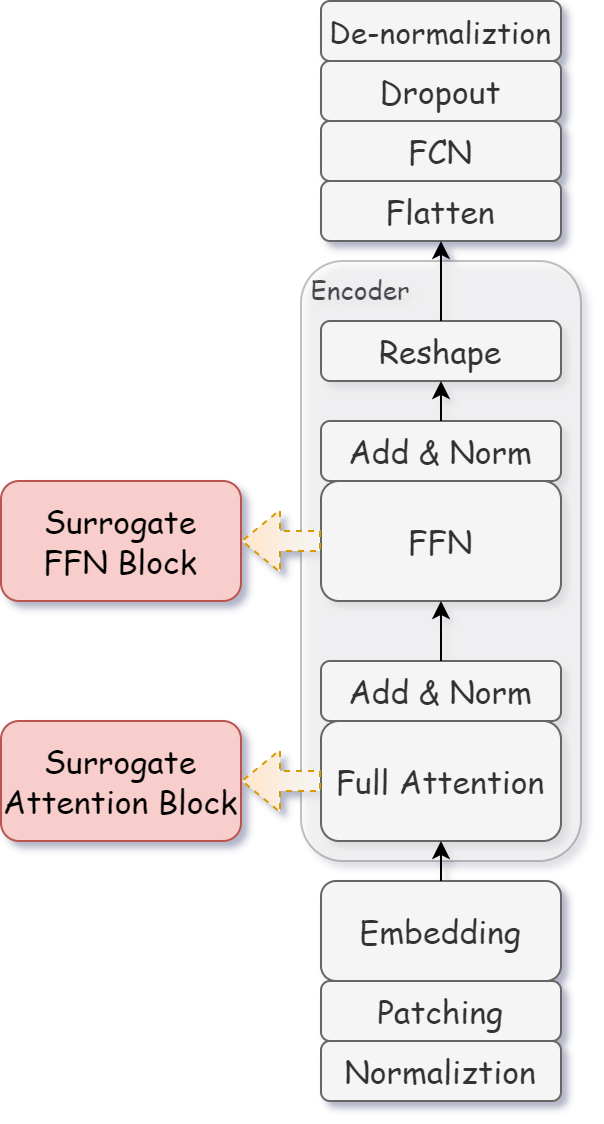}
	\caption{Architecture of the improved \underline{PatchTST} with structured matrices. The \fcolorbox{myred}{mypink}{red blocks} represent the replaced blocks.}
	\label{fig:xforemr-patchtst}
\end{figure}
\clearpage

\begin{table}[htbp]
	\centering
	\caption{Details of the improved \underline{iTransformer} with structured matrices.}
	\begin{tabular}{lccc|lccccc|c}
		\toprule
		\textbf{Encoder:} &       &       & \multicolumn{1}{c}{} & \multicolumn{1}{c}{} &       &       &       &       & \multicolumn{1}{c}{} & \multicolumn{1}{c}{N} \\
		\midrule
		\multicolumn{4}{c|}{\multirow{3}[6]{*}{Inputs}} & \multicolumn{6}{c|}{Normaliztion (dim = 0)} & \multirow{7}[14]{*}{2} \\
		\cmidrule{5-10}    \multicolumn{4}{c|}{}         & \multicolumn{6}{c|}{Inverse}         &  \\
		\cmidrule{5-10}    \multicolumn{4}{c|}{}         & \multicolumn{6}{c|}{Embedding(d = 512)} &  \\
		\cmidrule{1-10}    \multicolumn{4}{c|}{\multirow{4}[8]{*}{}} & \multicolumn{6}{c|}{\textbf{Surrogate Attention Block} (h = 8, d = 64)} &  \\
		\cmidrule{5-10}    \multicolumn{4}{c|}{}         & \multicolumn{6}{c|}{Add \& Norm}     &  \\
		\cmidrule{5-10}    \multicolumn{4}{c|}{}         & \multicolumn{6}{c|}{\textbf{Surrogate FFN Block}} &  \\
		\cmidrule{5-10}    \multicolumn{4}{c|}{}         & \multicolumn{6}{c|}{Add \& Norm}     &  \\
		\midrule
		\textbf{Decoder:} &       &       & \multicolumn{1}{c}{} & \multicolumn{1}{c}{} &       &       &       &       & \multicolumn{1}{c}{} & \multicolumn{1}{c}{N} \\
		\midrule
		\multicolumn{4}{c|}{}         & \multicolumn{6}{c|}{Linear Projection (d = pred\_len)} & 1 \\
		\midrule
		\textbf{Final:} &       &       & \multicolumn{1}{c}{} & \multicolumn{1}{c}{} &       &       &       &       & \multicolumn{1}{c}{} &  \\
		\midrule
		\multicolumn{4}{c|}{Outputs} & \multicolumn{6}{c|}{De-normaliztion ($\mu_x, \sigma_x$)} &  \\
		\bottomrule
	\end{tabular}%
	
\end{table}%
\begin{figure}[H]
	\centering
	\includegraphics[width=0.4\linewidth]{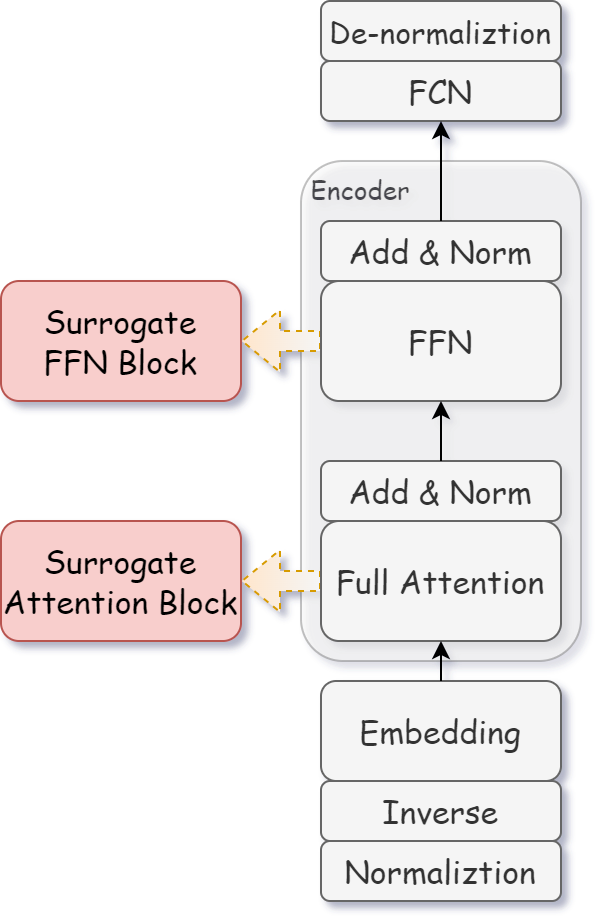}
	\caption{Architecture of the improved \underline{iTransformer} with structured matrices. The \fcolorbox{myred}{mypink}{red blocks} represent the replaced blocks.}
	\label{fig:xforemr-itransformer}
\end{figure}

\end{sloppypar}
\end{document}